\documentclass[lettersize,journal]{IEEEtran}
\usepackage{amsthm}
\usepackage{amsmath,amsfonts}
\usepackage{algorithmic}
\usepackage{algorithm}
\usepackage{array}
\usepackage[caption=false,font=normalsize,labelfont=sf,textfont=sf]{subfig}
\usepackage{textcomp}
\usepackage{stfloats}
\usepackage{url}
\usepackage{verbatim}
\usepackage{graphicx}
\usepackage{cite}
\usepackage{booktabs} 
\usepackage{balance}
\usepackage{amsmath}
\usepackage{epsfig}
\usepackage{graphicx}
\usepackage{algorithm}
\usepackage{algorithmic}
\usepackage{float}
\usepackage{subeqnarray}
\usepackage{cases}
\usepackage{amssymb}
\usepackage{amsmath}

\usepackage{bm}
\usepackage{url}
\usepackage{setspace}

\newtheorem{prop}{Proposition}
\newtheorem{con}{Condition}

\usepackage{multirow}
\usepackage{makecell}
\usepackage{threeparttable}
\usepackage{color}

\theoremstyle{definition}

\theoremstyle{remark}

\usepackage{pifont}
\begin{document}
	
	\title{ Optimization-Inspired Learning with  Architecture Augmentations and Control Mechanisms  for Low-Level Vision}
	
	\author{Risheng~Liu,~\IEEEmembership{Member,~IEEE,}
		Zhu~Liu, Pan~Mu, Xin~Fan,~\IEEEmembership{Senior Member,~IEEE,} 
		and~Zhongxuan~Luo
		\thanks{This work is partially supported by the  National Natural Science Foundation of China (No. U22B2052).}
		\thanks{ Risheng Liu is with the School of Software Technology and the
			Key Laboratory for Ubiquitous Network and Service Software of Liaoning
			Province, Dalian University of Technology, Dalian 116024, China. (Corresponding author,
			e-mail: rsliu@dlut.edu.cn).}
		\thanks{ Zhu Liu, Xin Fan and Zhongxuan Luo are with the School of Software Technology and the
			Key Laboratory for Ubiquitous Network and Service Software of Liaoning
			Province, Dalian University of Technology, Dalian 116024, China.  (email:
			liuzhu@mail.dlut.edu.cn, xin.fan@dlut.edu.cn and zxluo@dlut.edu.cn). }
		\thanks{  Pan Mu is with the College of Computer Science and Technology, Zhejiang
			University of Technology, Hangzhou 310023, China (e-mail: panmu@zjut.edu.cn).}
	}
	
	\markboth{Journal of \LaTeX\ Class Files,~Vol.~14, No.~8, August~2021}%
	{Shell \MakeLowercase{\textit{et al.}}: A Sample Article Using IEEEtran.cls for IEEE Journals}
	
	
	\maketitle
	
	\begin{abstract}
		In recent years, there has been a growing interest in combining learnable modules with numerical optimization to solve low-level vision tasks. However, most existing approaches focus on designing specialized schemes to generate image/feature propagation. There is a lack of unified consideration to construct propagative modules,  provide theoretical analysis tools, and design effective learning mechanisms.
To mitigate the above issues, this paper proposes a unified optimization-inspired learning framework to aggregate Generative, Discriminative, and Corrective (GDC for short) principles with strong generalization for diverse optimization models. Specifically, by introducing a general energy minimization model and formulating its descent direction from different viewpoints (\textit{i.e.,} in a generative manner, based on the discriminative metric and with optimality-based correction), we construct three propagative modules to effectively solve the optimization models with flexible combinations.
We design two control mechanisms that provide the non-trivial theoretical guarantees for both fully- and partially-defined optimization formulations. Under the support of theoretical guarantees, we can introduce diverse architecture augmentation strategies such as normalization and search to ensure stable propagation with convergence and seamlessly integrate the suitable modules into the propagation respectively. Extensive experiments across varied low-level vision tasks validate the efficacy and adaptability of GDC.
	\end{abstract}
	
	
	\section{Introduction}
	
	\IEEEPARstart{R}{ecent}   decades have witnessed dramatic advancements in low-level vision applications, ranging from image restoration~\cite{beck2009a,liu2020retinex}, enhancement~\cite{liu2020bilevel,liu2020investigating} and medical imaging~\cite{liubi}, just to name a few, based on numerical optimization~\cite{beck2009a,li2015accelerated}, deep learning~\cite{pan2022dual,gould2021deep} and  the combined learning methodologies~\cite{zhang2021plug,liu2022task}. 
	
	Among them, combining learnable modules and numerical optimizations has become the mainstream scheme for low-level vision tasks. These schemes can effectively aggregate the advantages of numerical optimization and learning-based methods. Specifically, optimization-based schemes~\cite{beck2009a,li2015accelerated} are flexible for diverse low-level vision tasks, generating solutions with concrete formation and task-related priors. However, these methods suffer from heavy numerical computation and are fully dependent on handcrafted priors, which may not be  regularized in real-world distribution. Learning-based methods~\cite{pan2022dual,gould2021deep} realize fast inference and high performance, benefiting from the learning of large-scale datasets.  These schemes are fully dependent on concrete CNN architectures and discard physical rules completely, limiting the generalization of adoption into other tasks.  
Combined schemes have obtained widespread attention in recent years.
Unfortunately, several obstacles limit the development of these methods.
	
 Most of these methods are designed by the direct combination of numerical solutions and trainable architectures, providing specialized schemes with fixed solving paradigms. For instance, the representative category is the unrolling-based methods.   Various CNN modules~\cite{zhang2017learning,zhang2021plug,liu2022task} have been directly integrated into numerical optimization iterations as the implicitly-defined data-driven priors, such as CNN-based denoisers for deconvolution~\cite{zhang2017learning}, Retinex-based modules for low-light enhancement~\cite{liu2022task}, and discriminative classifiers for deblurring~\cite{li2019blind}.
		Unfortunately,
current approaches design specialized and fixed solution paradigms, ignoring the roles of statistical priors. These fixed solutions are incapable of wide application
		scopes for diverse low-level tasks and scenarios.

	Furthermore, less progress has been made in constructing theoretical analysis tools to control propagative behaviors. Existing methods mostly leverage the data-driven priors to replace the
statistic priors in an implicit manner, solving the prior sub-problem using
learnable networks. Nevertheless, the original theoretical guarantee is inevitably broken by the difficult analysis with these data-driven modules. Therefore, it is difficult to rely on existing theoretical analysis tools to establish comprehensive analysis guarantees.  Recently, based on the assumption of bounded denoisers, Chan~\textit{et.al.}~\cite{Chan2017Plug} establish the convergence guarantee for the Plug-and-Play scenario. Liu~\textit{et.al.}~\cite{liu2019convergence} propose the scheduling policies with flexible iterative modularization for rigorous analysis of the convergence behaviors.  However, introducing learnable modules in the pre-trained or joint training manner would damage the optimization rule under diverse optimization models (w/ and w/o task-related priors).

 Moreover, the straightforward integration between learnable architectures and optimization is easy to lead to unstable propagation, due to the sensitive parameter changes of learnable architectures. Some methods aim to avoid unstable iterations by controlling the denoising levels ({\textit{e.g.,}} fixing at a small level~\cite{DongDnoiser} or
		adjusting denoising dynamically~\cite{zhang2021plug}). 	We argue that such manual adjustments cannot guarantee propagation toward the convergence and are not flexible for diverse low-level vision tasks.  In order to further improve performance, several works~\cite{zhang2021plug,zhang2020deep2} introduce more powerful learnable architectures by increasing the depth and width of the network. For instance, the enhanced version of IRCNN~\cite{zhang2021plug} replaces the shallow denoisers~\cite{zhang2017learning} with the combination of UNet and ResNet, without the theoretical convergences. Existing architectures are fragile with the tasks and data changes,  relying on customized architectures and training strategies,
			in order to stratify the theoretical convergences, that limit the performances
			for addressing low-level vision tasks.

	\subsection{Our Contributions}
	It can be observed that current methods provide specialized schemes for specific applications. These schemes have suffered from several fundamental limitations, including a lack of a unified framework to consider diverse optimization models, the shortage of tools for theoretical analyses, and effective learning strategies with convergence guarantees.
To overcome the above limitations, this paper proposes a unified optimization-inspired propagation framework for different low-level vision tasks.
	
	To be specific, we consider a general energy minimization model (\textit{i.e.,} Fully-Defined Model (FDM) with explicit priors and Partially-Defined Model (PDM) with implicitly defined priors) to abstractly formulate low-level vision tasks. We introduce three categories of principles to construct modules for investigating descent directions, \textit{i.e.,} from \textbf{G}enerative learning perspective, based on the \textbf{D}iscriminative metric and with model information \textbf{C}orrection, which can provide the generalized scheme for diverse low-level vision tasks. 

In order to overcome the convergence difficulties of the integrated propagations, we design a series of Propagative Control Mechanisms (PCM) to guide the propagation sequences toward our desired solutions. Convergence properties of GDC can be proved for both FDM and PDM, (\textit{i.e.,} critical point convergence for FDM and fixed point convergence for PDM), which is nontrivial, compared with existing works. Based on this, we can design various combinations of propagation modules, which largely extend the range of applications.

To avoid the conflicts of introducing learnable modules, the GDC framework is agile in introducing diverse Architecture Augmentation Strategies (AAS), including Architecture Normalization (AN) and Architecture Search (AS), to stabilize the propagation and discover the suitable architectures. We demonstrate that using architecture normalization can have fixed point convergences with arbitrary networks. Joint training and other augmentation techniques (\textit{e.g.,} search) can be performed for further improving performance.
Extensive experimental results demonstrate that our GDC achieves state-of-the-art performances on a variety of low-level vision tasks.
In summary, our contributions are listed as follows:
	\begin{itemize}
		\item  We provide a generic framework to integrate learning-based solutions and knowledge-based  optimization with sufficient theoretical analysis to address diverse optimization models from the methodological perspective.
		
		\item Propagative control mechanisms are proposed to guide the  optimization  learning behaviors, which provides nontrivial theoretical analysis  that strictly guarantee the convergence properties for GDC in both fully- and partially-defined optimization model. Under the theoretical guarantee, we can construct diverse  combinations of propagative modules, which break down the limitations of specialized and fixed solution paradigms of existing methods.

		\item Our framework is flexible enough to introduce diverse augmentation strategies under the theoretical guarantee for consistent propagation, including normalization for stable parameters and joint training, and searching for  task/data adaptive architectures.

	\item On the application side, we demonstrate that GDC can obtain state-of-the-art results on a variety of low-level vision applications, such as non-blind deblurring, blind image deblurring,  interpolation, {smoothing} and rain removal, thus verifying the efficiency of our proposed framework. Moreover, GDC scheme can be expanded to feature-level  optimization tasks (\textit{e.g.}, optical flow).
	\end{itemize}

	\section{Related Works}
	\label{sec:related-work}

	\subsection{Classical Optimization Methods}
	Classical optimization methods often formulate the physical rules as particular kinds of energy models. Thus by optimizing the obtained models with different numerical schemes~\cite{li2015accelerated}, these methods can generate optimization iterations towards desired solutions. One prominent category of these iterative methods is the MAP-type schemes. Owing to the ill-posedness of most image-based tasks, MAP-type schemes have introduced explicit sparse priors to construct certain optimization formulations for regularizing the structures of latent solutions and characterizing task-oriented knowledge. Some ingenious priors such as dark channel prior~\cite{pan2016blind}, and extreme channel prior~\cite{yan2017image} are introduced to regularize the latent solutions space. Tensor Low-Cascaded Rank Minimization~\cite{sun2022tensor} mainly addresses the problem of hyperspectral low-level vision based on the subspace low-rank minimization technique.
	These carefully designed priors provide the mathematical principles for regularizing the propagations.
	Nevertheless, it brings difficulties to introduce priors that regularize the latent distribution space and characterize the generative formations completely for real-world tasks.  These limitations make the purely prior-regularized propagations difficult to obtain solutions efficiently and accurately.   
	\subsection{Learning-based Methods}
	Learning-based methods have achieved considerable advancements in various low-level vision tasks. Recently, many efforts have been put into designing versatile CNN-based architectures, based on the commonality of low-level vision tasks.
DualCNN~\cite{pan2022dual} proposes two parallel branches based on the structure and details of image features. DGUNet~\cite{Mou2022DGUNet} is proposed to parameterize the optimization step to investigate the intrinsic information of degradation.
However, similar to most learning-based methods, these approaches are highly dependent on the quantity and quality of training data and lack guidance and correction from optimization priors. Lately, Transformer-based models~\cite{zamir2022restormer,Wang_2022_CVPR} mitigates the limitations of CNN-based networks, which can capture long-range interactions for various low-level vision tasks.
Neural Architecture Search (NAS) methods~\cite{liu2018darts,liu2020retinex} pioneered the design of innovative paradigms to obtain task-specific architecture for vision tasks, which can effectively enhance the task/data adaptation ability.  However, without the explicit enforcement of physical rules, these solutions lack transparency and interpretability.

	\subsection{Combining Learnable Modules with Numerical Iteration}
	By unrolling the iterations of numerical optimization objectives, some methods attempt to utilize diverse principled modules as priors to plug into the numerical iterations. Recent studies introduce many plug-in schemes by unrolling-based techniques to deal with fidelity term and prior term respectively for generating the hybrid propagations. In~\cite{Chan2017Plug}, this method introduces the implicit denoiser modules as regularization to plug into numerical iterations. Very lately, pre-trained CNN-based modules~\cite{zhang2017learning,ryu2019plug,zhang2021plug} are implemented to solve many image restoration tasks such as image denoising, deblurring, and super-resolution, based on the optimization unrolling. However, these methods may lack mechanisms to control iteration to avoid local minimum. In recent years, optimization layer-based methods are proposed to introduce the optimization operators (\textit{\textit{e.g.,},} Total Variation solver~\cite{yeh2022total}, differentiable optimization~\cite{amos2017optnet, sun2022alternating}) as the building-blocks into the network design, which can provide effective  inductive bias. Deep implicit model~\cite{gould2021deep} is also introduced to define the network layer as the solution to an optimization problem implicitly. In the following, we will demonstrate GDC framework
		not only can be considered as the optimization scheme towards the desired solution but also plays a role as the deep network to improve performance by joint training.
Monga \textit{et.al}~\cite{monga2021algorithm} summarize the mainstream unrolled schemes and applications based on these underlying iterative algorithms. Compared with the existing advanced scheme, our framework has three major advantages.  Our framework is generalized enough to address fully/partially-defined models with sufficient theoretical guarantees.  Facing different tasks and demands, we can have diverse solving paradigms through the flexible combinations of propagative modules.  Supported by convergence guarantees, we introduce various augmentation techniques for performance improvement. 
	In the following, we will elaborate on the details of the proposed framework.

	\begin{figure*}[htb]
		\centering
		\includegraphics[width=0.98\textwidth]{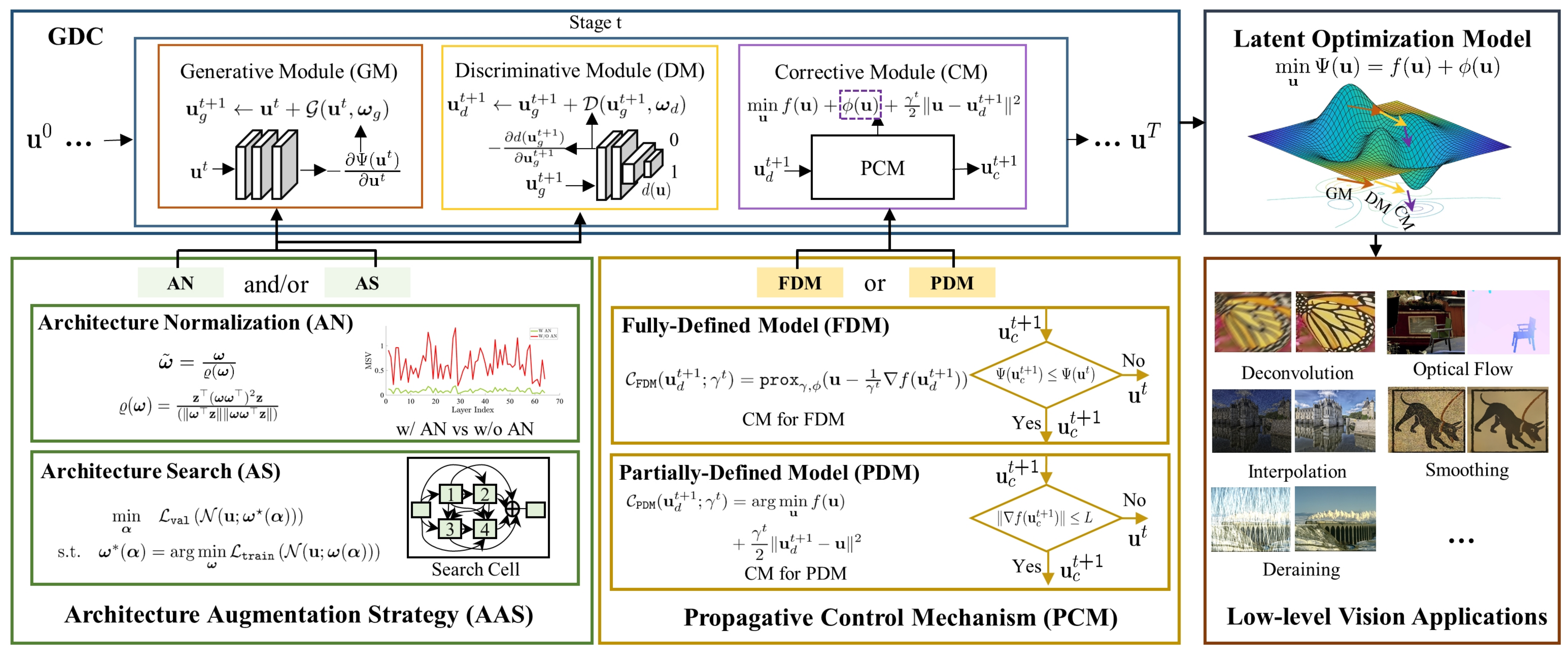}
		\caption{{Overview of the main workflow and components of GDC. We first illustrate GDC and the corresponding learning-based optimization trajectory on the top row. We then demonstrate the architecture augmentation strategies, the propagative control mechanisms and our considered low-level vision applications on the bottom row.}}
		
		\label{fig:illustration}
	\end{figure*}
	\section{The Proposed GDC Framework}
	
	In this section, we provide a general optimization learning framework to construct fundamental  modules for low-level vision tasks. In particular, we first abstractly formulate various low-level vision tasks using the following energy  model:
	\begin{equation}
	\min\limits_{\mathbf{u}}\Psi(\mathbf{u}):=f(\mathbf{u}) +  \phi(\mathbf{u}),\label{eq:model-u}
	\end{equation}
	where we denote $\mathbf{u}$ as our desired solutions (i.e, clear image), $f$ and $\phi$ are the data fidelity term for capturing the task principles (\textit{e.g.,}, the degradation formations) and the regularization term related to our prior on the solutions, respectively. 
	In this paper, we will consider two categories of models to formulate Eq.~\eqref{eq:model-u} for particular applications. Specifically, if we explicitly define $\Psi$ (\textit{i.e.,} $f$ and $\phi$) to formulate both the task information and the prior of our solution, we can obtain a determined form of Eq.~\eqref{eq:model-u} and call it Fully-Defined Model (FDM for short). On the other hand, in some challenging scenarios, we may only explicitly define $f$ to formulate the task but consider priors $\phi$ in some implicit manners. In this case, we will call Eq.~\eqref{eq:model-u} as a Partially-Defined Model (PDM for short).

	\subsection{Fundamental Propagation Modules}
	
	{In this work, we consider the optimization of Eq.~\eqref{eq:model-u} as an image/feature propagation process and formulate it by the universal descent-direction-based scheme, which iteratively estimates the descent direction (denoted as $\mathbf{d}^t$ at $t$-th stage) and performs $\mathbf{u}^{t+1}\leftarrow\mathbf{u}^t+\mathbf{d}^t$. However, rather than exactly calculating $\mathbf{d}^t$, which is  difficult or even impossible for complex $\Psi$, we would like to design propagation modules from different learning perspectives to investigate $\mathbf{d}^t$.  
		Specifically, here we design three fundamental modules as our task-originated decent directions, which are constructed from the generative learning perspective (\textit{i.e.,} Generative Module), based on the discriminative metric (\textit{i.e.,} Discriminative Module) and utilizing model information as corrections (\textit{i.e.,} Corrective Module).}

	\textbf{Generative Module (GM):} { We first predict the descent direction by learning it from the generative perspective. That is,
		by parameterizing $\mathbf{d}^t$ using the network architecture $\mathcal{G}(\mathbf{u}^t;\bm{\omega}_g)$, we can rewrite the updating scheme as
		\begin{equation}\label{eq:g_opt}
		\mathbf{u}^{t+1}_{g}\leftarrow\mathbf{u}^t+\mathcal{G}(\mathbf{u}^t;\bm{\omega}_g),
		\end{equation}
		where $\bm{\omega}_{g}$ denotes the network parameters. In this way, we  aim to obtain a module that can directly propagate $\mathbf{u}$ towards the optimal solution. In fact, GM can be understood as a residual network structure that has the ability to predict the task-desired solutions based on the collected training data.}

	\textbf{Discriminative Module (DM):}
	{We further construct another learning module, that aims to investigate the optimal solution from the binary classification perspective. That is, we would like to distinguish the optimal solution (of the considered task) from the corrupted observations based on a discriminative metric. Specifically, we introduce a binary classification network $d(\mathbf{u};\bm{\omega}_{d})$ with parameter $\bm{\omega}_{d}$, which can predict the task observation as positive (\textit{i.e.,} with label 1) and desired task-specific solution as negative (\textit{i.e.,} with label 0). Thus given the current state ${\mathbf{u}}^{t}$, we define the following optimization model to update $\mathbf{u}$:
		\begin{equation}
		\min\limits_{\mathbf{u}}\frac{1}{2}\|\mathbf{u}-{\mathbf{u}}^{t}\|^2 + d(\mathbf{u};\bm{\omega}_{d}), \label{eq:dm_opt}
		\end{equation}
		where $\|\cdot\|$ denotes $\ell_2$-norm. Then by calculating the gradient of Eq.~\eqref{eq:dm_opt} w.r.t. $\mathbf{u}$ and denoting the descent direction as $\mathcal{D}({\mathbf{u}};\bm{\omega}_{d}):=-\frac{\partial d({\mathbf{u}};\bm{\omega}_{d})}{\partial{\mathbf{u}}}$, we obtain 
		our discriminative metric based updating scheme (\textit{i.e.,} DM) as follows:
		\begin{equation}
		\mathbf{u}_d^{t+1}\leftarrow {\mathbf{u}}^{t} + \mathcal{D}({\mathbf{u}}^{t};\bm{\omega}_{d}).  \label{eq:dm_opt1}
		\end{equation}
		In fact, here we  learn a discriminative network architecture and calculate its gradient as our descent direction. We will also discuss the relation between our DM and the Discriminator in GAN in the following subsection.}
	
	\textbf{Corrective Module (CM):}
	{It should be noticed that both GM and DM are fully data-driven updating schemes, thus their generated propagation may lack task information (formulated by the original objective in Eq.~\eqref{eq:model-u}). Therefore, it is necessary to infuse the optimality condition of Eq.~\eqref{eq:model-u} to correct these fully data-driven propagation and promote the quality of solutions. In order to achieve this goal, we construct CM, a non-parametric module, from the numerical perspective. Specifically, we first consider the following penalized objective for Eq.~\eqref{eq:model-u}:
		\begin{equation}
		\min\limits_{\mathbf{u}}f(\mathbf{u}) +\phi(\mathbf{u})+ \frac{\gamma^t}{2}\|\mathbf{u}-{\mathbf{u}}^{t}\|^2, \label{eq:cm_opt}
		\end{equation}
		where $\gamma^t$ is a non-negative parameter for the penalty\footnote{Here we just follow standard optimization strategy to set $\gamma^{t+1} = \eta\gamma^{t}$ with $\eta>1$ during propagation.}. 
		Then we design CM based on Eq.~\eqref{eq:cm_opt} in both FDM (with explicit $f$ and $\phi$) and PDM (only with explicit $f$) scenarios.
		In detail, if Eq.~\eqref{eq:cm_opt} is fully-defined (\textit{i.e.,} FDM), we define a proximal gradient operation $\mathcal{P}_{\mathtt{FDM}}^{f+\phi}(\mathbf{u}^t;\gamma^t)$ on $\mathbf{u}^{t}$, \textit{i.e.,}
		\begin{equation}
		\mathcal{P}_{\mathtt{FDM}}^{f+\phi}(\mathbf{u}^t;\gamma^t)=\mathtt{prox}_{\gamma^t,\phi} \left({\mathbf{u}}-\frac{1}{\gamma^t}\nabla f(\mathbf{u}^{t})\right).
		\end{equation}
		While if we do not have explicit prior information $\phi$, the formulation of Eq.~\eqref{eq:cm_opt} can only be  partially-defined (\textit{i.e.,} PDM). We  consider the solution of the simplified penalization model as an operation $\mathcal{P}_{\mathtt{PDM}}^{f}$ on $\mathbf{u}^{t}$, \textit{i.e.,}
		\begin{equation}
		\mathcal{P}_{\mathtt{PDM}}^{f}(\mathbf{u}^t;\gamma^t)=\arg\min_{\mathbf{u}} f(\mathbf{u}) + \frac{\gamma^t}{2}\|\mathbf{u}- {\mathbf{u}}^{t}\|^2.
		\end{equation}
		Then the descent direction for CM can be uniformly formulated as 
		\begin{equation}
		\mathcal{C}(\mathbf{u}^{t})=\left\{\begin{array}{cc}
		\mathcal{P}_{\mathtt{FDM}}^{f+\phi}(\mathbf{u}^t;\gamma^t)-\mathbf{u}^t,&\mbox{for FDM},\\
		\mathcal{P}_{\mathtt{PDM}}^{f}(\mathbf{u}^t;\gamma^t)-\mathbf{u}^t,&\mbox{for PDM},
		\end{array}\right.
		\end{equation}
		and the corresponding propagation scheme is written as
		\begin{equation}
		\mathbf{u}_c^{t+1}\leftarrow {\mathbf{u}}^{t} + \mathcal{C}({\mathbf{u}}^{t}).
		\end{equation}}

	\begin{algorithm}
		\caption{PCM for Fully-Defined Model}\label{alg:correction1}
		\begin{algorithmic}[1]
			\REQUIRE The observation $\mathbf{y}$ and necessary parameters.
			\ENSURE Latent image estimation $\mathbf{u}^T$.	
			\STATE Initialization $\mathbf{u}^0 = \mathbf{y} $. 
			\FOR{$t=0,\dots,T-1$}
			\STATE   ${\mathbf{u}}^{t+1}_{g}\leftarrow\mathbf{u}^{t}+ \mathcal{G}(\mathbf{u}^{t};{\bm{\omega}}^{t}_g)$.\label{step:guide-g}
			\STATE   ${\mathbf{u}}^{t+1}_{d}\leftarrow{\mathbf{u}}^{t+1}_{g}+ \mathcal{D}({\mathbf{u}}^{t+1}_{g};{\bm{\omega}}^{t}_d)$.\label{step:guide-d}
			\STATE ${\mathbf{u}}^{t+1}_{c}\leftarrow {\mathbf{u}}^{t+1}_{d}+ \mathcal{C}({\mathbf{u}}^{t+1}_{d})$.
			\IF {$\Psi({\mathbf{u}}^{t+1}_{c}) \leq \Psi(\mathbf{u}^{t})$ }	\label{step:critreion-begin}
			\STATE $\mathbf{v}^{t} \leftarrow {\mathbf{u}}^{t+1}_{c}$.
			\ELSE
			\STATE $\mathbf{v}^{t} \leftarrow \mathbf{u}^t $.
			\ENDIF \label{step:criterion-end}
			\STATE $\mathbf{u}^{t+1} \leftarrow \mathtt{prox}_{\gamma^t,\phi} (\mathbf{v}^{t}- \frac{1}{\gamma^{t}} \nabla f(\mathbf{v}^{t})).$	\label{step:pg_u_new}
			\ENDFOR	
		\end{algorithmic}
	\end{algorithm}
	\section{Propagative Control Mechanisms for Theoretical Convergence}\label{sec:propagation}
	With the model information-based penalty formulation (\textit{i.e.,} Eq.\eqref{eq:cm_opt}),   CM can correct solutions at propagation for diverse optimization models (\textit{i.e.,} FDM and PDM) with domain knowledge. However, it is still challenging to obtain theoretical convergence guarantees for the above straightforward schemes, since the integration of deep modules breaks the rules of optimization iterations and leads to the failure of theoretical guarantee.
	Thus, as shown in Fig.~\ref{fig:illustration}, based on the correction of CM, we design two kinds of
	Propagative Control Mechanisms (\textit{i.e.,} PCM) as modularized techniques for the whole iterative process to guide the convergent propagation towards our desired solutions. We can obtain different convergence properties in both the FDM and PDM  scenarios with  PCMs.
	\subsection{PCM for Fully-Defined Model}
	Facing with the convergence of  general FDM, we propose the optimality checking  mechanism to control  output $\mathbf{u}^{t+1}_{c}$  based on the objective function's monotonicity.
	Obtaining $\Psi({\mathbf{u}}^{t+1}_{c})$ from CM, we choose the optimal objective value $\mathbf{v}^{t}$ through explicitly comparing $\Psi({\mathbf{u}}^{t})$ with $\Psi({\mathbf{u}}^{t+1}_{c})$.
	Then the final solution of $\mathbf{u}^{t+1}$ can be computed by optimizing  concrete FDM with  temporary variable $\mathbf{v}^{t}$, using  proximal-gradient scheme. 
	The whole   GDC scheme for FDM with PCM is summarized in Alg.~\ref{alg:correction1}.
	
	As PCM for  FDM,  we also assume that $f$ has the Lipschitz constant $L$, $\phi$ is a nonsmooth and nonconvex function with lower semi-continuity and $\Psi(\mathbf{x})$ is coercive. Exploiting with Proposition~\ref{prop:p4}, it can verify the sufficient descent property of objectives directly, which indicates Alg.~\ref{alg:correction1} can converge to the critical points. 
	\begin{prop}\label{prop:p4}
		(Critical Point Convergence of  PCM for FDM) 
		Let $\{\mathbf{u}^{t},\mathbf{v}^{t}\}$ be the sequences generated by Alg.~\ref{alg:correction1}. With PCM, we have the following conclusions: 
		(1) The objective function satisfies $\Psi(\mathbf{u}^{t})\geq\Psi(\mathbf{u}^{t+1})+\beta^t d_t$, where  $d_t$ denotes the distance between the output $\mathbf{u}^{t+1}$ and the selected $\mathbf{v}^{t}$ (with $\beta^t>0$).
		(2) Any accumulation points ($\mathbf{u}^{\ast}$) of $\{\mathbf{u}^{t}\}$ are the critical points of Eq.~\eqref{eq:model-u}, \textit{i.e.,} they satisfy $\mathbf{0}\in\partial \Psi(\mathbf{u}^{\ast})$. \footnote{Please notice that, the detailed proofs of all propositions are provided in Appendix.} 
	\end{prop}

	\subsection{{PCM for Partially-Defined Model}}
	
	Considering  complex low-level vision tasks, it is challenging to introduce  explicit priors to formulate the distribution (\textit{\textit{e.g.,},}, the complicated statistics of rain streaks). We also introduce another PCM to control the  propagation of PDM with explicit partially-defined fidelity and the output of CM (\textit{i.e.,} $\mathbf{u}_c^{t+1}$).
	We define the optimality checking condition as $\|\nabla f(\mathbf{u}_c^{t+1})\| \leq  L $. Specifically, targeting to the checking of latent solution $\mathbf{u}^{t+1}_{c}$,  we define  final output $\mathbf{u}^{t+1}$ as $\mathbf{u}^{t+1}\leftarrow{\mathbf{u}}^{t+1}_{c}$ if $\|\nabla f(\mathbf{u}_c^{t+1})\| \leq  L $ and $\mathbf{u}^{t+1}\leftarrow\mathbf{u}^{t}$ otherwise.  With   Proposition~\ref{prop:p5},  the fixed point convergence of this PCM for PDM can be guaranteed. Alg.~\ref{alg:correction3} summarizes the whole GDC with PCM scheme for PDM. In this way,  GDC is more flexible to extend  more complex low-level vision applications, which may have difficulty in  formulating the prior information explicitly.
	
	\begin{algorithm}[htb!]
		\caption{PCM for Partially-Defined Model}\label{alg:correction3}
		\begin{algorithmic}[1]
			\REQUIRE The observation $\mathbf{y}$ and necessary parameters.
			\ENSURE Latent image estimation $\mathbf{u}^T$.	
			\STATE Initialization $\mathbf{u}^0 = \mathbf{y} $.
			\FOR{$t=0,\dots,T-1$}
			\STATE   ${\mathbf{u}}^{t+1}_{g}\leftarrow\mathbf{u}^{t}+ \mathcal{G}({\mathbf{u}}^t; \bm{\omega}_{g}^t)$.\label{step:non-g}
			\STATE   ${\mathbf{u}}^{t+1}_{d}\leftarrow{\mathbf{u}}^{t+1}_{g}+  \mathcal{D}({\mathbf{u}}^{t+1}_{g};  \bm{\omega}_{d}^t)$.\label{step:non-d}
			\STATE ${\mathbf{u}}^{t+1}_{c}\leftarrow {\mathbf{u}}^{t+1}_{d}+ \mathcal{C}({\mathbf{u}}^{t+1}_{d})$.
			\IF  {$\|\nabla f(\mathbf{u}_c^{t+1})\| \leq  L $}	\label{step:critreion-1}
			\STATE $\mathbf{u}^{t+1}\leftarrow {\mathbf{u}}^{t+1}_{c}$.
			
			\ELSE
			\STATE $\mathbf{u}^{t+1}\leftarrow \mathbf{u}^t $.
			\ENDIF \label{step:criterion-2}
			
			\ENDFOR	
		\end{algorithmic}
	\end{algorithm}
	
	\begin{prop}
		\label{prop:p5}
		(Fixed Point Convergence of  PCM for PDM) 
		Let $\{\mathbf{u}^{t}\}$ to be the sequence generated by Alg.~\ref{alg:correction3} and $f$ be continuous differential with Lipschitz constant $L$. Denote $\mathcal{T}_{\mathcal{G}}$ and $\mathcal{T}_{\mathcal{D}}$ as the operators of GM and DM, respectively. Assume
		$\|\mathcal{T}_{\mathcal{D}} \!\circ\! \mathcal{T}_{\mathcal{G}}({\mathbf{u}^{t}}) - {\mathbf{u}^{t}} \| \leq   \sqrt{c/\gamma^{t}}$ ($c >0$) is satisfied for any  $\mathbf{u}^{t}$, then we have  the sequence $\{\mathbf{u}^t\}$ converges to a fixed-point.  
	\end{prop}

	
	
	
	\section{Architecture Augmentation Strategies for {Consistent Propagation}}
	To integrate the learnable modules seamlessly into the convergent optimization learning procedure without conflicts, we investigate two Architecture Augmentation Strategies (\textit{i.e.,} AAS) include Architecture Normalization (\textit{i.e.,} AN)  and Architecture Search (\textit{i.e.,} AS) to enhance the stability of learned propagation and discover  automated module architectures respectively. As shown in Fig.~\ref{fig:illustration},  AAS are performed for learning-based modules, where we use $\mathcal{N}$, $\bm{\omega}$ to denote these  modules and weights, \textit{i.e.,} $\mathcal{N}:=\{\mathcal{G },\mathcal{D}\}$ and $\bm{\omega}:=\{\bm{\omega}_g,\bm{\omega}_d\}$.
	
	\subsection{Architecture Normalization}
	In order to avoid  large oscillations generated by  manually designed architectures and enhance the stability of propagation, we propose architecture normalization, which investigates their contraction property of these architectures (\textit{i.e.,} $\mathcal{N}$). {The contraction property of  $\mathcal{N}$ indicates that
		$\|\mathcal{N}(\mathbf{u}_{1}) - \mathcal{N}(\mathbf{u}_{2})\|\leq \delta \|\mathbf{u}_{1} - \mathbf{u}_{2}\|$ with Lipschitz constant $\delta \textless 1$. In this way, if architecture $\mathcal{N}$ satisfies this property, the inequality $\|\mathcal{N}(\mathbf{u}^{t+1}) - \mathcal{N}(\mathbf{u}^{t})\| / \|\mathbf{u}^{t+1} - \mathbf{u}^{t}\|\leq \delta $ could be obtained to guarantee the stability of propagation.}

	{Indeed, above inequality can be converted as to control the largest singular values of $\mathcal{N}$. Since it is well known that its Lipschitz norm $\|\mathcal{N}\|_{\mathtt{Lip}}$ is just the largest singular value of weight matrix~\cite{SN}.}
	In  nutshell, architecture normalization controls the largest singular values of  weights  per layer to enforce $\|\mathcal{N}\|_{\mathtt{Lip}}$ from above 1.
	For instance, we define that $\mathcal{N}$ is a  network with $K$
	convolution layers.
	We normalize the weights ${\bm{\omega}}^{k}$ of the $k$-th layer to achieve
	\begin{equation}
	\max\limits_{\mathbf{u}^t \neq \mathbf{0}} {\|{\bm{\omega}}^{k}{\mathbf{u}}^{t}\|}/{\|{\mathbf{u}}^{t}\|} =\max\limits_{\|\mathbf{u}^t\|\leq 1} \|{\bm{\omega}}^{k} \mathbf{u}^t\| \leq (\delta)^{\frac{1}{K}},\label{eq:an1}
	\end{equation}
	where we define the constant $\delta \in (0,1]$  and  above equation implies that $\|\mathcal{N}\|_{\mathtt{Lip}}\leq \delta$.
	We utilize one matrix power accelerated iteration method~\cite{SN} with a random Guassian vector $\mathbf{z}$ to perform AN, \textit{i.e.,} 
	\begin{equation}
	\varrho({\bm{\omega}})  =   \mathbf{z}^{\top}(\bm{\omega}\bm{\omega}^{\top})^{2}\mathbf{z}~/(\|\bm{\omega}^{\top}\mathbf{z}\|\|\bm{\omega}\bm{\omega}^{\top}\mathbf{z}\|).
	\label{eq:sn}
	\end{equation}
	We leverage $\tilde{\bm{\omega}}=\bm{\omega}/\varrho({\bm{\omega}})$ to update the parameters of each layer in deep modules at the forward propagation. As shown in the AN part of Fig.~\ref{fig:illustration}, the largest singular value of $\mathcal{N}$ with (w/) AN is much smaller than
	$\mathcal{N}$ without (w/o) AN.
	
	We propose the architecture normalization strategy to normalize the weights per layer of data-driven modules for the stable propagation.  We  demonstrate that the architecture normalization guarantee the convergent propagation for the convex fully-defined objectives  (\textit{e.g.,}, image deblurring, image denoising and optical flow estimation) in Proposition~\ref{prop:p2}. Based on this, we can further improve performance by introducing joint training and other powerful augmentation strategies (\textit{e.g.,} architecture search) under convergence guarantees. 
	
	\begin{prop}\label{prop:p2}
		(Fixed-point Convergence with AN)  
		Let $\{\mathbf{u}^{t}\}$ to be the sequence generated by GDC with AN.
		Assume $f$ is $\rho$-strong convex with Lipschitz constant $L$, then the sequence $\{\mathbf{u}^{t}\}$ converges to a fixed-point if $(1+\delta_d)(1+\delta_g) < \frac{\rho+L}{|\rho -L|}$, where $\delta_g$ and $\delta_d$ denote the Lipschitz constants of GM and DM, respectively. 	
	\end{prop}
	\subsection{{Architecture Search}}

As aforementioned,  GM and DM in our GDC are CNN-based networks. However, the manual design of these deep modules is a challenging task, requiring extensive architecture engineering and subtle adjustments. Introducing existing networks directly into the iterations of numerical optimization is easy to induce inconsistency, causing propagation oscillations and inexact convergence. Our motivation is to better discover task/data-adaptive architectures, which can be seamlessly embedded in the optimization process and promote performance.
Inspired by the success of NAS in high-level vision field~\cite{liu2018darts}, we propose an architecture search strategy by introducing a compact search space and proposing a differentiable NAS strategy to discover one task/data-specific architecture N (\textit{i.e.,} GM or DM)  automatically for low-level vision.

	It is worth emphasizing that, we propose an effective task-oriented search space for low-level vision, instead of utilizing the primitive operators~\cite{liu2018darts} (\textit{e.g.,}, normal  $3\times3$ convolution  and max pooling). The search space is a set that contains the possible selections for designing task-specific $\mathcal{N}$, which includes
	
	\begin{table}[h]
		\begin{tabular}{c}
			\begin{minipage}{0.23\textwidth}  
				\begin{itemize} 
					\item\scriptsize$3\times3$ Dilated Convolution (3-DC)
					\item\scriptsize$3\times3$ Residual Blocks (3-RB)
					\item\scriptsize$3\times3$ Dense Blocks (3-DB)
					\item\scriptsize Spatial Attention (SA)
				\end{itemize}
			\end{minipage}
			\begin{minipage}{0.23\textwidth}
				\begin{itemize}
					\item\scriptsize$5\times5$ Dilated Convolution (5-DC)
					\item\scriptsize$5\times5$ Residual Blocks (5-RB)
					\item\scriptsize$5\times5$ Dense Blocks (5-DB)
					\item\scriptsize Channel Attention (CA)
				\end{itemize}
			\end{minipage} 
		\end{tabular}
	\end{table}
	The above operations have been demonstrated to deal with a series of low-level vision tasks effectively, which also guarantees the generalization and adaption ability for addressing diverse low-level vision tasks.
Then we consider a  differentiable search optimization strategy to perform AS under GDC scheme, which is
	\begin{equation}\begin{array}{ll}\label{eq:bilevel}
	\min \limits_{\bm{\alpha}} & \mathcal{L}_{\mathtt{val}}\left(\mathcal{N}(\mathbf{u};\bm{\omega}^{\star}(\bm{\alpha}))\right), \\
	\text {s.t.} & \bm{\omega}^{\star}(\bm{\alpha})=\arg\min \limits_{\bm{\omega}} \mathcal{L}_{\mathtt{train}}\left(\mathcal{N}(\mathbf{u};\bm{\omega}(\bm{\alpha}))\right),
	\end{array}\end{equation}
	where $\mathcal{L}_\mathtt{train}$ and $\mathcal{L}_\mathtt{val}$ denote  task-oriented losses on the training and validation dataset. $\bm{\alpha}$ represents the relaxation weights of architecture, following with~\cite{liu2018darts}.
	
	In detail, as shown in Fig.~\ref{fig:illustration}, we leverage the basic block (\textit{i.e.,} search cell) to simplify the search process of concrete architecture. The basic cell can be implemented as a directed acyclic graph, where the graph includes one node of input, 
a sequence of four inner nodes, and one node of the output. Compared with existing NAS schemes~\cite{liu2018darts, ye2022b}, we argue that architecture search is only one augmentation strategy in our paper. Under the convergence guarantee, more powerful augmentation strategies can be introduced to  improve performances for specific vision tasks.

	\subsection{Highlights and Discussions}
	In this subsection, we  highlight the distinct goals of GDC in different application scenarios.

It should be emphasized that there are two distinct goals of our GDC: either to exactly minimize the optimization model, or just to properly recover the task-required images/features. In particular, the distinction is that in some tasks (should be with FDM) our aim is to learn different propagative modules to find an exact minimizer of the model's objective function, thus the practicality is to speed up the convergence. In contrast, the optimization formulation can also be used to motivate our learning model only (with either FDM or PDM). In this case, our main goal is just to learn the propagation parameters so that we can obtain the desired solutions for the particular low-level vision task. Thus we do not require GDC to exactly minimize the optimization model. In summary, our GDC can be used as either \textbf{a learning-based optimizer} or \textbf{an optimization-derived learner} for applications in different scenarios.

	\section{Applications}\label{sec:applications}
	In this section, we demonstrate how to apply GDC to construct propagations for a series of low-level vision applications. In Table~\ref{tab:application}, we display  the clear energy formulations of considered low-level vision tasks.  We also summarize the abbreviations of essential notations in Table.~\ref{tab:notations}. 

	\subsection{GDC with Fully-Defined Model}
	FDM has both the fidelity $f$ and explicit prior $\phi$ in Eq.~\eqref{eq:model-u}, where  physical rules from $f$ and regularization from $\phi$ provide task-oriented domain knowledge for diverse tasks.  We apply GDC for four low-level vision applications (\textit{i.e.,} non-blind deblurring, blind deblurring, and interpretation) with 
	clear FDMs. Note that, for the particular task, we would like to  choose the prior term following existing works, thus guaranteeing relatively fair comparisons.
	\begin{table}[!]
		\centering
		\begin{threeparttable}
			\renewcommand{\arraystretch}{1.2}\footnotesize
			\caption{Summary of the optimization formulations for  low-level vision applications.}
			\label{tab:application}
					\vspace{-0.2cm}
			\setlength{\tabcolsep}{1.8mm}{
				\begin{tabular}{|c |c |c |c|}
					\hline
					Types  &   Tasks &  $f(\mathbf{u})$ &  $\phi(\mathbf{u})$ \\ \hline
					\multirow{4}{*}{FDM} & Non-blind deblurring & $\|\mathbf{u} \otimes \mathbf{k} - \mathbf{y}\|^2$& $\|\mathbf{B}^{T}\mathbf{u}\|_{1}$ \\
					& Blind Deblurring &   $\|\mathbf{u} \otimes \mathbf{k} - \mathbf{y}\|^2$ &  $\|\mathbf{B}^{T}\mathbf{u}\|_{0.8}$   \\
					&  Interpolation &  $\|\mathbf{u} \odot \mathbf{M} - {\mathbf{y}} \|^2$ & $\|\mathbf{B}^{T}\mathbf{u}\|_{0.8}$ \\
					& Smoothing  &  $\|\mathbf{u} - {\mathbf{y}} \|^2$  & $\|\mathbf{B}^{T}\mathbf{u}\|_{0}$  \\
					\hline
					
					PDM	& Rain Removal  &  $\|\mathbf{u} - {\mathbf{y}} \|^2$ & -- \\ \hline
				\end{tabular}
				\begin{tablenotes}
					\footnotesize
					\item $\mathbf{B}$ denotes the inverse wavelet transform and is utilized to compute the hyper-Laplacian prior~\cite{krishnan2009fast}.
				\end{tablenotes}
			}
		\end{threeparttable}
	\end{table}

		\begin{table}[!]
		\renewcommand{\arraystretch}{1.2}
		\caption{Summary of  essential notations.}
				\vspace{-0.2cm}
		\label{tab:notations}
		\centering\scriptsize
		\setlength{\tabcolsep}{1mm}{
			\begin{tabular}{|c| c| c| c|}
				\hline
				Notation   & Description & Notation   & Description \\
				\hline
				FDM/PDM &Fully/Partially-Defined Model & GM & Generative Model\\ \hline
				DM &Discriminative Model&CM&Corrective Module\\\hline
				PCM&Propagative Control Module&AA&Architecture Augmentation\\\hline
				AS&Architecture Search&AN&Architecture Normalization\\ \hline
				GC&GM+CM&GDC&GM+DM+CM\\ \hline
				
			\end{tabular}	
		}
	\end{table}
	
	\textbf{Non-blind Deblurring:} 
	The purpose of non-blind deblurring is to restore the latent images from  corrupted observations with known blur kernels. We formulate the specific case of Eq.~(\ref{eq:model-u}) for non-blind deblurring as $f(\mathbf{u})= \|\mathbf{u} \otimes \mathbf{k} - \mathbf{y}\|^2$ and $\phi(\mathbf{u})=\|\mathbf{B}^{T}\mathbf{u}\|_{1}$, where $\mathbf{u}$, $\mathbf{y}$, $\mathbf{k}$ and $\otimes$ are corresponding to the latent image, observation, blur kernel and convolution operator respectively.  According to related works~\cite{beck2009a}, we adopt the $\ell_1$-norm.
	For GM, we establish a residual network with seven convolution layers and six ReLU blocks, that are plugged behind each convolution layer. The DM is constructed as a standard CNN-based classifier~\cite{SRGAN} to promote the latent solutions. Particularly emphasized that we also use the same architectures to deal with the image restoration tasks.

	\textbf{Blind Image Deblurring:}
	Blind image deblurring includes estimating the blur kernel and recovering the sharp image from a blurred input. Here we demonstrate how to incorporate GDC into a kernel estimation process to nest the blind image deblurring process. We leverage the similar fidelity of non-blind deblurring to model  relations of image pairs in the gradient domain, where $\mathbf{u}$, $\mathbf{y}$ denote  gradients of solutions and corruptions in this application. 
The nonconvex hyper-Laplacian with $\ell_{0.8}$-norm is used as one regularization, aiming to model the distribution at the gradient domain better and preserve  clear structures for this challenging task. We train the GM with  noised images as inputs and the clear gradients (including the horizontal and  vertical gradients) as labels.
Then we just adopt  most normally utilized framework~\cite{pan2014deblurring} to estimate blur kernel $\mathbf{k}$ at $t$-th, \textit{i.e.,}
$
\mathbf{k}^{t+1}=\arg\min_{\mathbf{k}\in\Delta}\frac{1}{2}\|\mathbf{u}^{t+1}\otimes\mathbf{k}-\mathbf{y}\|^2 + \mu\|\mathbf{k}\|^2,\label{eq:solve-k}
$
where $\Delta=\{\mathbf{k}|\mathbf{1}^T\mathbf{k}=1,[\mathbf{k}]_i\geq 0\}$ is corresponding to the complex constraint, $\mu$ is a weighting parameter and $\mathbf{u}^{t+1}$ is the output of  $t$-th. We also introduce a classical coarse-to-fine strategy~\cite{sun2013edge,pan2014deblurring} for the kernel estimation. With the guidance of PCM,  convergent propagation can be guaranteed. 

\textbf{Image Interpolation:}
The main purpose of image interpolation is to partially restore invisible measurements, which are corrupted by some masks, such as pixel missing and text.
When using our paradigm to address these ill-posed tasks, the prominent difference with non-blind deblurring is fidelity term, which can be formulated as $f(\mathbf{u})= \|\mathbf{u} \odot \mathbf{M} - {\mathbf{y}} \|^2$, where $\odot$ represents the pixel multiplication and $\mathbf{M}$ denotes occluded corrections. We also utilize hyper-Laplacian with $\ell_{0.8}$-norm as the regularization.
	
	\textbf{Edge-preserved Smoothing:}
	Edge-preserved smoothing attempts to extract  dominant structures of observations and eliminate  undesired textures and noises. We construct the FDM formulation of this application by changing the fidelity formulation with the desired smooth solution $\mathbf{u}$ and the observation $\mathbf{y}$.  
		As for this application, we establish the fidelity as $f(\mathbf{u}) =  \|\mathbf{u} - \mathbf{y} \|^2$ and build widely used $\phi$ as $\ell_{0}$-norm prior~\cite{xu2011image} to smooth the textures. 
	\subsection{GDC with Partially-Defined Model }
	PDM only observes the fidelity $f$ explicitly but overlooks the information on  prior $\phi$, since it remains challenging to introduce proper and effective priors to model precise distribution for complex vision tasks. Rain removal is a representative task due to  the difficulty of introducing   priors to formulate the complex statistics of rain streaks.

		\textbf{Rain Removal:}
		Recovering the rain-degraded images is a challenging and  crucial task for the outdoor visual scenario, aiming to remove  rain streaks and recover  details simultaneously.
		We leverage  PDM to characterize the rain removal task based on  Alg.~\ref{alg:correction3}. 
	Considering  rain streaks can be regarded as a category of signal-dependent noises, we define the fidelity term as $\|\mathbf{u} - \mathbf{y}\|^{2}$, aiming to persevere the structure of latent images. 	
	Furthermore, due to existing NAS-based schemes~\cite{gou2020clearer,zhang2020memory} mainly focus on image deraining, we leverage AS to construct GDC scheme, and we perform AS to
	discover the concrete GM and DM for rain removal respectively. GM consists of four search cells and DM contains two  cells.

	\subsection{Training Strategy}
	The most straightforward way to establish our propagative schemes is just to cascade  three kinds of principled modules hierarchically in practice.
	Specifically, as for  image restoration and blind deblurring, we first leverage the task-based fidelity ${\mathbf{u}} = \arg\min_{\mathbf{u}}  f(\mathbf{u}) + \frac{\gamma^t}{2}\|\mathbf{u}-{\mathbf{u}^{t}}\|^2$ to obtain an initial solution based on the closed-form solution. Then we construct a GM to learn the dependencies between  closed-form outputs   and task-specific solutions, aiming to avoid  complicated training phrases.  Finally, we cascade  three primary proposed modules repeatedly to construct our propagation to deduce the procedure of estimating the desired outputs progressively.

	We  can perform two different training strategies (\textit{i.e.,} greedy and joint) or their combination to optimize the weight parameters in GM and DM. 
	The greedy training strategies~\cite{zhang2017learning} have been widely exploited  for low-level vision, aiming to train each module separately and incorporate them into our GDC optimization. 
	With such separate training strategies, we can obtain convergence guarantees for GDC with the following control mechanisms.
	On the other hand, the overall scheme after the direct cascade
	can be considered a deep network.
	Adopting the  joint training with task-oriented loss functions, we can optimize the parametric modules across stages jointly to preserve the consistency of architecture in an end-to-end manner. 
	
	\subsection{Implementation Details}
	As for greedy training, we can gradually train our deep modules to refine the output of each stage progressively. Within this training strategy, we sampled images from BSD benchmark~\cite{martin2001database} and added Gaussian noise with small interval levels with a standard deviation of 0.1  to construct the training pairs, following with~\cite{zhang2017learning}. Moreover, we trained GM using these noised-clear training pairs, supervised by the MSE loss. As for  DM, we labeled the noised images as one and the desired solution as zero and leveraged the classification criterion to train our DM. We leverage the Adam optimizer with a learning rate of $1e^{-3}$ and a multi-step learning rate decay strategy to train these modules. As
for joint training, we additionally utilize data augmentation (\textit{e.g.,}, flip and rotation) with the task-specific dataset to train the whole network until convergence.
	
	\section{Ablation Study}
	\label{sec:alb}
	In this section, we  investigate the properties of these propagative modules and the efficiency of PCM and AAS on  non-blind deblurring tasks.

\begin{figure}
		\centering \begin{tabular}{c@{\extracolsep{0.2em}}c@{\extracolsep{0.2em}}c@{\extracolsep{0.2em}}c}
			\includegraphics[width=0.112\textwidth,height=0.08\textheight]{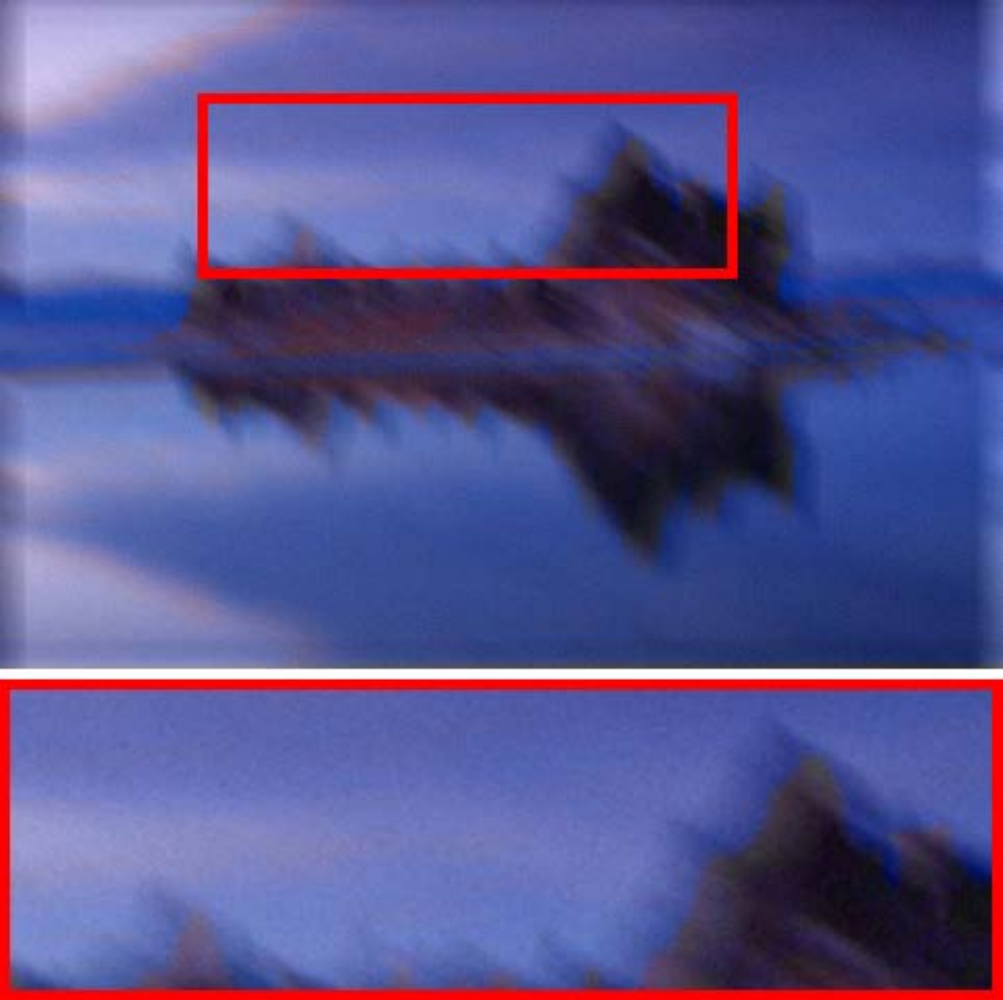}
			&\includegraphics[width=0.112\textwidth,height=0.08\textheight]{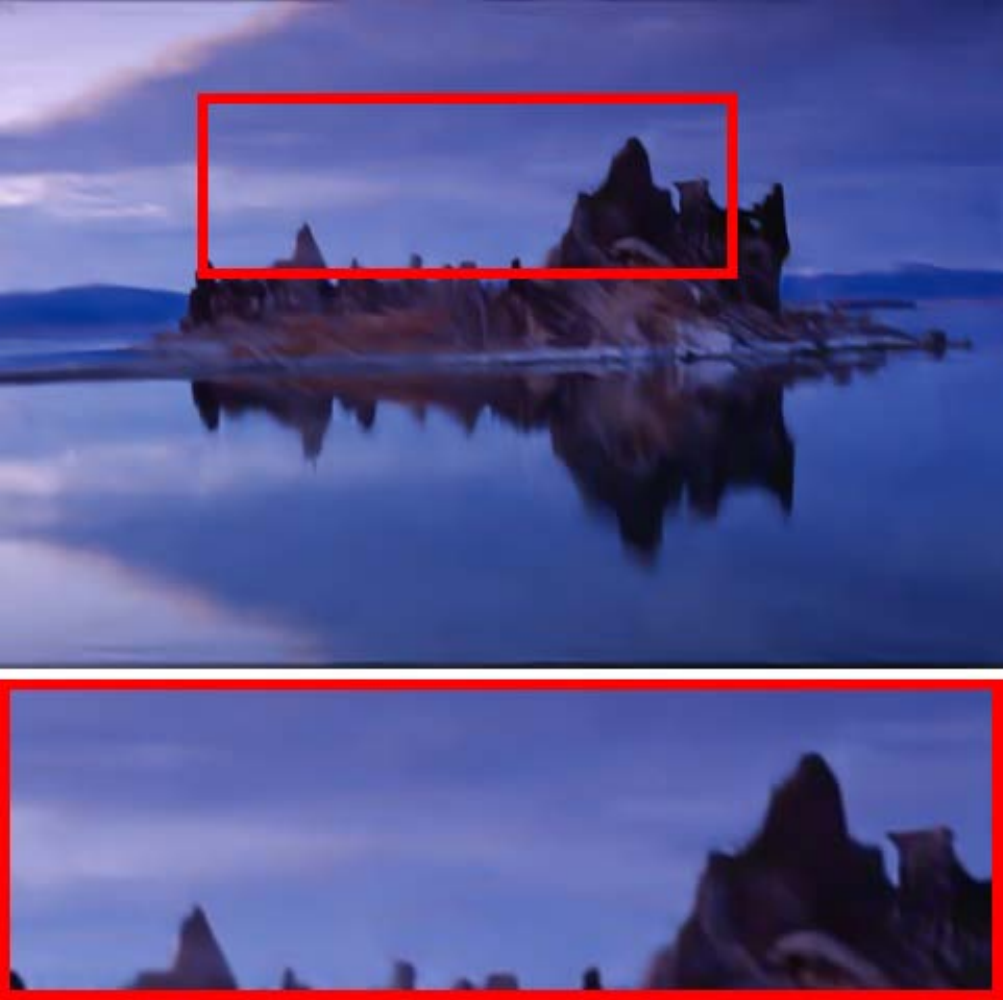}
			&\includegraphics[width=0.112\textwidth,height=0.08\textheight]{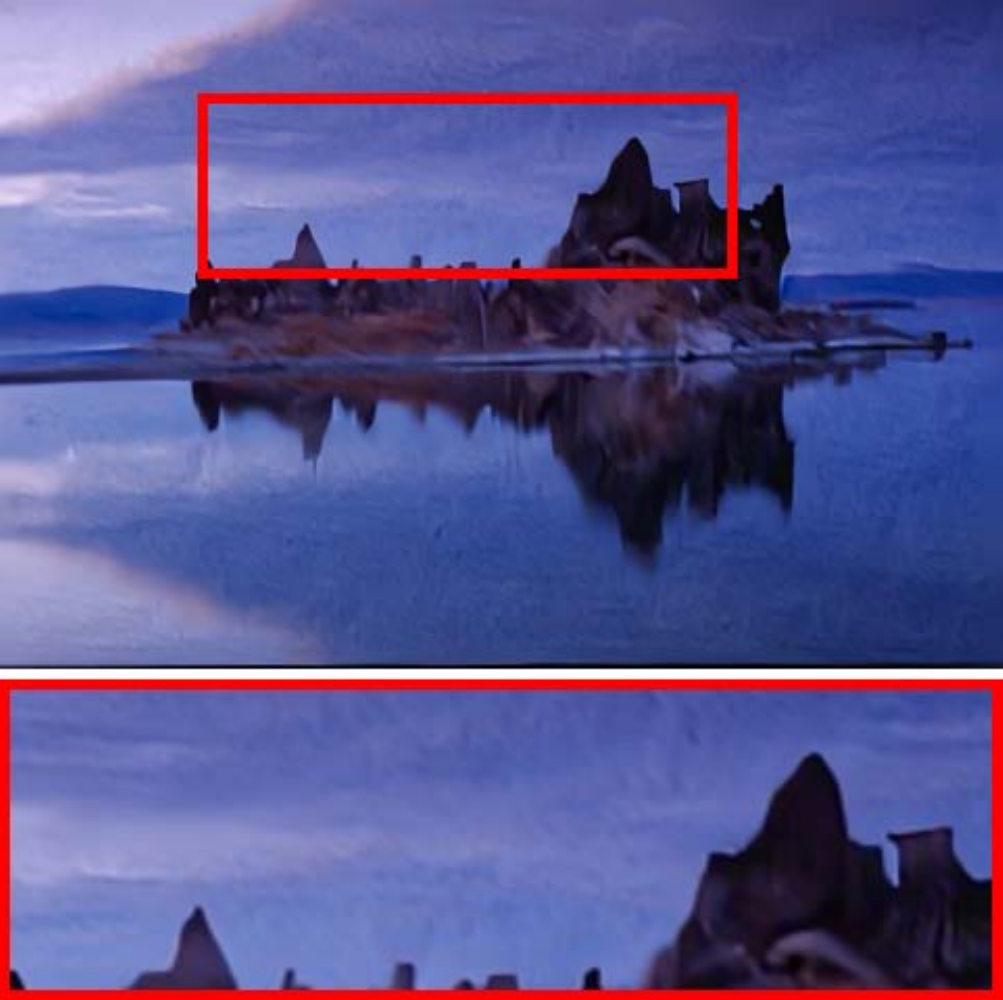}
			&\includegraphics[width=0.112\textwidth,height=0.08\textheight]{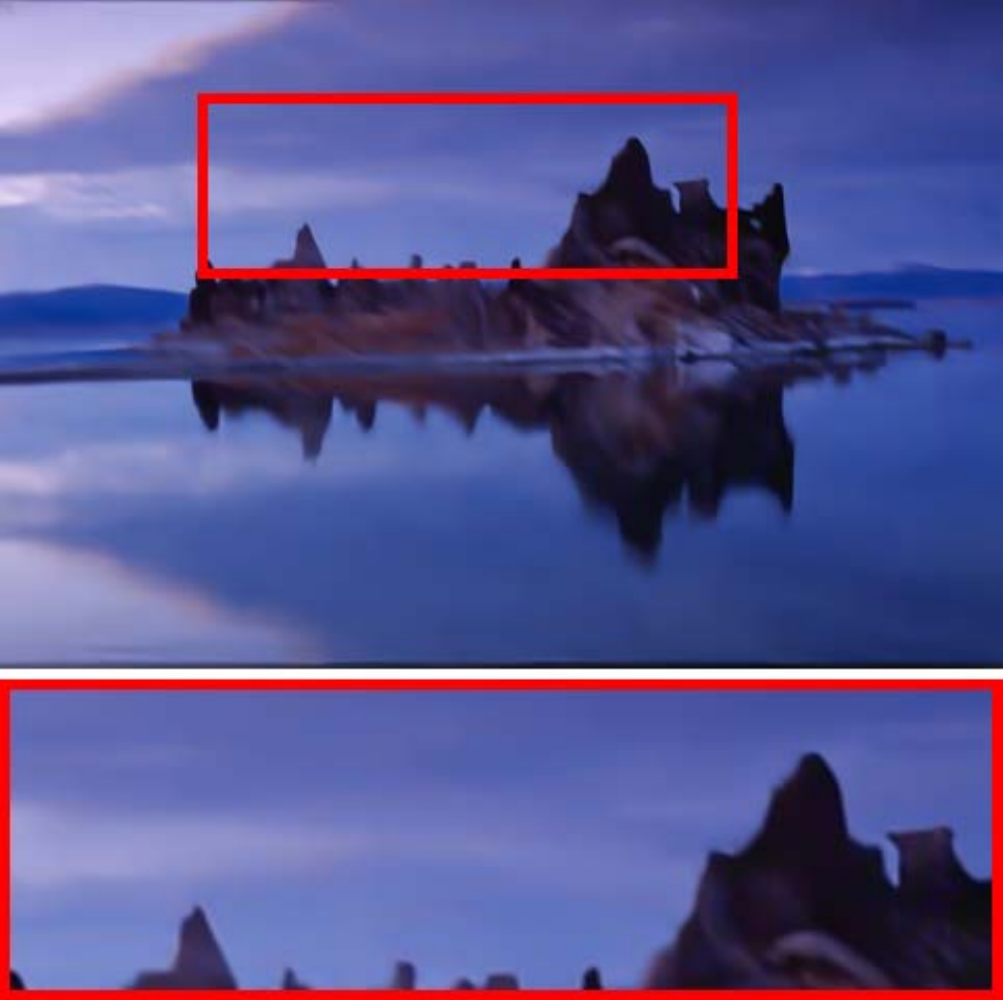}\\
			\footnotesize	(a) Input & \footnotesize(b) GD & \footnotesize(c) GC &\footnotesize (d) GDC \\
		\end{tabular}
		\caption{The visual results of series schemes  under  3\%  noise level. }
		\label{fig:alb1}
	\end{figure}

	\textbf{Foundational Modules Evaluation:}
	In this part, the properties of each basic module are carefully discussed. In  Table~\ref{tab:correct}, we reported the experimental results of different cascade schemes (\textit{i.e.,} GM, GC, GD, and GDC) under  diverse Gaussian noise levels. $\sigma$ denotes the percent form of the Gaussian noise level. GM obtained  considerable performances and GC achieved  better quantitative results under the disturbance of small noise particularly. 
Although GDC obtained  similar performances with GC, the visual result of Fig.~\ref{fig:alb1} showed that  GDC had fewer artifacts than GC.
	\begin{table}[!]
		\renewcommand{\arraystretch}{1.2}
		\caption{Averaged PSNR  on Levin \emph{et al.}~\cite{levin2009understanding} and BSD68~\cite{roth2009fields}.}
		\label{tab:correct}
				\vspace{-0.2cm}
		\centering\footnotesize
		\setlength{\tabcolsep}{2.5mm}{
			\begin{tabular}{|c |c| c| c| c|}
				\hline
				$\sigma$  & GM &GD& GC&  GDC\\
				\hline
				1\%&34.81/29.51&34.89/29.79&34.86/29.80&34.90/29.79 \\ \hline
				2\%&32.12/27.84&32.26/27.91&32.41/27.85&32.36/28.00 \\ \hline
				3\%&30.35/26.88&30.51/26.83&30.52/26.83&30.69/26.97 \\ \hline
				
			\end{tabular}	
			
		}
	\end{table}
	
	\begin{figure}
		\centering \begin{tabular}{c@{\extracolsep{0.2em}}c@{\extracolsep{0.2em}}c@{\extracolsep{0.2em}}c}
			
			\includegraphics[width=0.112\textwidth,height=0.08\textheight]{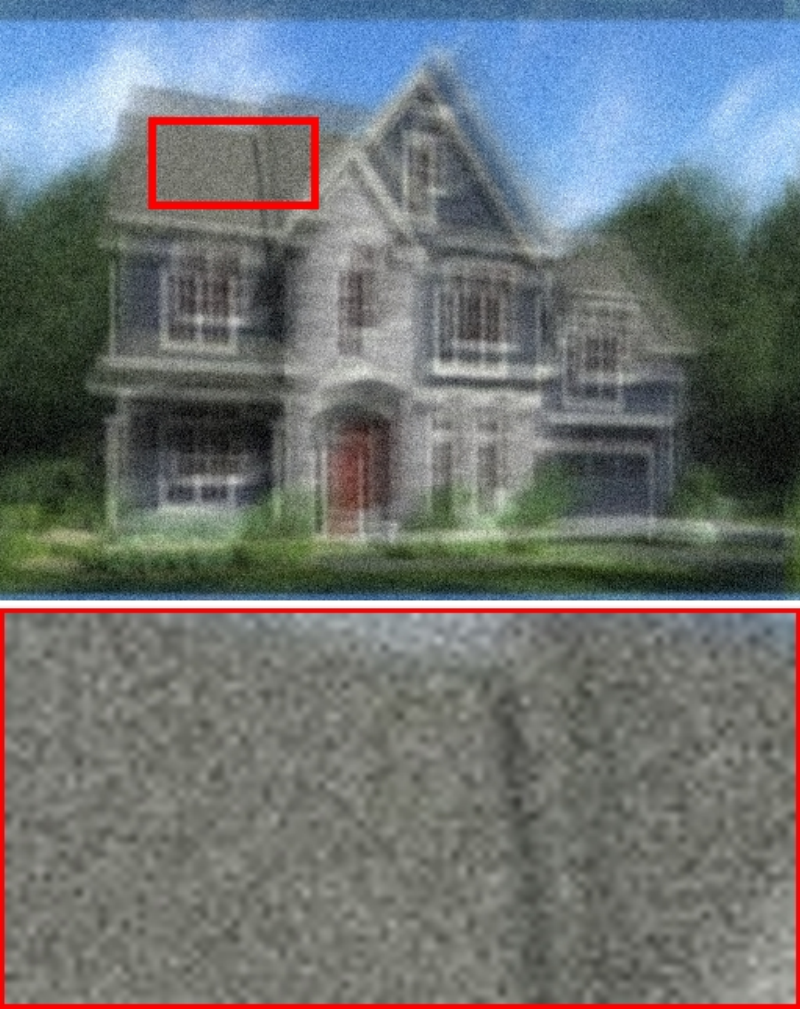}
			&	\includegraphics[width=0.112\textwidth,height=0.08\textheight]{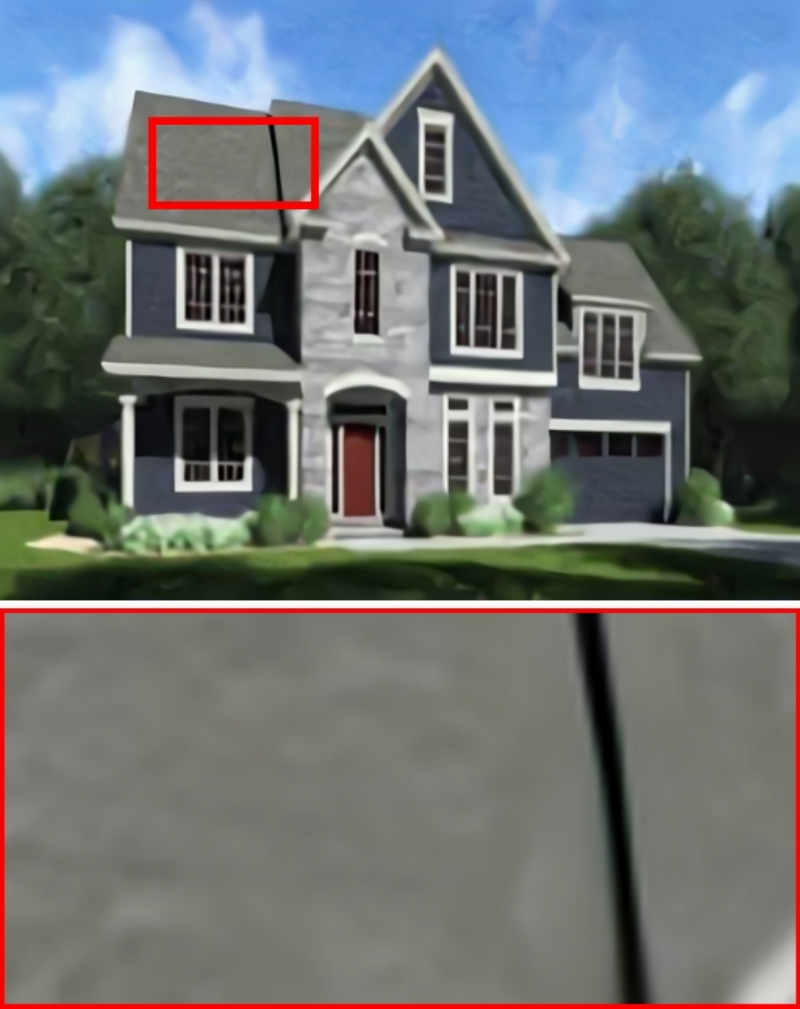}
			&\includegraphics[width=0.112\textwidth,height=0.08\textheight] {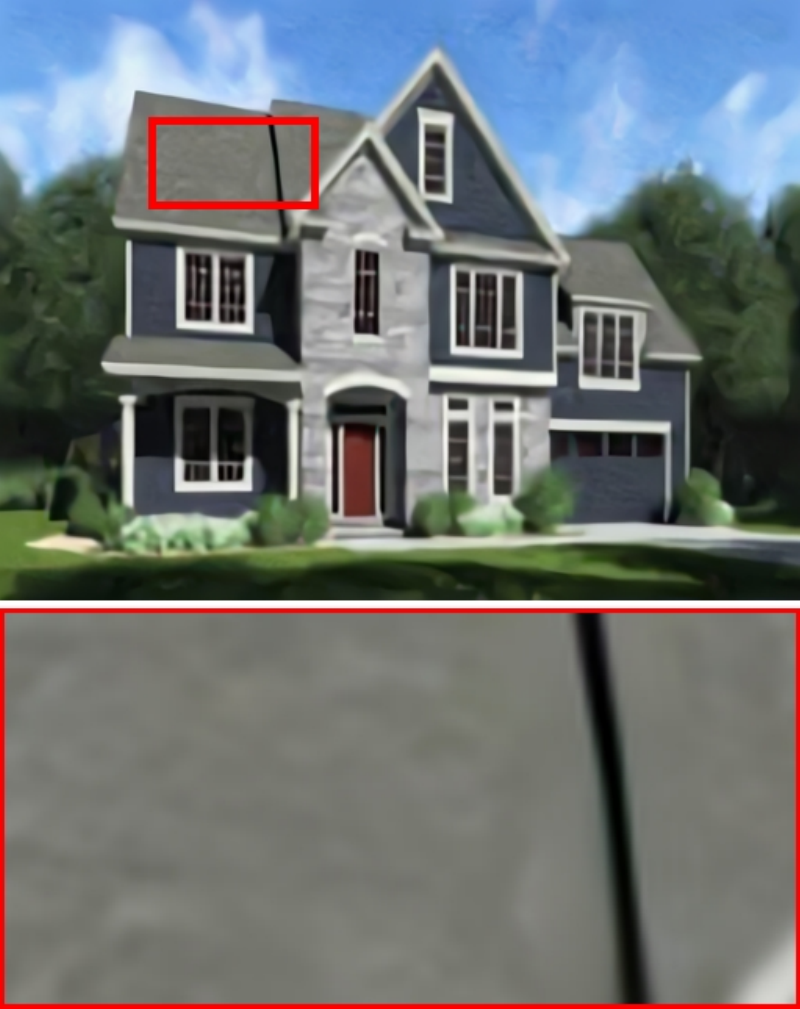}
			&\includegraphics[width=0.112\textwidth,height=0.08\textheight] {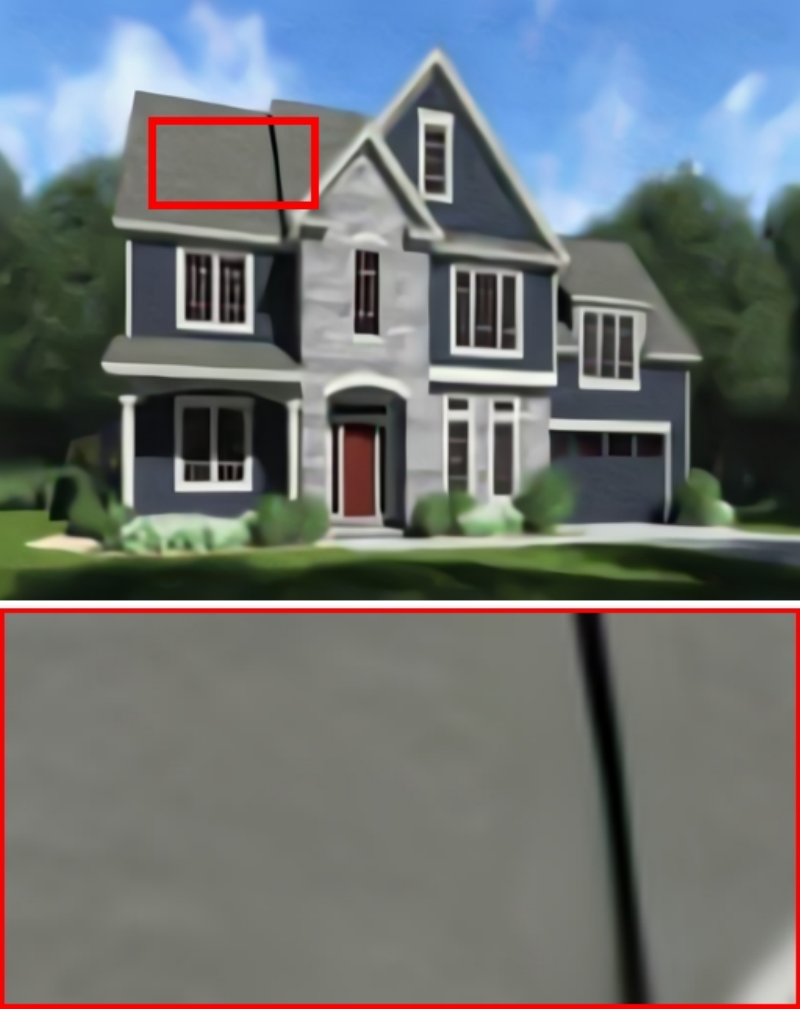}	\\
			\footnotesize	(a) Input &\footnotesize (b) {GM}$_\mathtt{Joint}$& \footnotesize(c) ({G}{C})$_\mathtt{Joint}$ &\footnotesize(d) ({G}D{C})$_\mathtt{Joint}$ \\
			
		\end{tabular}
		\caption{The visual results of joint training under 5\%  noise level. 
		}
		\label{fig:albstudy}
	\end{figure}
	We also illustrated the performances of  joint training for these schemes, denoted this strategy as ``$\mathtt{Joint}$'', where we reported quantitative results in Table~\ref{tab:big}. 
All these schemes with joint training strategy improved the performances greatly compared with  greedy training. 
We can see that (GDC)$_\mathtt{Joint}$ has better performances under the strong noise.
(GC)$_\mathtt{Joint}$ obtained the satisfied result under a small noise level. On the other hand,  visual results of different networks with  joint training are shown in Fig.~\ref{fig:albstudy}. (GDC)$_\mathtt{Joint}$ obtained 
more visually  favorable results, especially under  strong noises (5\%).
	
	\begin{table}[!]
		\renewcommand{\arraystretch}{1.2}
		\caption{The numerical results of joint training on Levin \emph{et al.}~\cite{levin2009understanding} and BSD68~\cite{roth2009fields}.}
		\label{tab:big}
				\vspace{-0.2cm}
		\centering\footnotesize
		\setlength{\tabcolsep}{4.5mm}{
			\begin{tabular}{|c| c| c| c|}
				\hline
				$\sigma$    & {GM}$_\mathtt{Joint}$& ({G}{C})$_\mathtt{Joint}$ & ({G}D{C})$_\mathtt{Joint}$\\
				\hline
				1\%&35.45/30.07&35.57/30.11&35.51/30.09 \\ \hline
				2\%&32.57/28.28&32.73/28.30&32.75/28.25\\\hline
				3\%&30.86/27.27&31.08/27.28&31.06/27.15\\\hline
				5\%&28.76/25.30&28.86/25.92&28.94/26.06\\ \hline
				10\%&26.04/24.35&26.05/24.43&26.25/24.57\\ \hline
				
			\end{tabular}	
		}
	\end{table}

\begin{figure}[htb!]
		\centering
		\begin{tabular}{c@{\extracolsep{0.1em}}c@{\extracolsep{0.1em}}c}
			\includegraphics[width=0.23\textwidth]{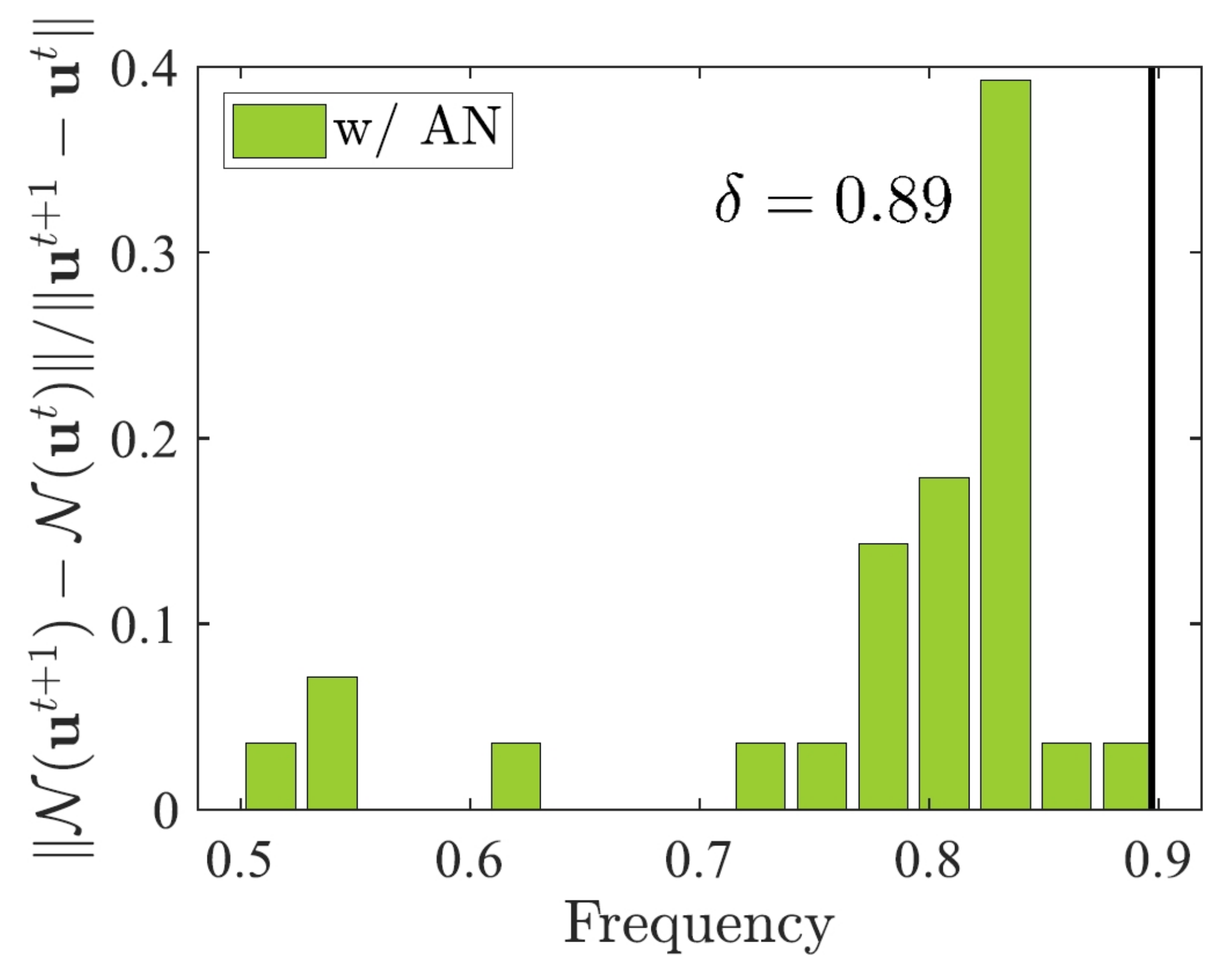}&
			\includegraphics[width=0.23\textwidth]{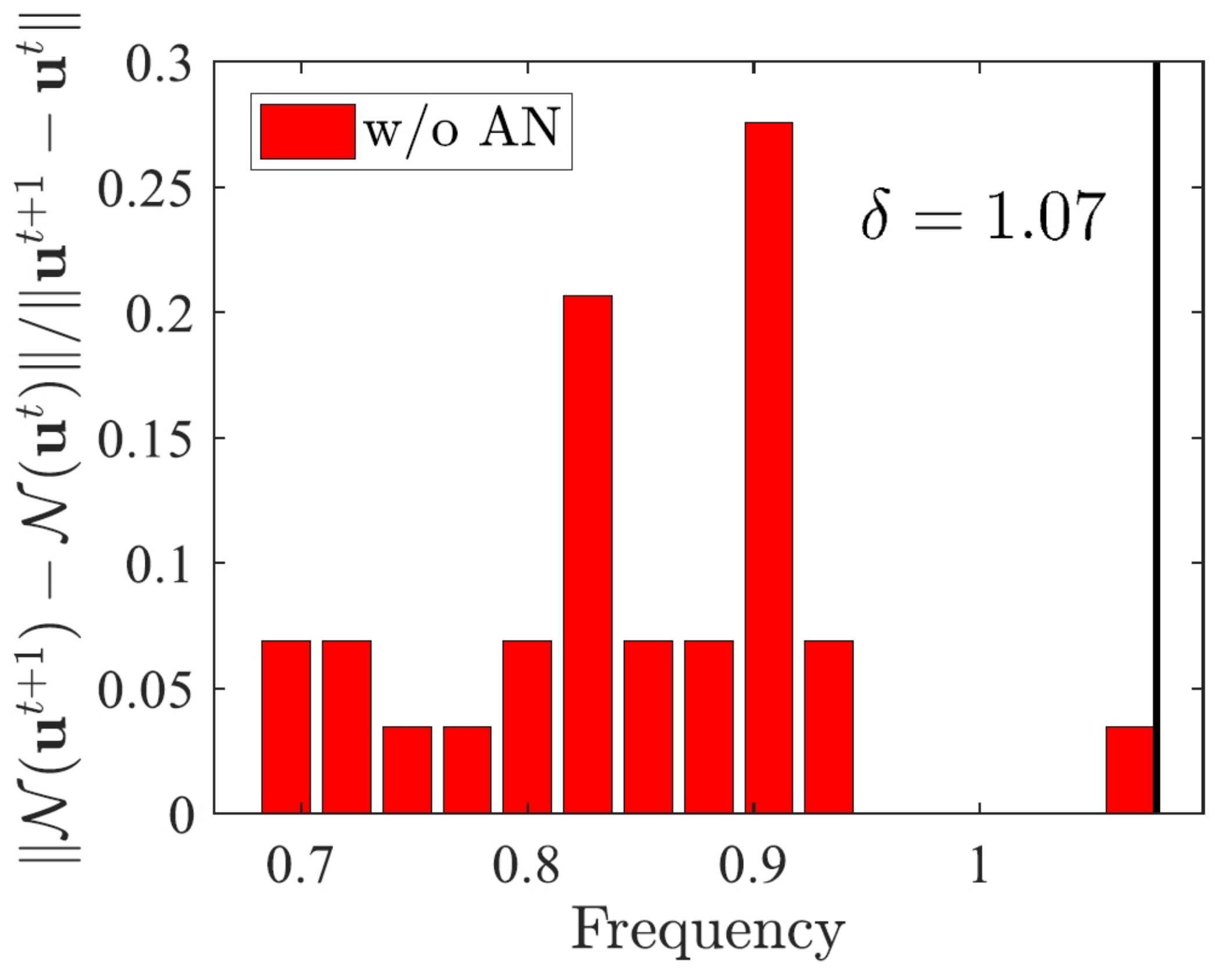}\\
			\footnotesize	(a) GM w/ AN  &\footnotesize (b) GM w/o AN\\
		\end{tabular}
		\caption{{The frequency statistics of $\frac{\|\mathcal{N}(\mathbf{u}^{t+1} -\mathbf{u}^{t} )\|} { \|\mathbf{u}^{t+1} - \mathbf{u}^{t} \|}$  for GM  w/  and w/o AN.}}
		\label{fig:snn}
	\end{figure}

	\textbf{AAS Evaluation:}
	As for AN, we adopted the GC scheme to verify the efficiency and compared the properties of our architectures $\mathcal{N}$ (\textit{i.e.,} GM in this scenario) with (w/)  and without (w/o) AN in Fig.~\ref{fig:snn}. The subfigure (a) and  subfigure (b)  plot the frequency of $\|\mathcal{N}(\mathbf{u}^{t+1}) -\mathcal{N}(\mathbf{u}^{t} )\| / \|\mathbf{u}^{t+1} - \mathbf{u}^{t} \|$, where the ratio for GM w/ AN is much smaller than GM w/o AN. Thereby, these  results justified the effectiveness of AN.

\begin{figure}[htb]
		\centering \begin{tabular}{c@{\extracolsep{0.1em}}c}							     
			\includegraphics[width=0.23\textwidth]{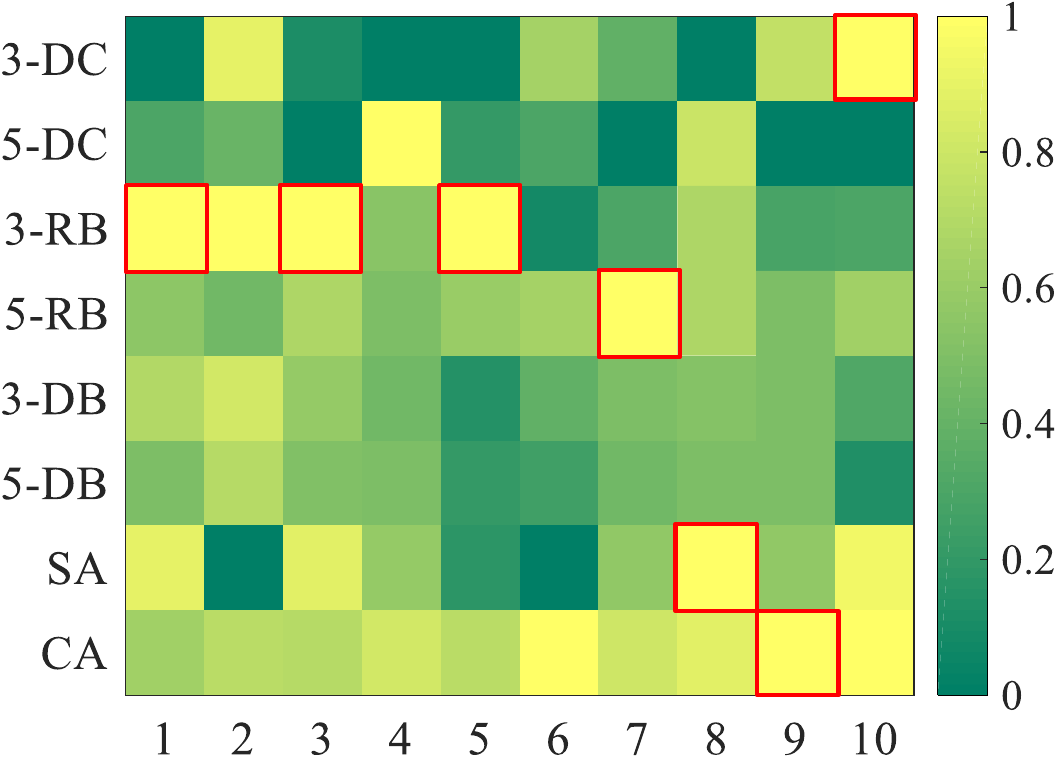}
			&\includegraphics[width=0.23\textwidth]{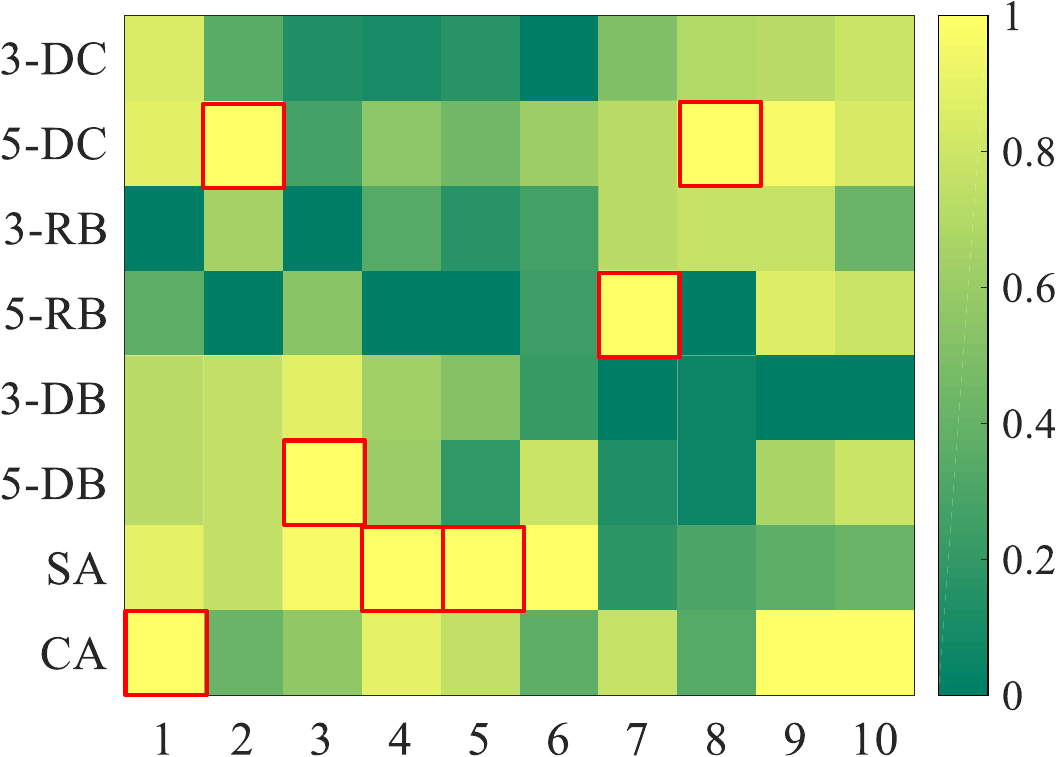}\\
			\footnotesize	(a) Searched GM & \footnotesize (b)  Searched  DM
		\end{tabular}
		\caption{{Heatmaps about  basic candidate operators of GM and DM  at  the last search epoch. The final  architectures are indicated by  red boxes.} }
		\label{fig:nas}
	\end{figure}

To validate the quantitative performance of GDC, we randomly select operators from the search space to construct GM and DM. To be specific, As for GM, we adopt the  successive architecture to propagate the feature with $3\times 3$ residual block, $3\times 3$ dense block, $3\times 3$ dilated convolution and attentions progressively. As for the DM, we introduce the pre-trained DM (with a similar architecture for deblurring). This scheme is denoted as  GDC$_\mathtt{MD}$.
We also analyzed the final AS-based architectures for the rain removal task.  The Fig.~\ref{fig:nas} shows the final searched  architectures of GM and DM respectively.  GM contains more residual blocks and dilated convolution with  small receptive fields, which is  reasonable and demonstrated in previous manually designed works.  DM consists of many dilated convolutions with large receptive fields, tending to capture more  contextual information of  observations. In this way, these results indicated the significance of AS in discovering task-specific architectures. We denote the AS-based GDC as  GDC$_\mathtt{AS}$.
In Table.~\ref{tab:alb_rain}, we compared the AS performance for image deraining. Obviously,  GDC$_\mathtt{AS}$ achieves the best numerical performance.
	
	\begin{table}	\renewcommand{\arraystretch}{1.2}
		\caption{{Quantitative results about rain removal on Rain800~\cite{li2018recurrent}.}}
		\label{tab:alb_rain}
				\vspace{-0.2cm}
		\centering\footnotesize
		\setlength{\tabcolsep}{1.8mm}{
			\begin{tabular}{|c| c| c| c| c|}
				\hline
				Metrics&	 HiNAS~\cite{zhang2020memory}  & CLEARER~\cite{gou2020clearer} & GDC$_\mathtt{MD}$ &  GDC$_\mathtt{AS}$  \\ \hline		
				PSNR &26.31 &27.23 & 27.65& \textbf{ 28.14} \\ \hline
				SSIM & 	 0.86 &0.86 & 0.87 & \textbf{0.88} \\ 
				\hline
			\end{tabular}	
		}
	\end{table}

	\begin{table}[!]
		\renewcommand{\arraystretch}{1.1}
		\caption{Quantitative results~(PSNR/SSIM) about  non-blind deblurring compared with state-of-the-art methods.}
		\label{tab:ablationtabless}
				\vspace{-0.2cm}
		\centering\footnotesize
		\setlength{\tabcolsep}{0.8mm}{
			\begin{tabular}{ |c | c| c| c| c| c| c| c| c| c|}
				\hline
				Datasets&Metrics    &~\cite{kruse2017learning}  &	 ~\cite{zhang2017learning}  & ~\cite{eboli2020end}&~\cite{nan2020deep}  &  ~\cite{zhang2021plug}   &  ~\cite{chen2022nonblind} & ~\cite{sanghvi2022photon} &	Ours \\ \hline
				\multirow{2}*{Levin}&PSNR  &33.59 & 34.61 &20.06 & 34.22& 34.85& 27.26& 26.62& \textbf{35.57}   \\ 
				&SSIM    &0.93  & 0.93 &0.65 & 0.94&	0.94& 0.90 &0.78 & \textbf{0.95} \\\hline
				\multirow{2}*{BSD68}&PSNR   &29.49 & 29.83 & 19.64& 28.87& 29.87 & 24.85 & 24.38& \textbf{30.11}  \\ 
				&SSIM  &0.89   & 0.85 &0.59 &0.83 &	0.85  & 0.77 &0.63   & \textbf{0.86} \\
				\hline
				
			\end{tabular}	
		}
	\end{table}
	
	\begin{figure}   
		\centering \begin{tabular}{c@{\extracolsep{0.1em}}c}							     
			\includegraphics[width=0.23\textwidth]{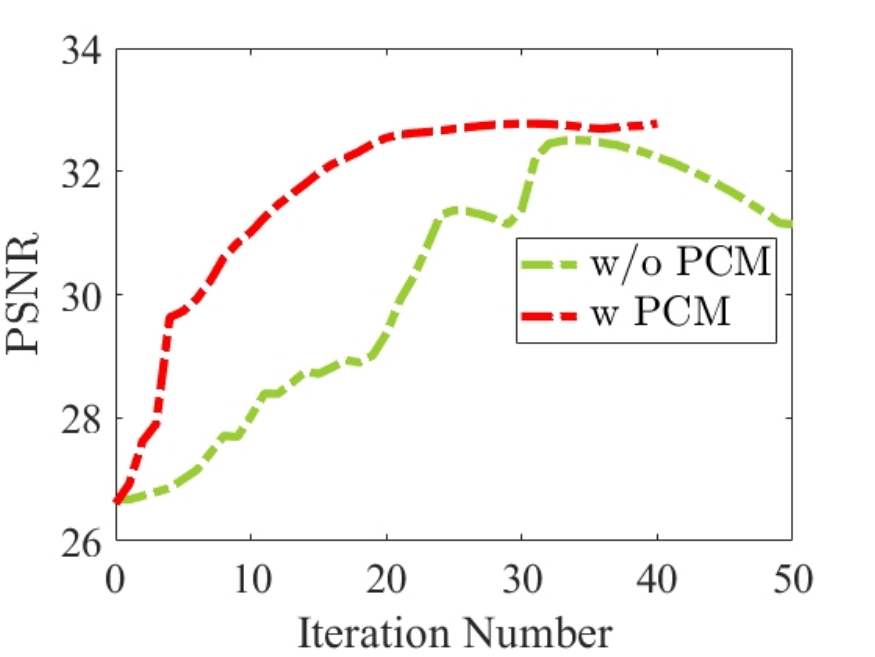}
			&\includegraphics[width=0.23\textwidth]{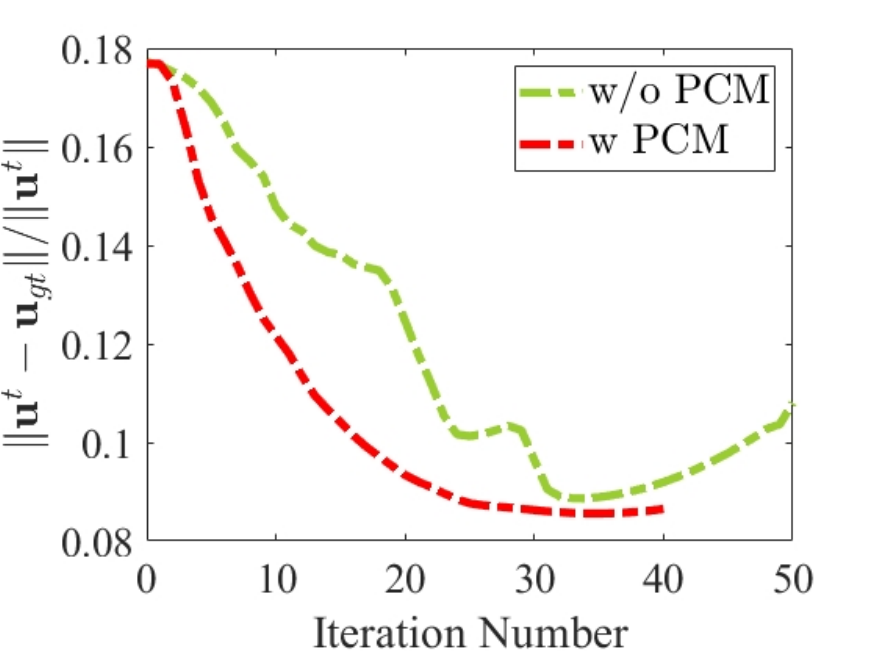}\\
			\footnotesize	(a)   PSNR  &\footnotesize (b)  Reconstruction Error
		\end{tabular}
		\caption{Convergence analysis for GDC w/ and w/o PCM.
		} 
		\label{fig:sns}
	\end{figure}
	
	\begin{figure*}
		\centering \begin{tabular}{c@{\extracolsep{0.2em}}c@{\extracolsep{0.2em}}c@{\extracolsep{0.2em}}c@{\extracolsep{0.2em}}c@{\extracolsep{0.2em}}c@{\extracolsep{0.2em}}c}
			
			\includegraphics[width=0.135\textwidth,height=0.09\textheight]{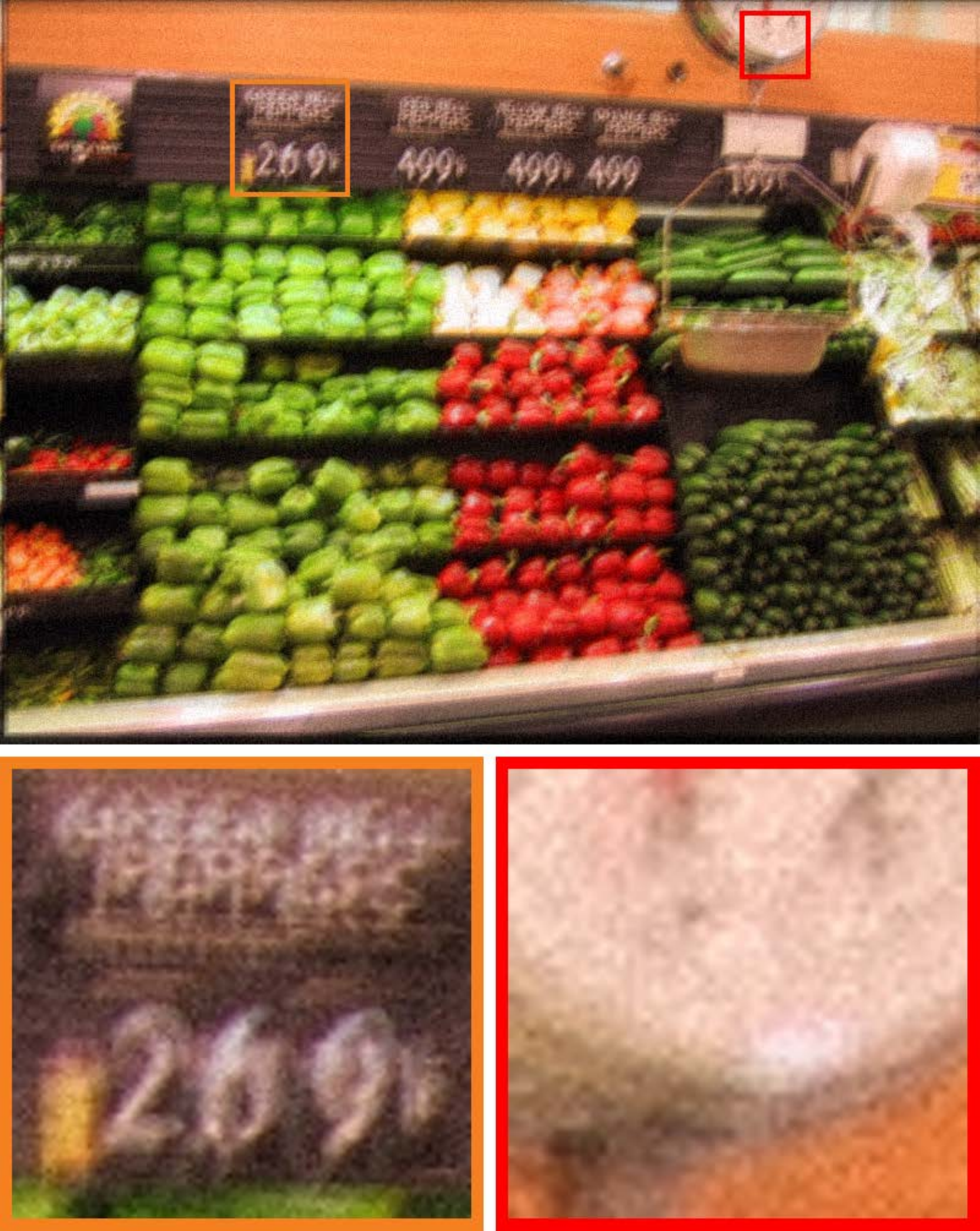}
			&\includegraphics[width=0.135\textwidth,height=0.09\textheight]{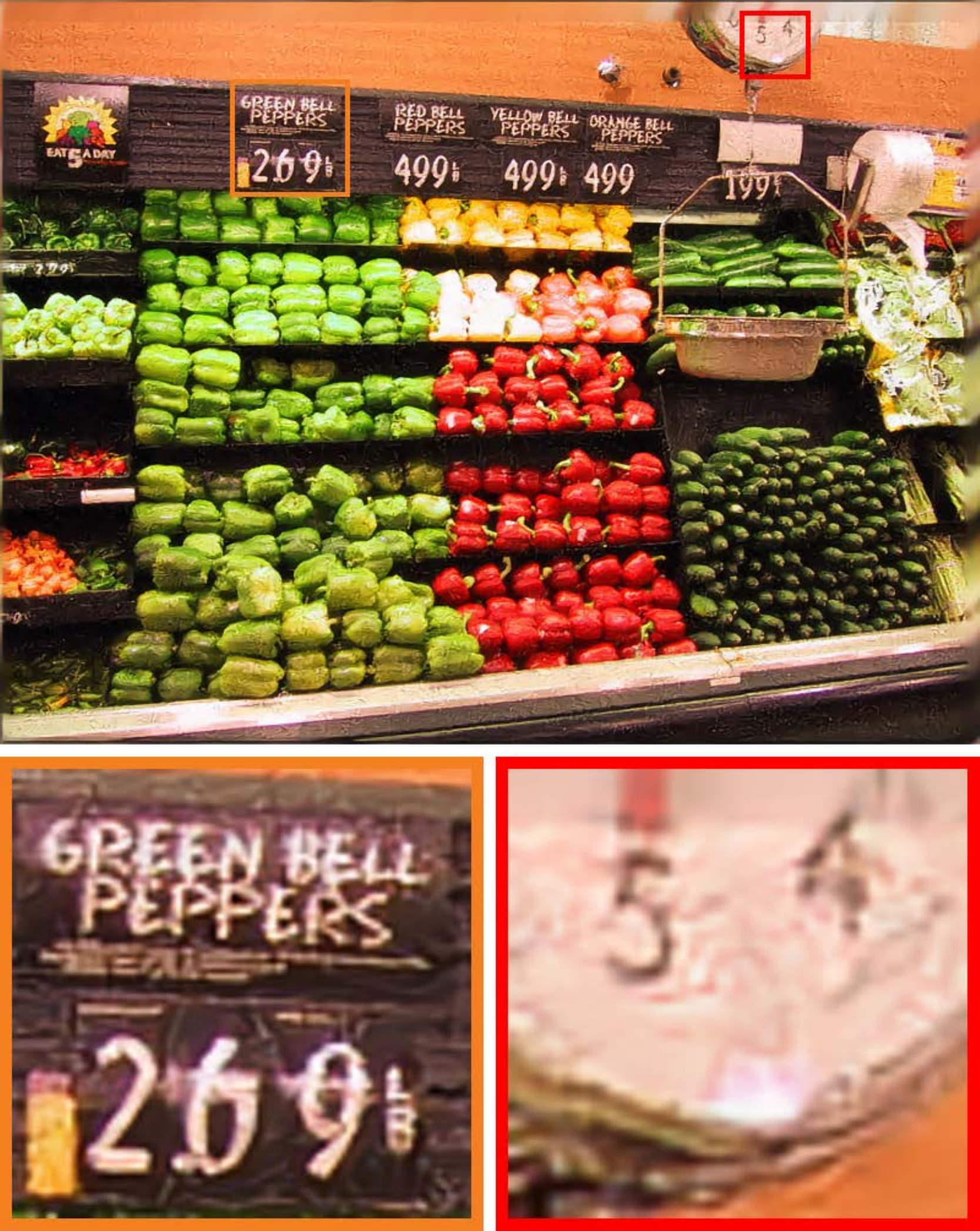}
			&\includegraphics[width=0.135\textwidth,height=0.09\textheight]{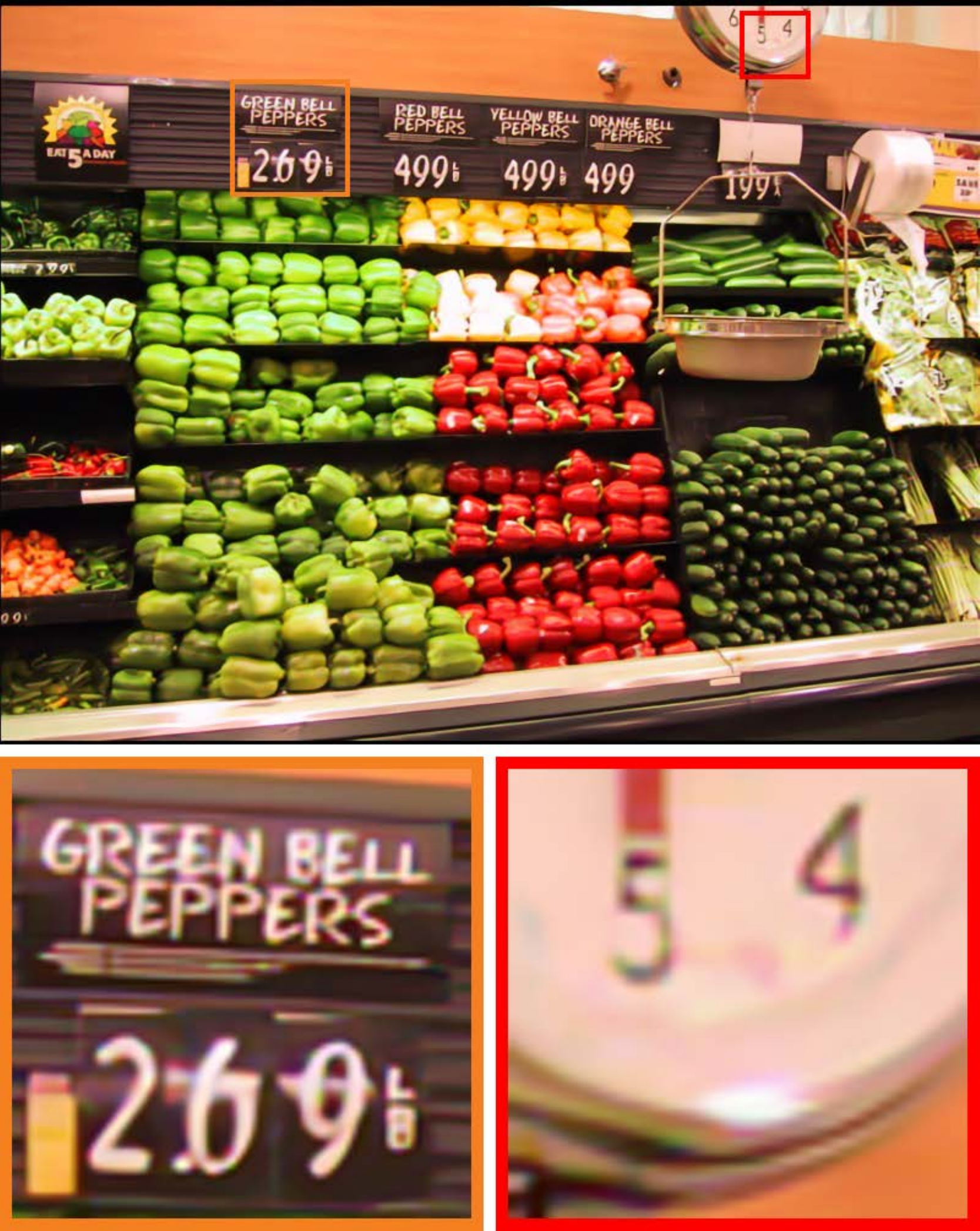}
			&\includegraphics[width=0.135\textwidth,height=0.09\textheight]{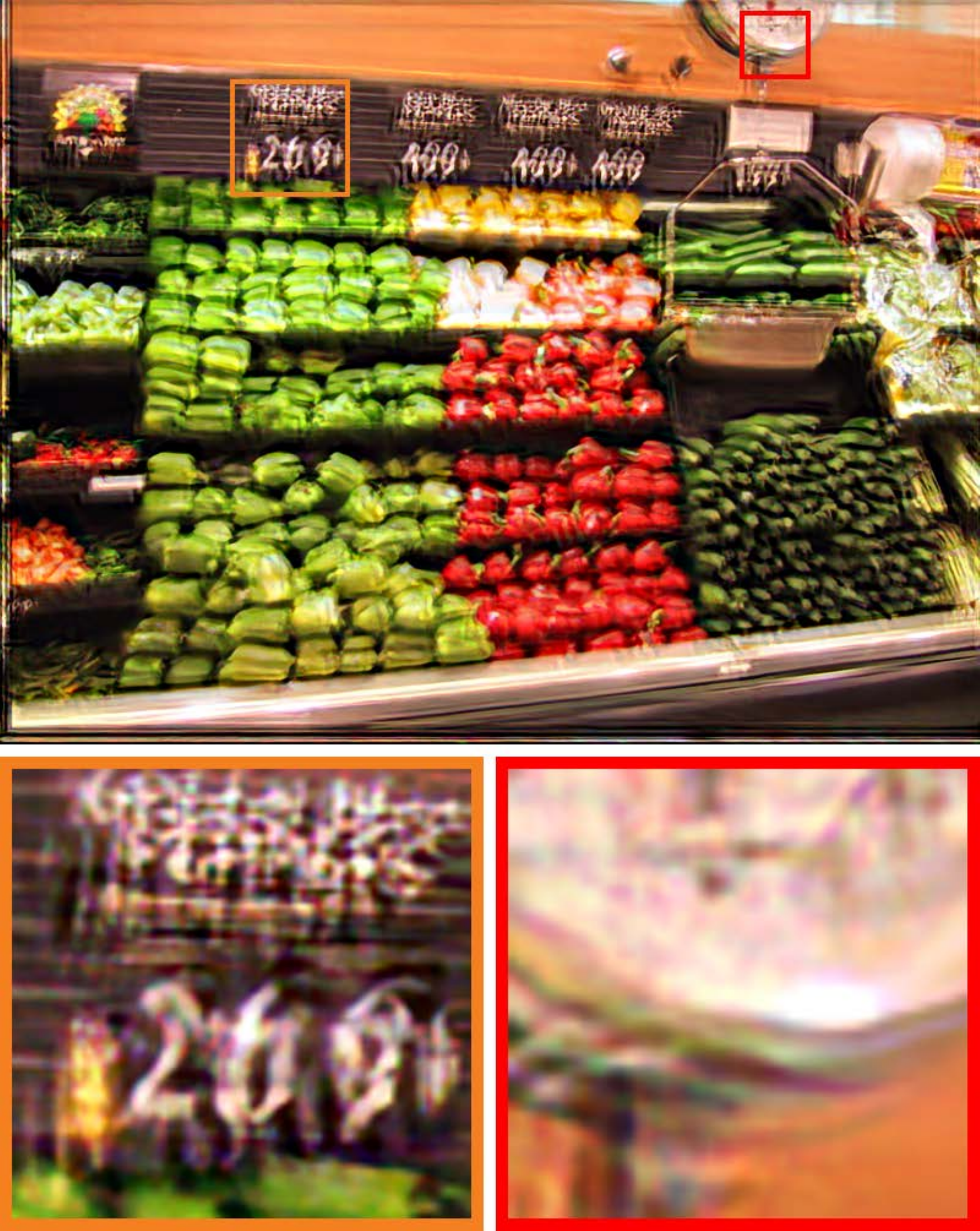}
			&\includegraphics[width=0.135\textwidth,height=0.09\textheight]{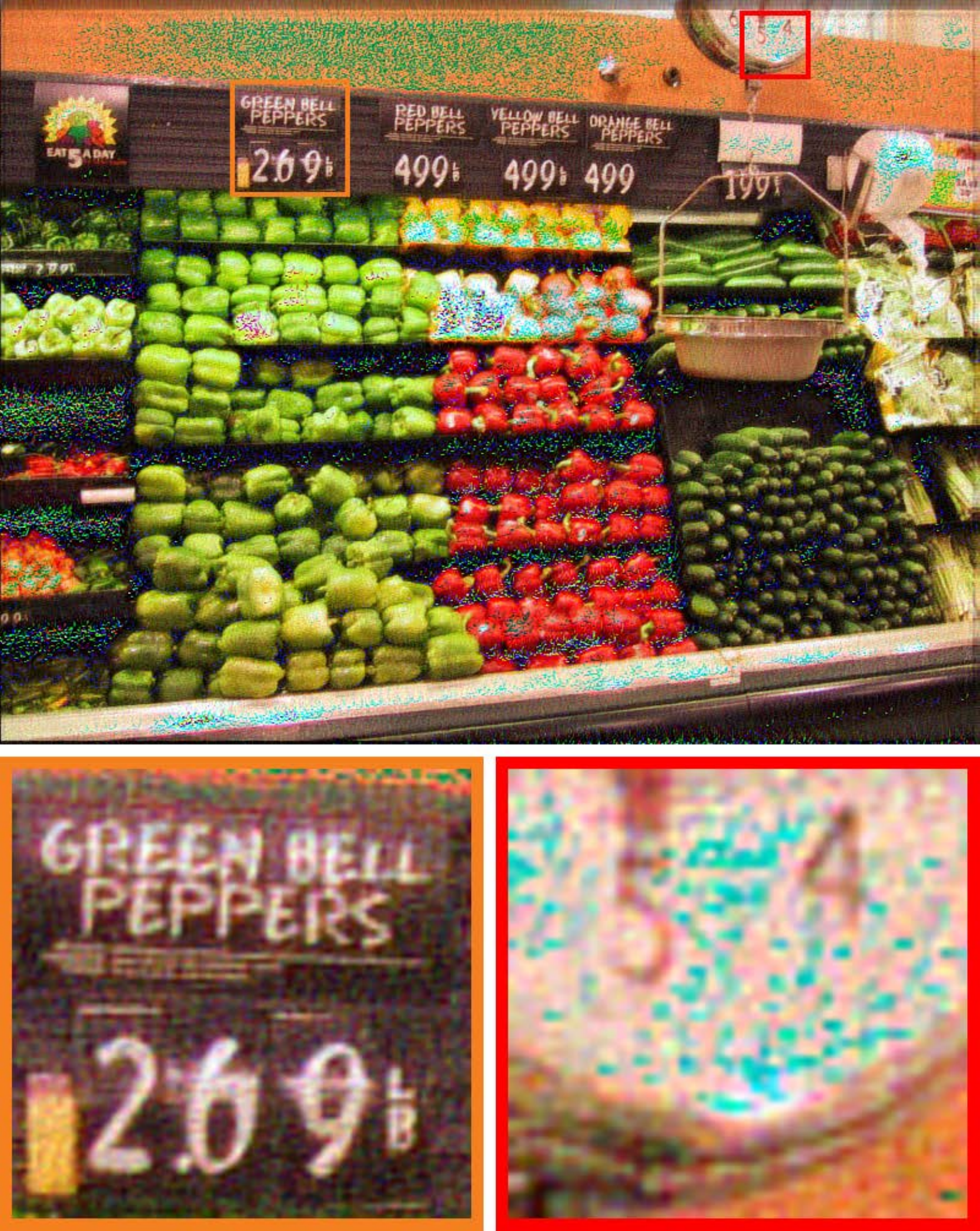}
			&\includegraphics[width=0.135\textwidth,height=0.09\textheight]{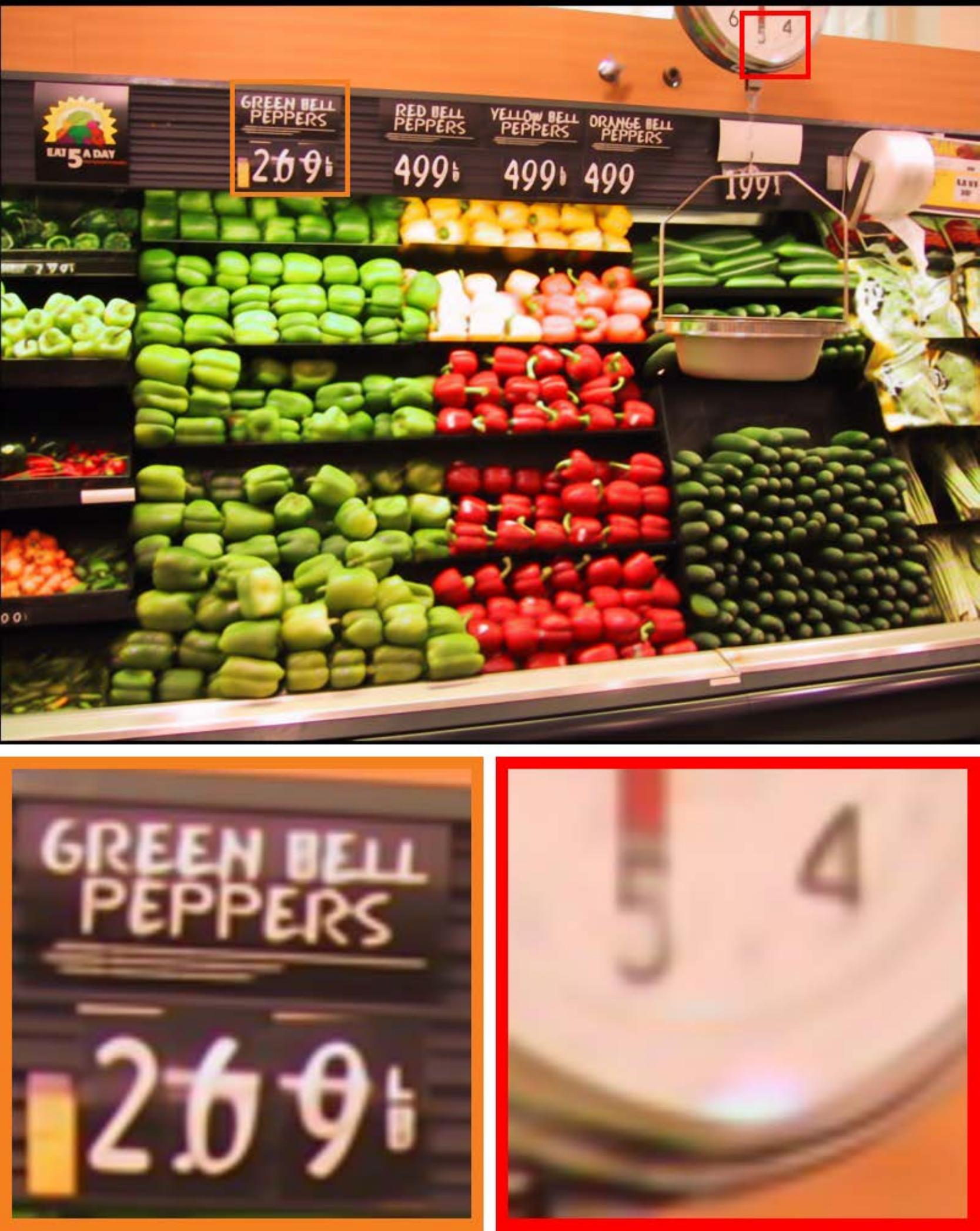}
			&\includegraphics[width=0.135\textwidth,height=0.09\textheight]{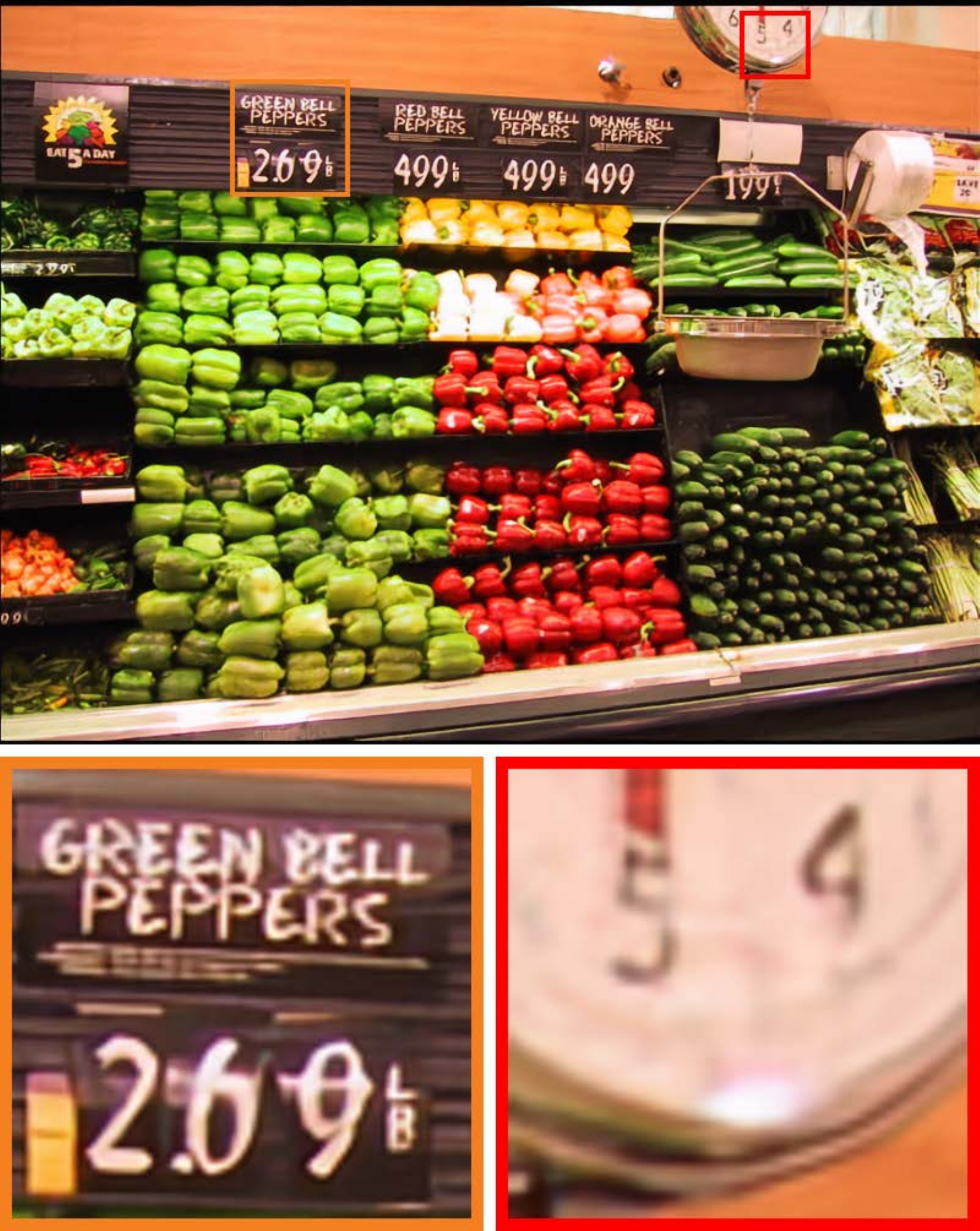}\\

			\includegraphics[width=0.135\textwidth,height=0.09\textheight]{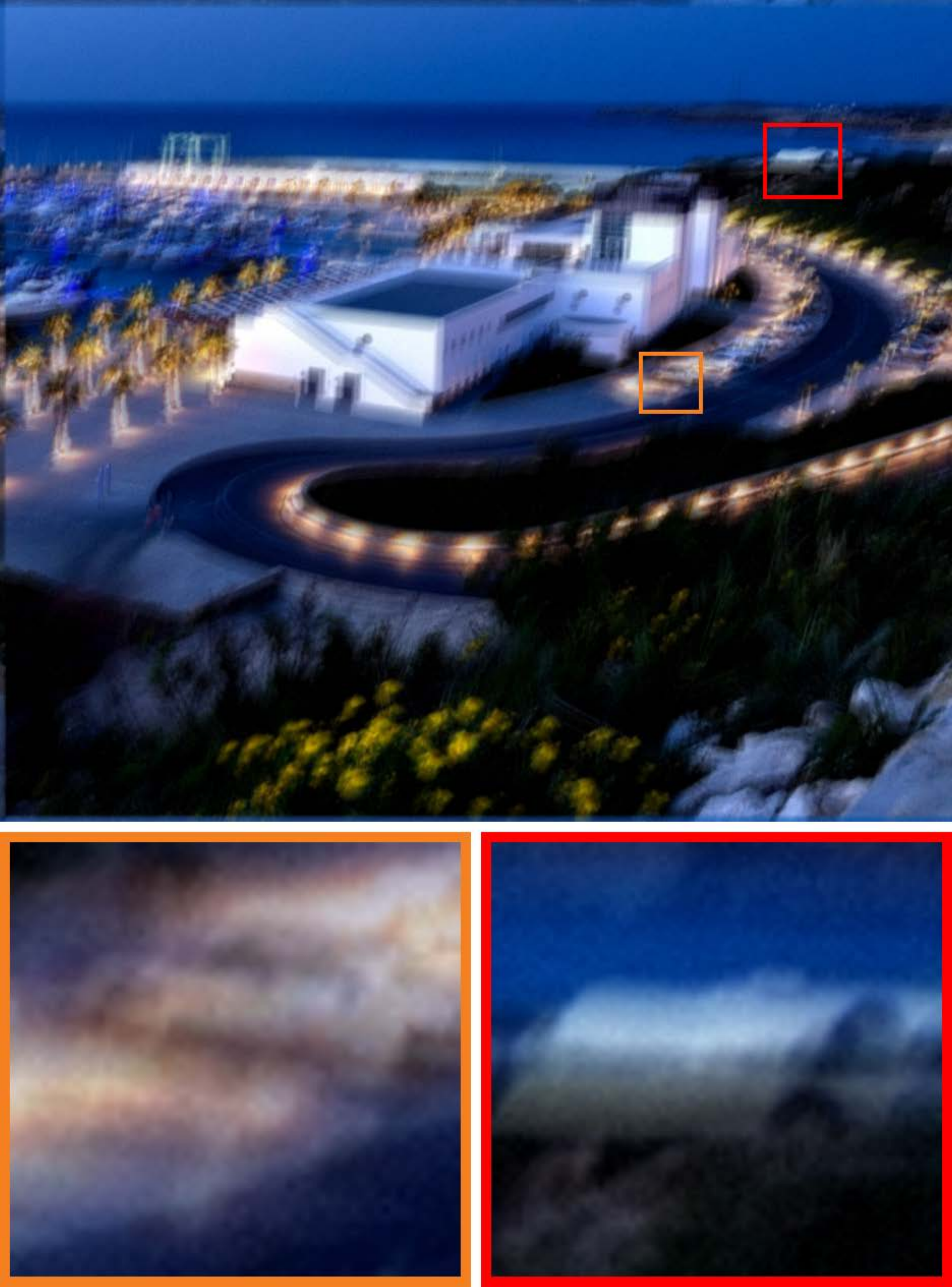}
			&\includegraphics[width=0.135\textwidth,height=0.09\textheight]{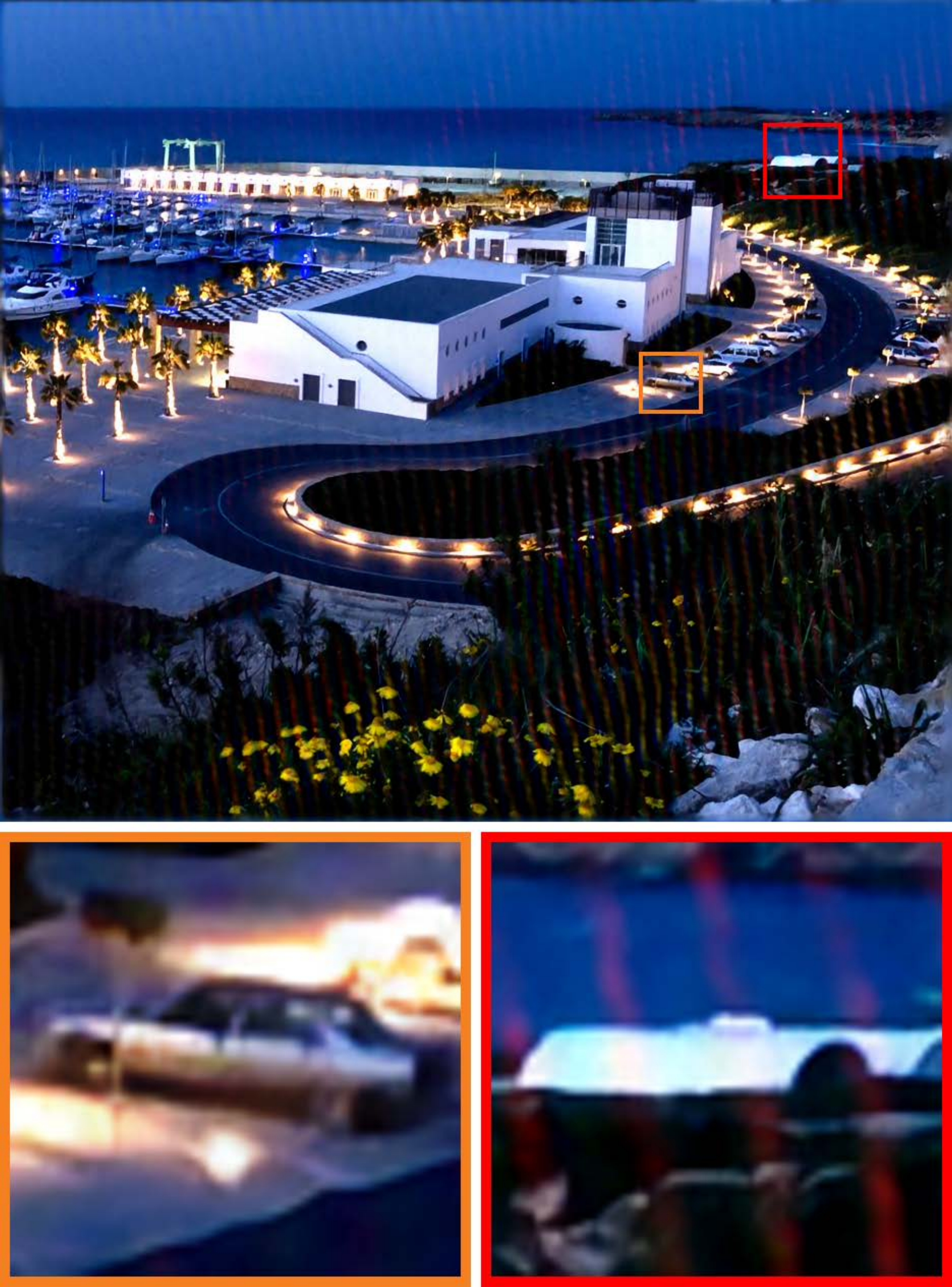}
			&\includegraphics[width=0.135\textwidth,height=0.09\textheight]{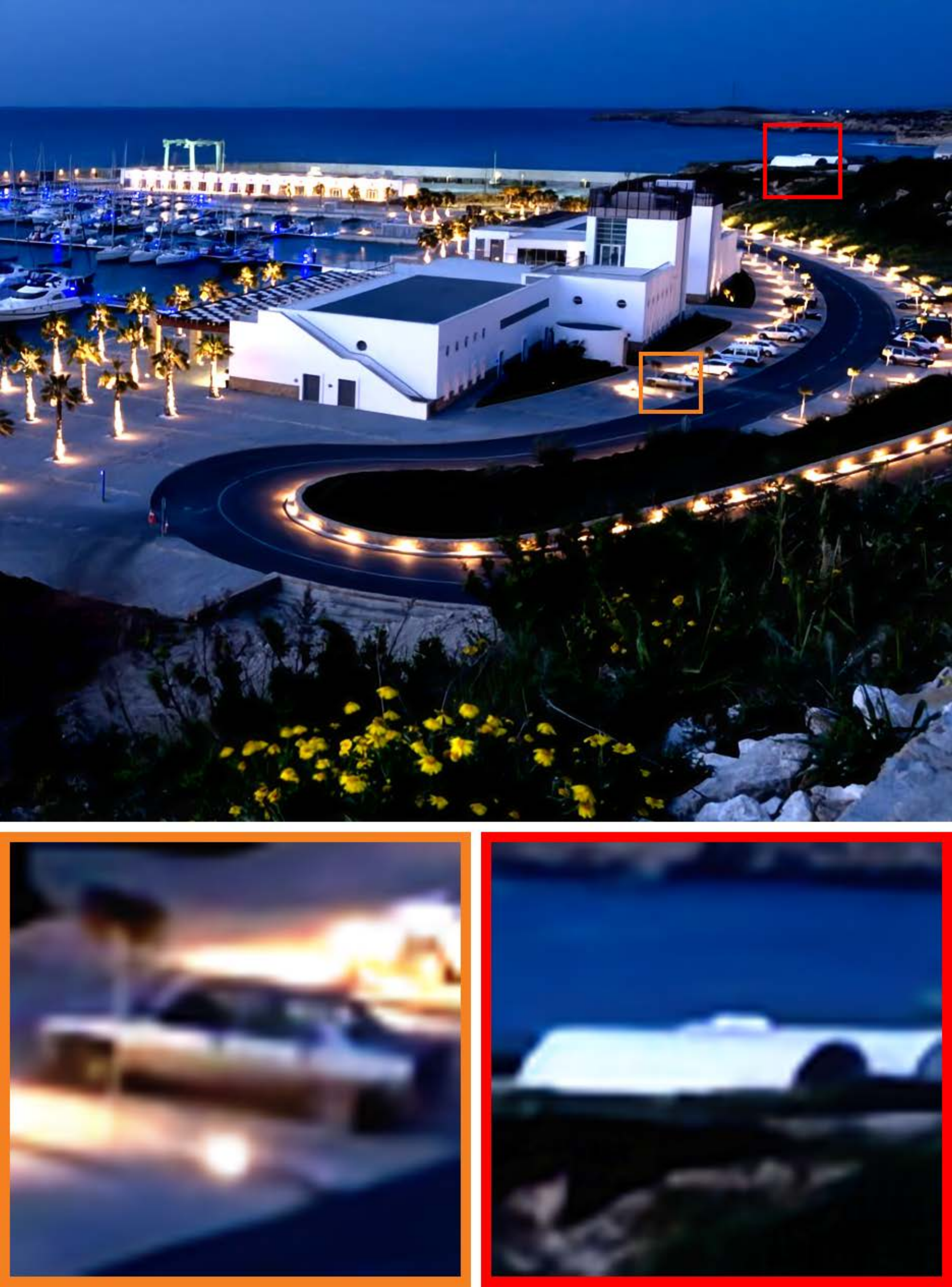}
			&\includegraphics[width=0.135\textwidth,height=0.09\textheight]{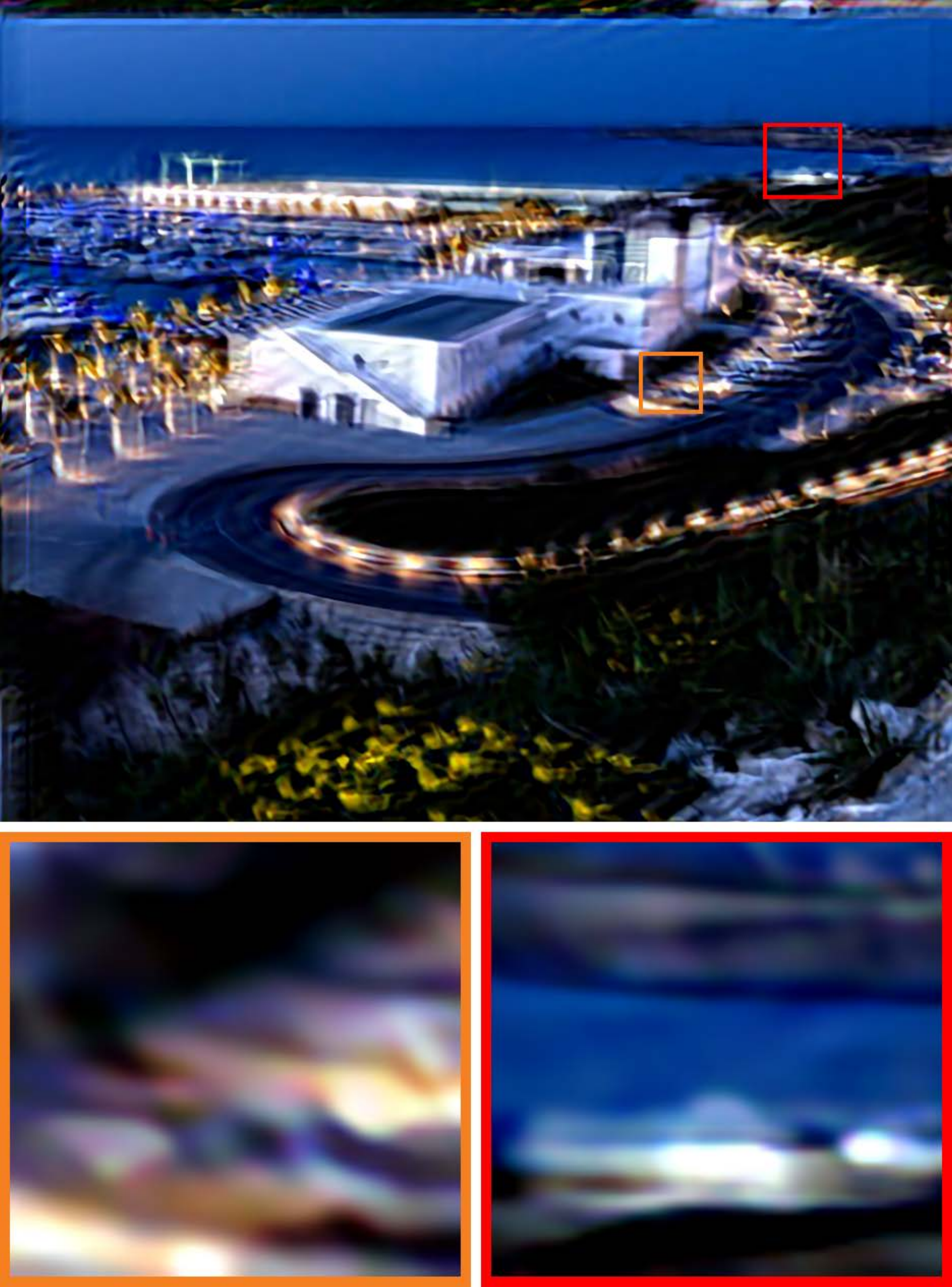}
			&\includegraphics[width=0.135\textwidth,height=0.09\textheight]{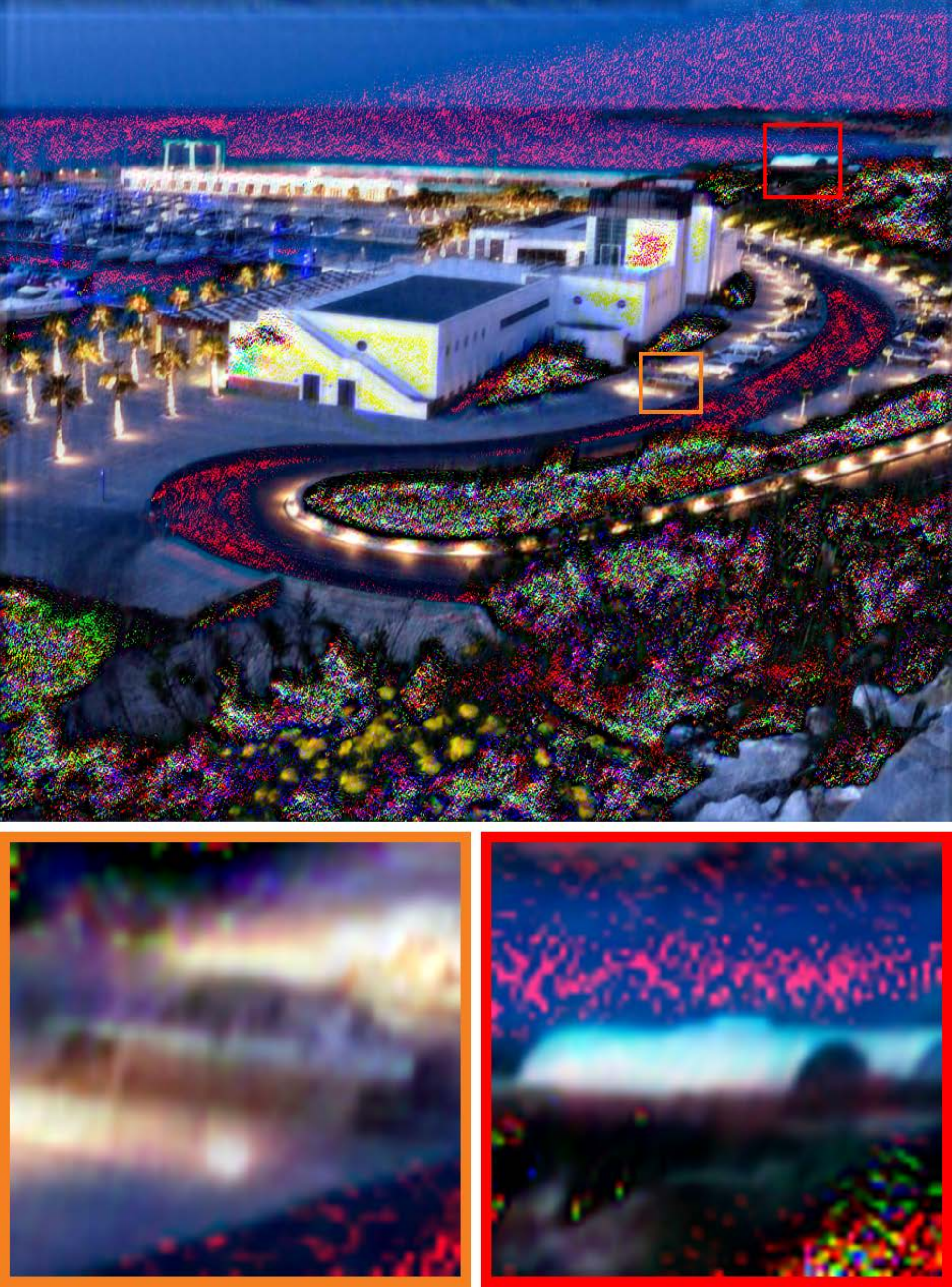}
			&\includegraphics[width=0.135\textwidth,height=0.09\textheight]{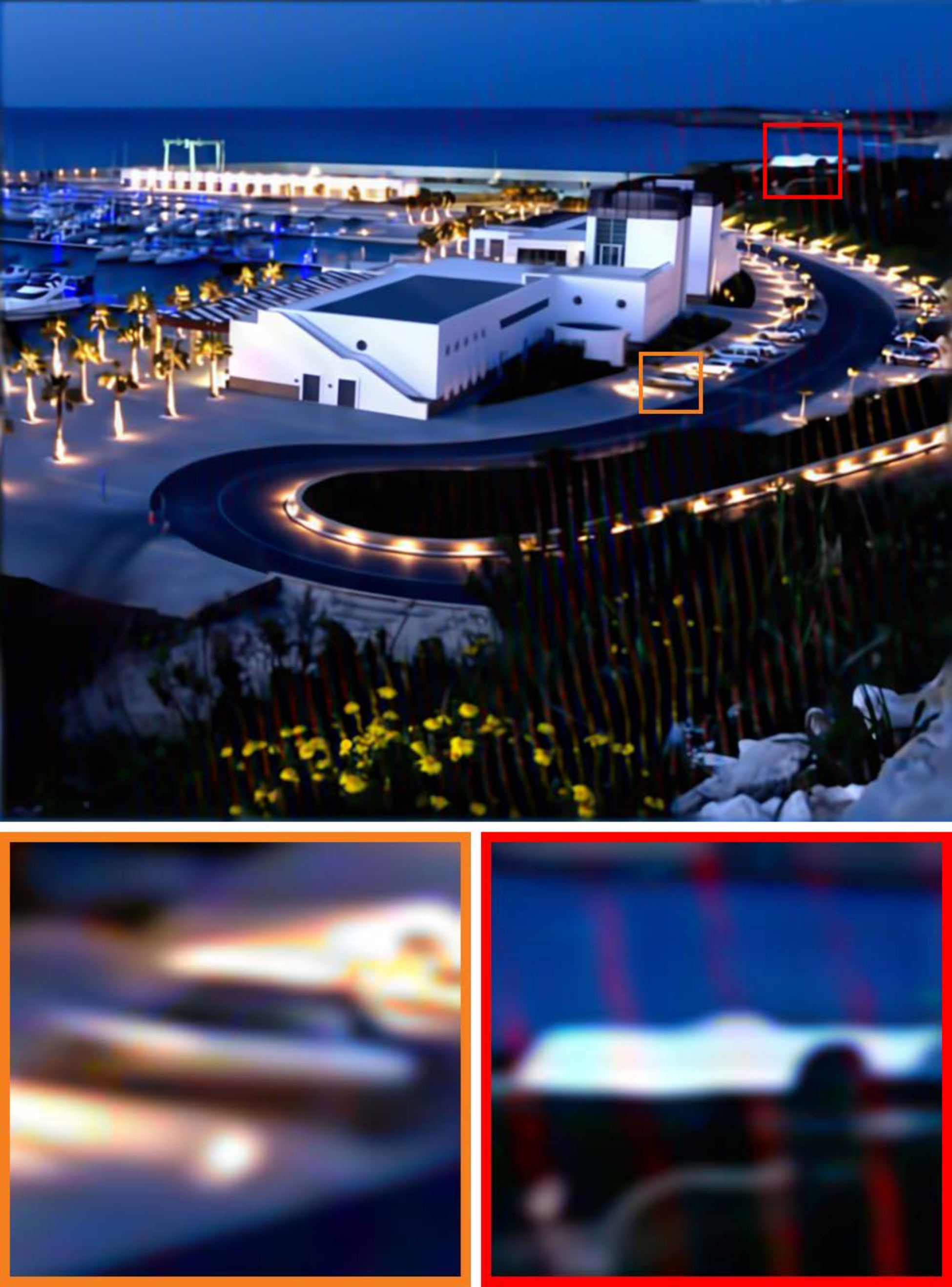}
			&\includegraphics[width=0.135\textwidth,height=0.09\textheight]{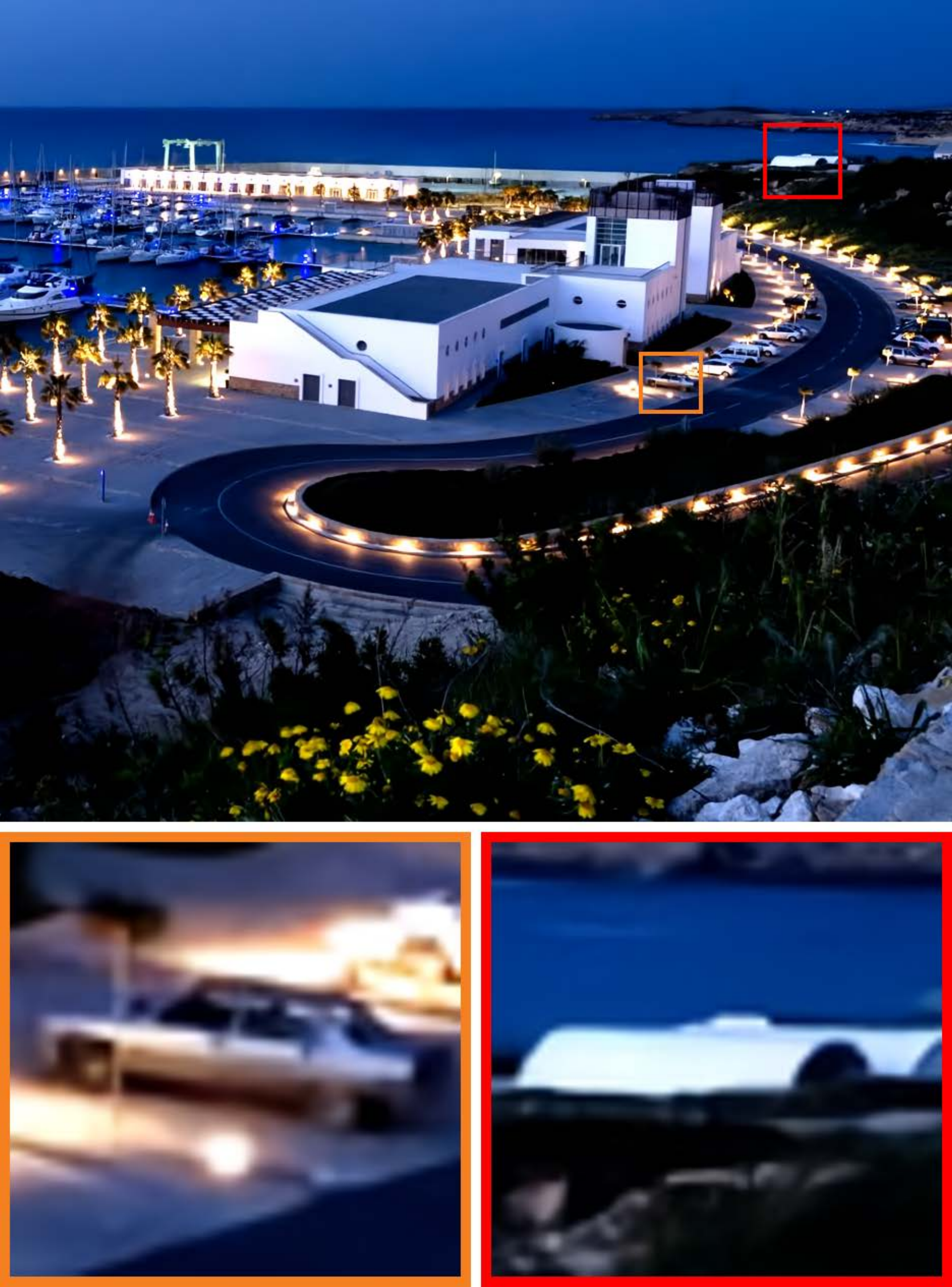}\\

			\footnotesize	(a) Input & \footnotesize (b) FDN~\cite{kruse2017learning}& \footnotesize (c) IRCNN~\cite{zhang2017learning} & \footnotesize (d) CPCR~\cite{eboli2020end} & \footnotesize (f) KerUnc~\cite{nan2020deep} & \footnotesize (e) DPIR~\cite{zhang2021plug}  & \footnotesize (f) Ours \\
		\end{tabular}
		\caption{Non-blind deconvolution  comparisons with other competitive methods under 1$\%$ and 3$\%$ Gaussian noise respectively. 
		}
		\label{fig:non-blind-state}
	\end{figure*}

	\textbf{PCM Evaluation:}
	We then evaluated the effectiveness of PCM for this FDM task. As shown in Fig.~\ref{fig:sns}, we compared the propagation behaviors with and without PCM, where  we exploited the reconstruction error to stop  iterations automatically. The  curves of the original GC were oscillated  and could not satisfy the automatic stop condition through  finite numbers of iterations. The scheme with PCM obtained  more stable outputs with fewer iterations that demonstrate the efficiency of pour proposed PCM.
	\begin{table}[htb]
		\renewcommand{\arraystretch}{1.2}
		\caption{ Quantitative results about image deblurring on image set (Levin et al.~\cite{levin2009understanding} / Sun et al.~\cite{sun2013edge}), lower ER is better.}
		\label{tab:data_set_results}
				\vspace{-0.2cm}
		\centering\footnotesize
		\setlength{\tabcolsep}{1.6mm}{
			\begin{tabular}{|c| c|  c| c| c|}
				\hline
				Methods &  PSNR &  SSIM &  ER &  Time (s)  \\
				\hline
		
				\cite{pan2016blind}          & 29.78/29.60  &  0.89/0.84  &  1.33/1.43  & 102.60/215.66 \\ \hline
				\cite{yan2017image}			 & 29.87/29.57  &  0.89/0.83  &  1.32/1.45  & 29.56/172.55  \\ \hline
				\cite{li2018learning}		 & 29.76/29.60  &  0.89/0.83  &  1.31/\textbf{1.41}  & 67.40/127.17  \\ \hline
				\cite{chen2019blind}&29.78/29.51&0.91/0.82&1.00/2.21&67.78/386.61\\\hline
				\cite{wen2020simple}&29.34/28.14&0.89/0.81&\textbf{0.99}/2.53&13.18/219.93\\\hline
				Ours  &\textbf{30.30}/\textbf{29.70}  &\textbf{0.92}/0.84 &{1.37/1.80} & \textbf{10.79}/194.9\\
				\hline
			\end{tabular}	
		}
	\end{table}

	\section{Experimental Results}\label{sec:result}
	
	In this section, comprehensive experimental results have demonstrated the superiority of GDC on various low-level vision applications. 
	\subsection{Low-Level Vision Tasks}
	\textbf{Non-blind deblurring:}
	We compared our method against state-of-the-art approaches, including  FDN~\cite{kruse2017learning}, IRCNN~\cite{zhang2017learning},  CPCR~\cite{eboli2020end} and KerUnc~\cite{nan2020deep}, and recent proposed DPIR~\cite{zhang2021plug},  UDN~\cite{chen2022nonblind} and PLD~\cite{sanghvi2022photon}. We conducted the comparisons  on widely used gray datasets, \textit{i.e.,} Levin \emph{et al.}~\cite{levin2009understanding}  and BSD68~\cite{roth2009fields} under  1\% Gaussian noise.  To compare our method, we employ the joint training version of GC. All the compared methods are unrolled schemes.
	As shown in Fig.~\ref{fig:non-blind-state}, our method is  effective in  preserving the clear structure  and generating perceptually-pleasant outputs.
	We also reported the average quantitative comparison in terms of PSNR and SSIM in Table~\ref{tab:ablationtabless}.
	Our  scheme with AN using joint training, performed favorably against a series of conventional and learning-based deblurring methods.

	\begin{table}[htb]	\renewcommand{\arraystretch}{1.1}
		\caption{Quantitative results with learning methods on Levin \textit{et.al}.}
		\label{tab:dataset_sun}
		\centering \footnotesize
				\vspace{-0.2cm}
		\setlength{\tabcolsep}{2.8mm}{
			\begin{tabular}{|c| c| c |c |c |c|}
				\hline
				Metrics &  ~\cite{Mou2022DGUNet} & ~\cite{ji2022xydeblur} &  ~\cite{zamir2022restormer} & ~\cite{Wang_2022_CVPR}  &  Ours  \\ \hline		
				PSNR  & 20.72 & 16.32 & 20.25& 21.28 & \textbf{30.30} \\ 
				SSIM &	0.57 & 0.37 & 0.56 &0.58  & \textbf{0.92} \\ 
				\hline
			\end{tabular}	
		}
	\end{table}

	\begin{figure*}[htbp] 
		\centering \begin{tabular}{c@{\extracolsep{0.1em}}c@{\extracolsep{0.1em}}c@{\extracolsep{0.1em}}c@{\extracolsep{0.1em}}c@{\extracolsep{0.1em}}c@{\extracolsep{0.1em}}c}
			\includegraphics[width=0.135\textwidth,height=0.09\textheight]{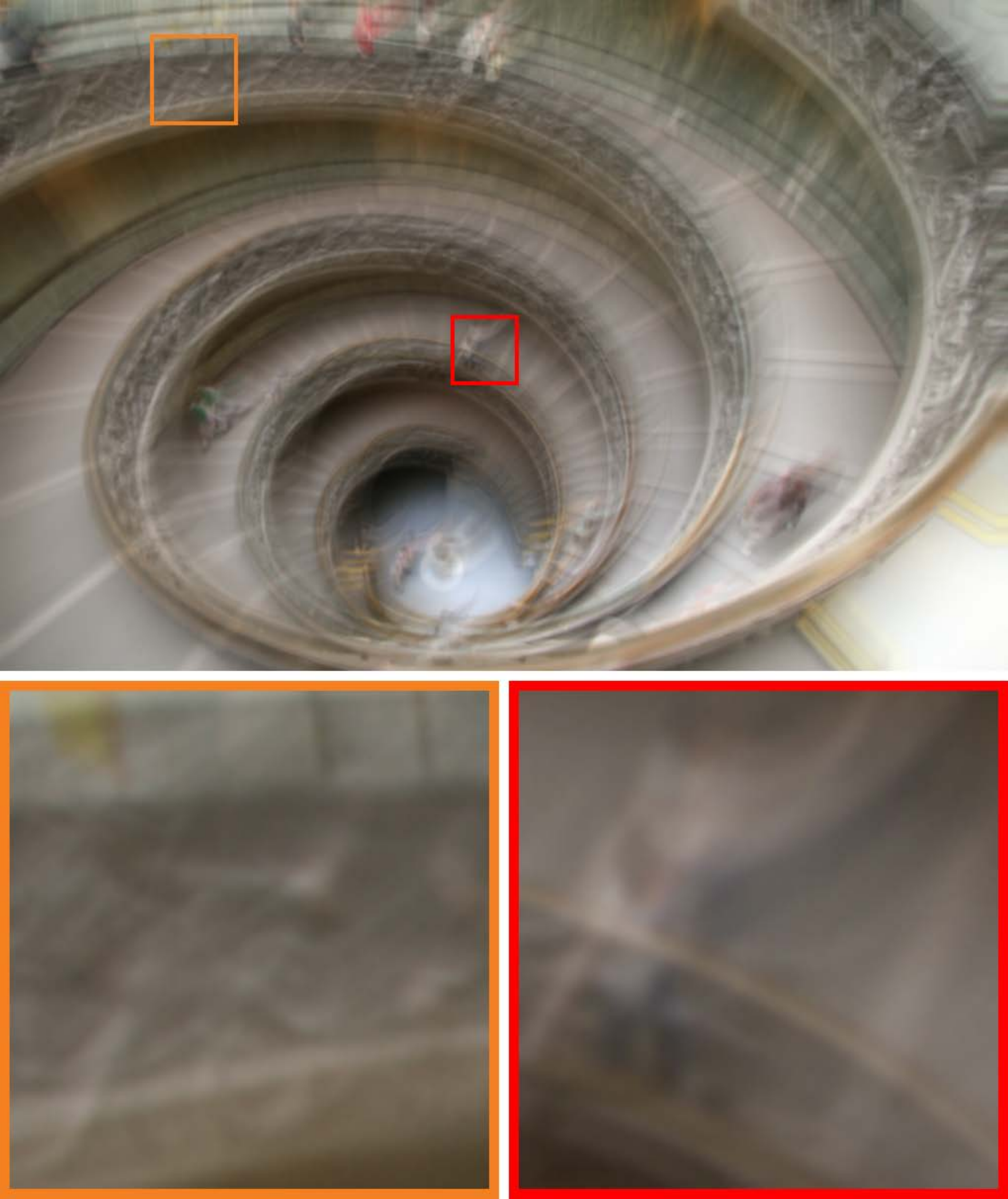}
			&\includegraphics[width=0.135\textwidth,height=0.09\textheight]{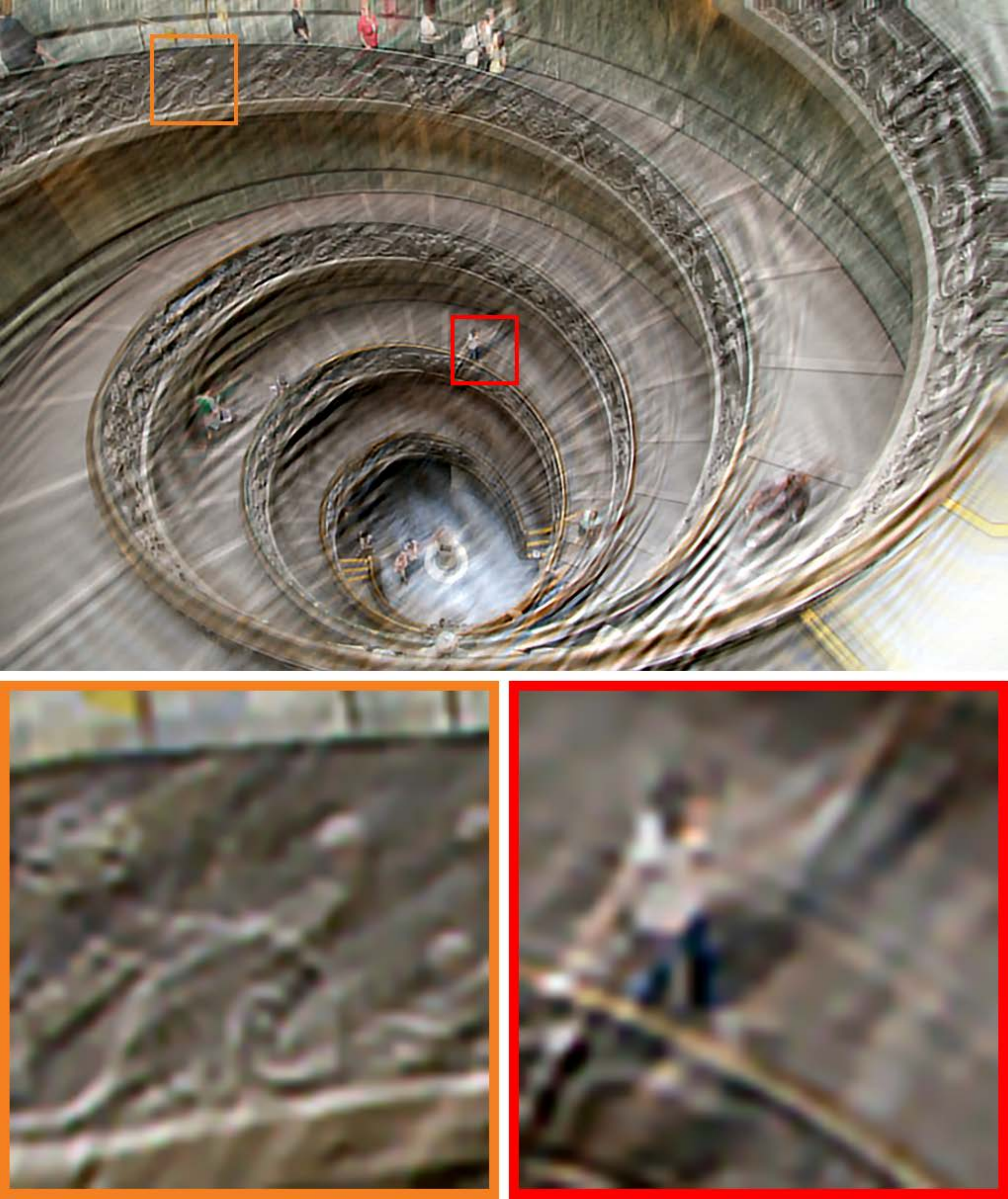}
			&\includegraphics[width=0.135\textwidth,height=0.09\textheight]{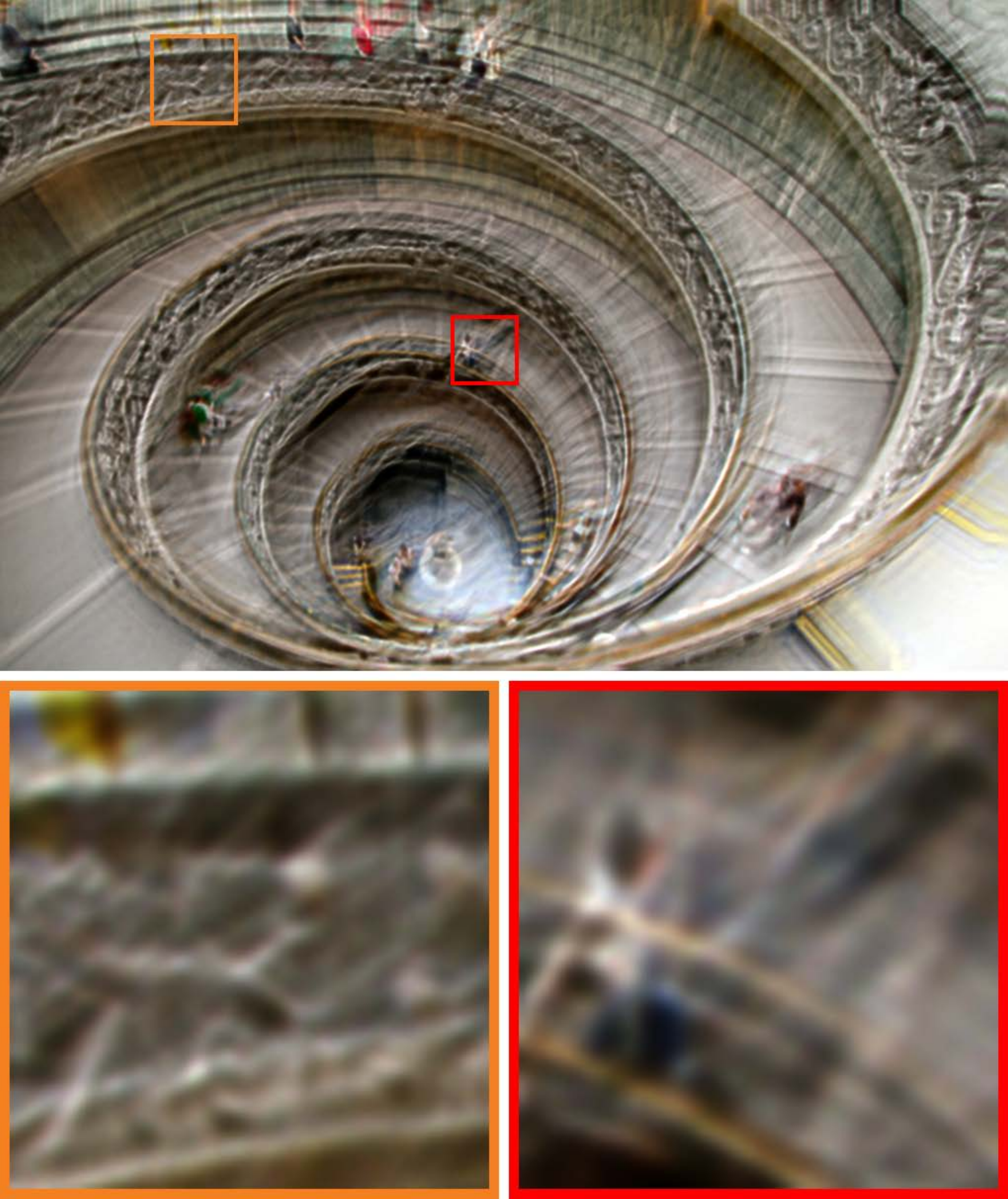}
			&\includegraphics[width=0.135\textwidth,height=0.09\textheight]{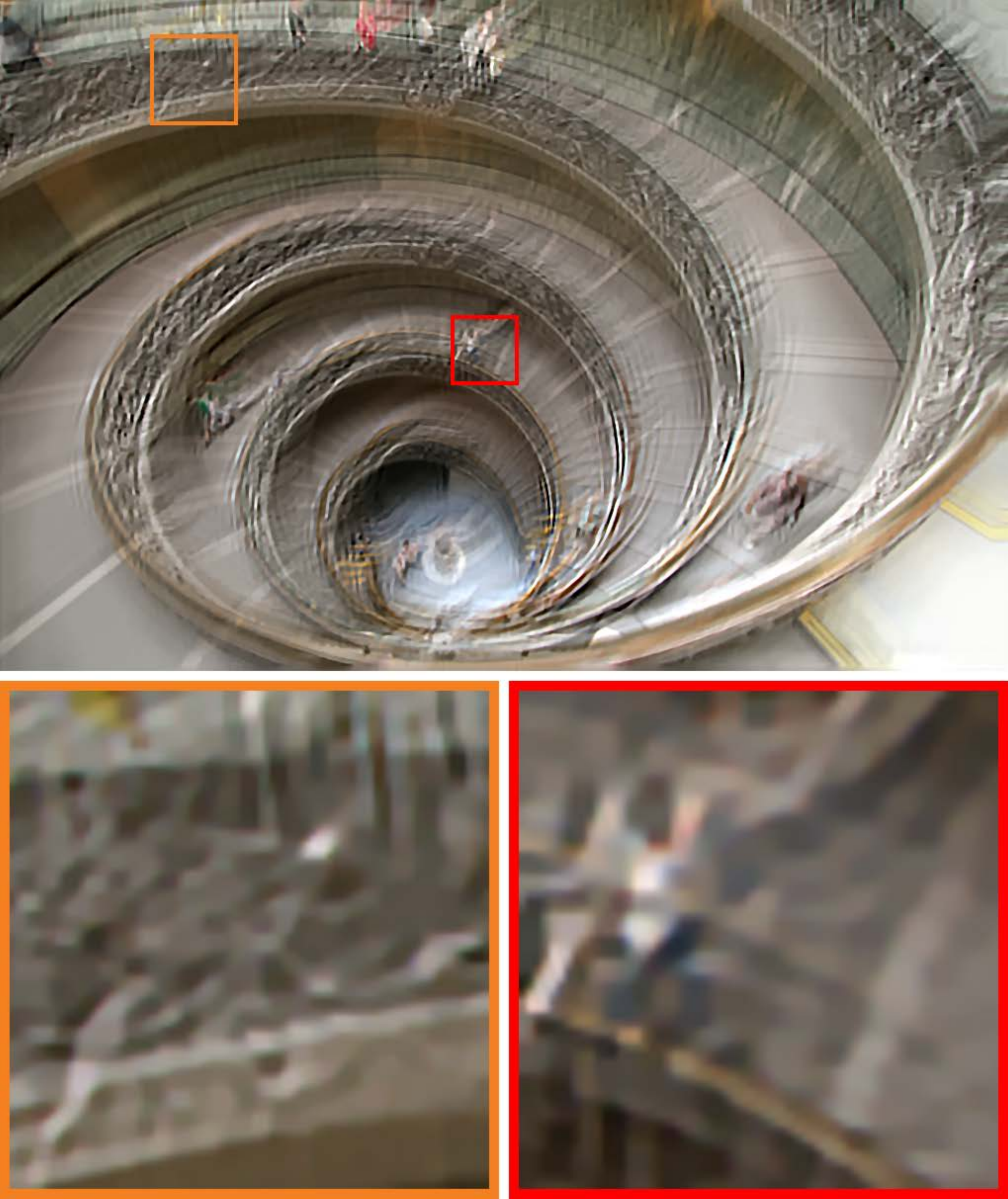}
			&\includegraphics[width=0.135\textwidth,height=0.09\textheight]{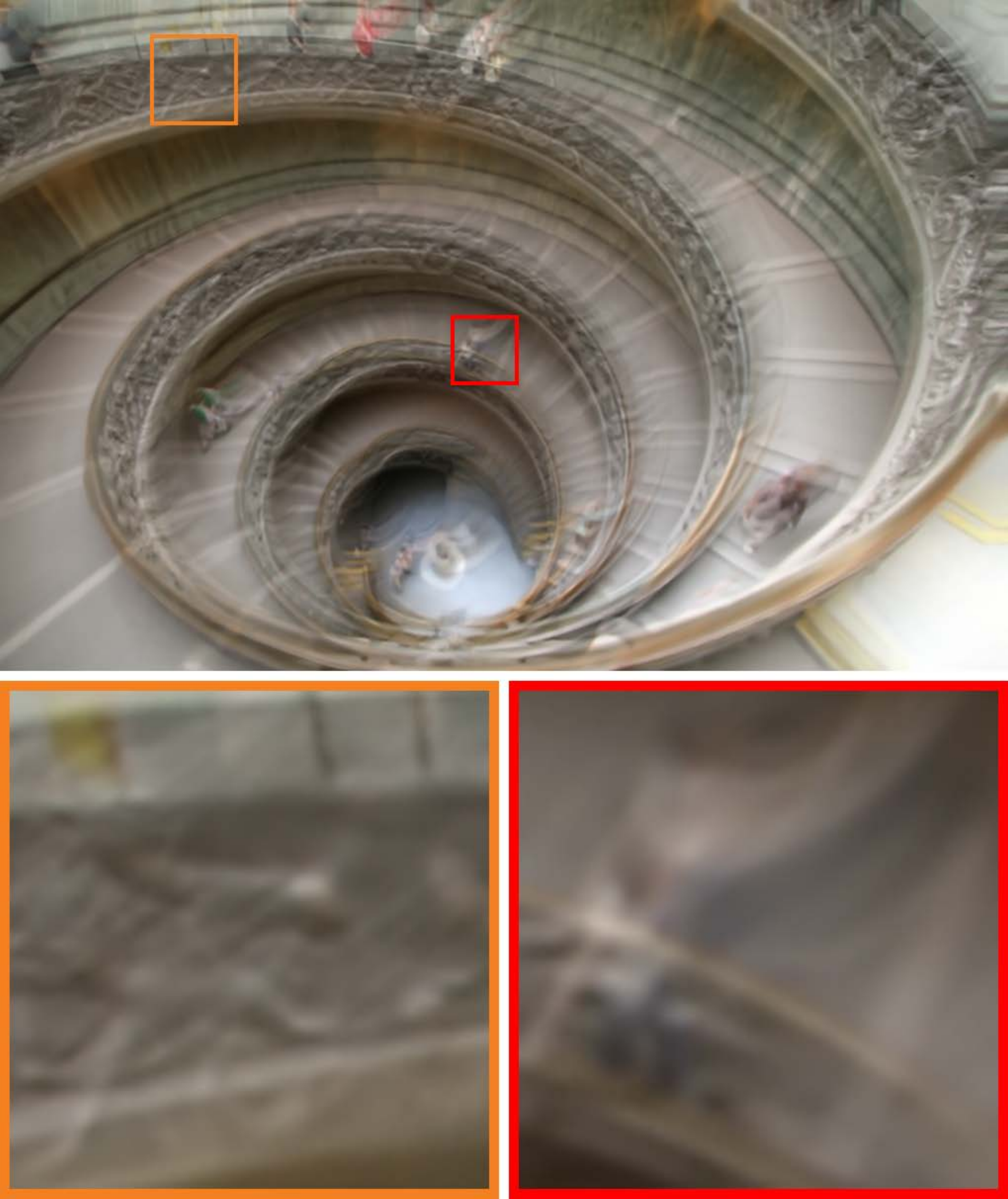}
			&\includegraphics[width=0.135\textwidth,height=0.09\textheight]{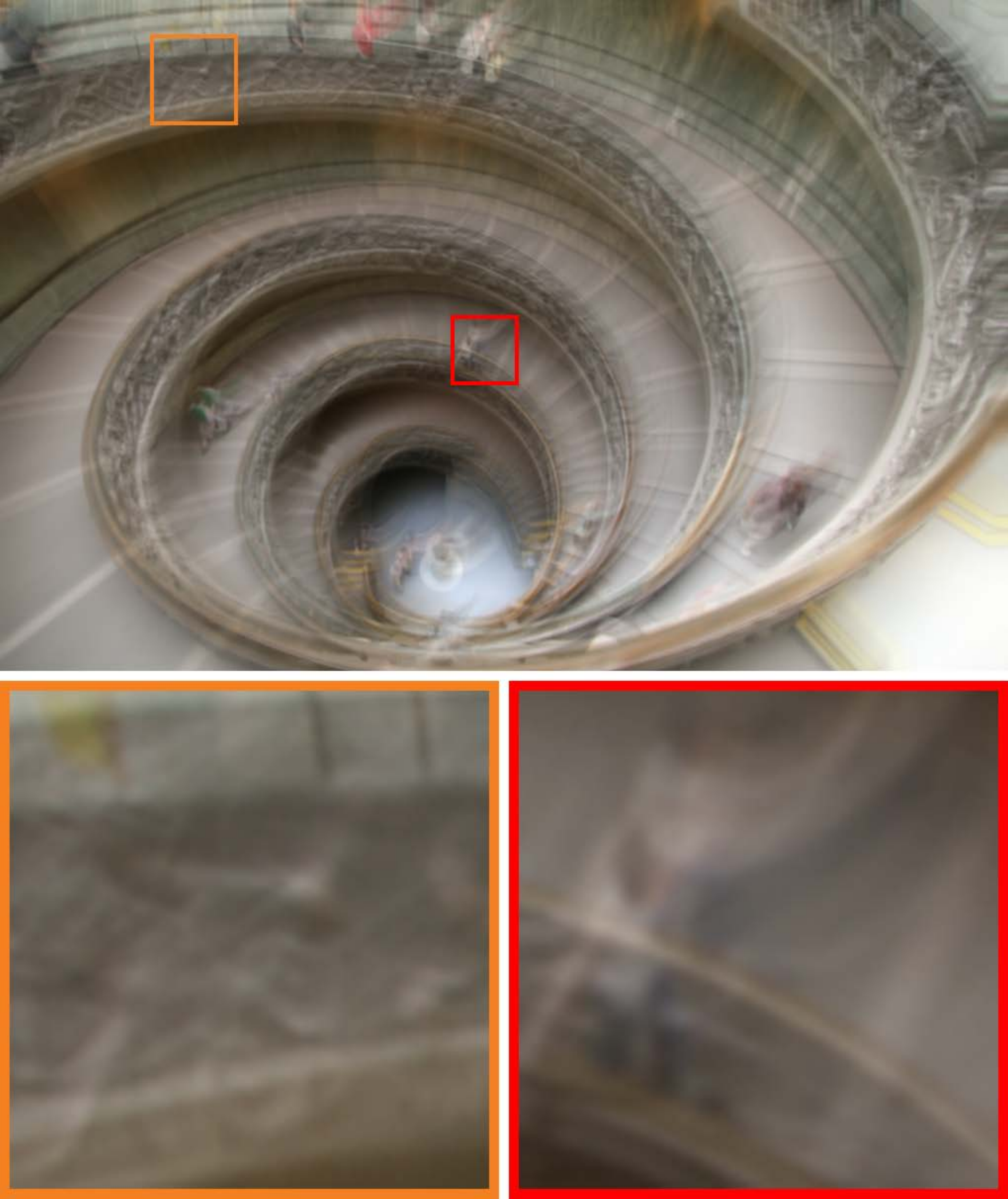}
			&\includegraphics[width=0.135\textwidth,height=0.09\textheight]{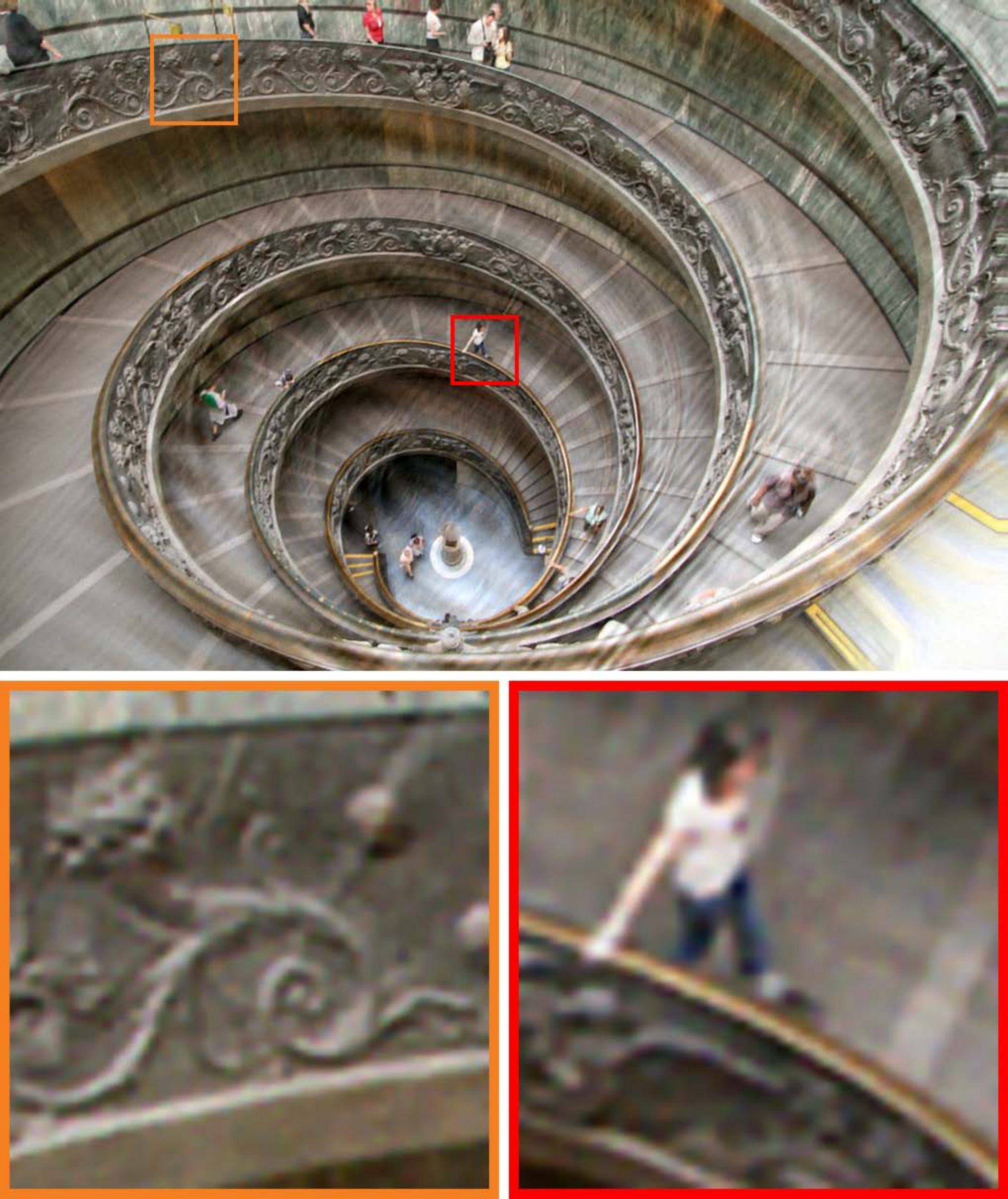}\\
			\includegraphics[width=0.135\textwidth,height=0.07\textheight]{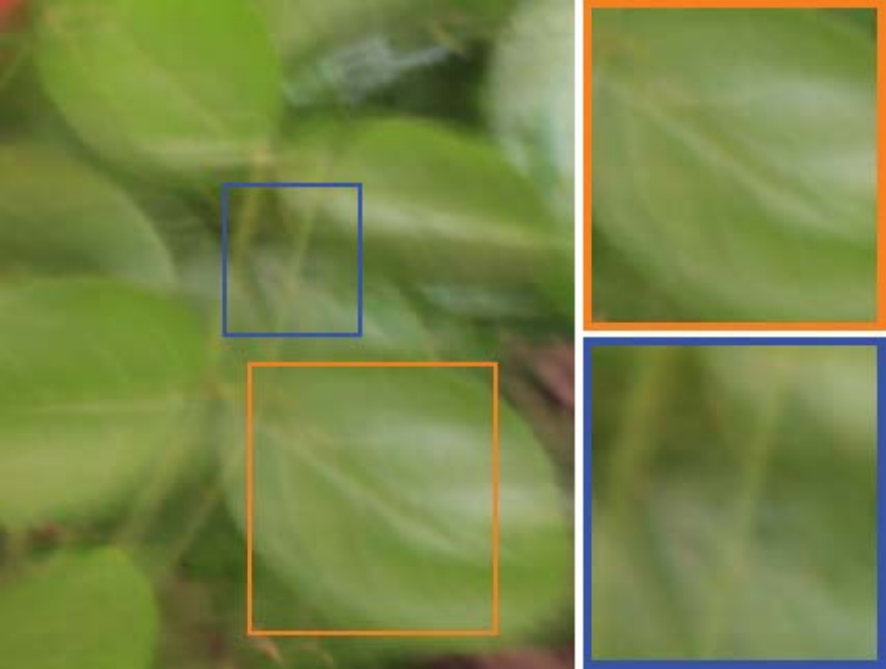}
			&\includegraphics[width=0.135\textwidth,height=0.07\textheight]{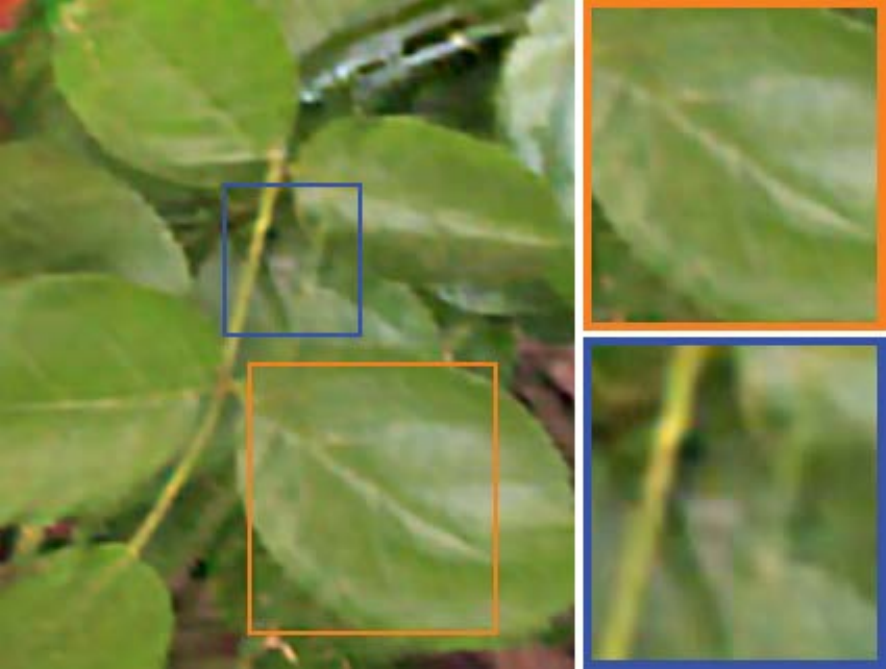}
			&\includegraphics[width=0.135\textwidth,height=0.07\textheight]{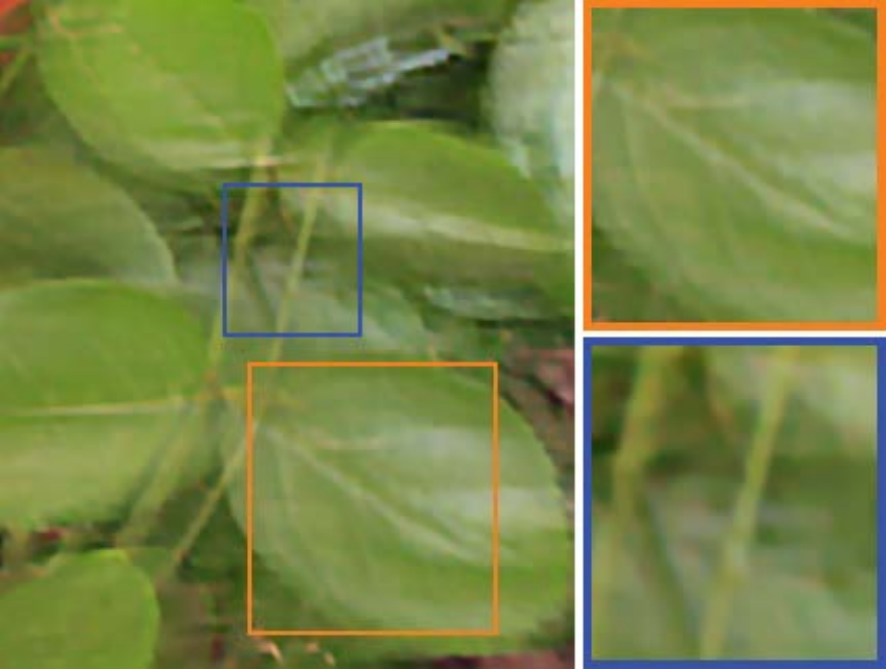}
			&\includegraphics[width=0.135\textwidth,height=0.07\textheight]{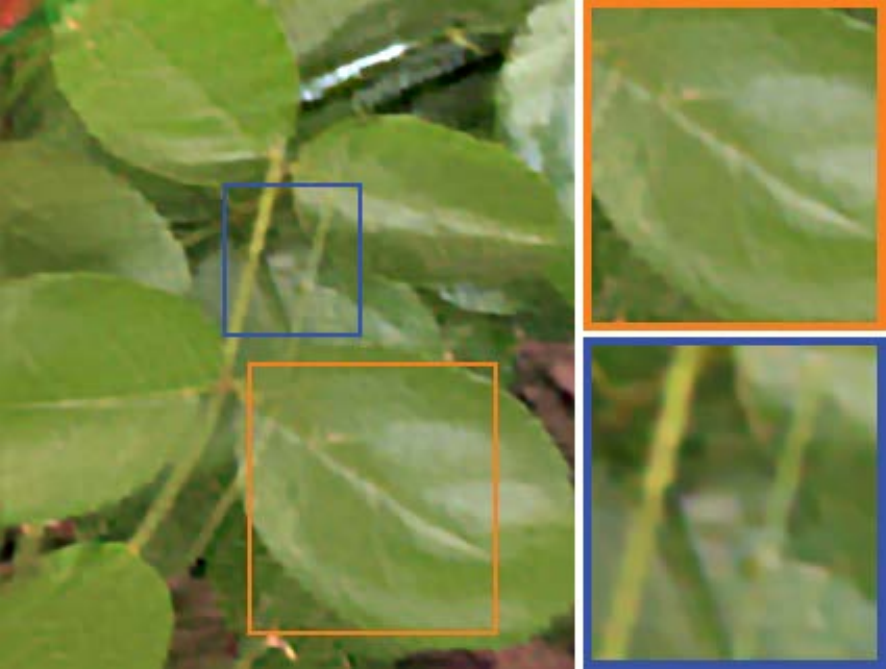}
			&\includegraphics[width=0.135\textwidth,height=0.07\textheight]{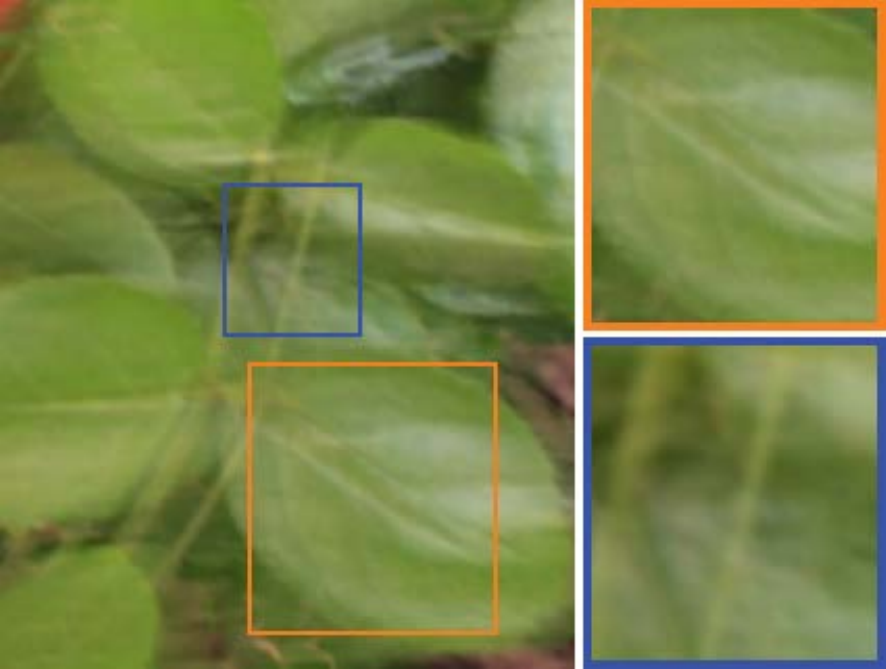}
			&\includegraphics[width=0.135\textwidth,height=0.07\textheight]{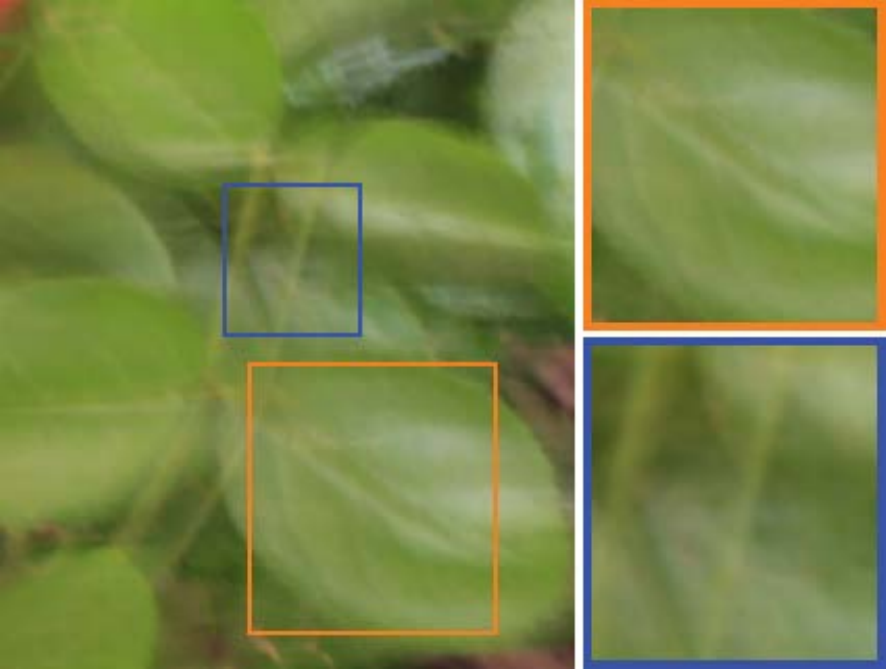}
			&\includegraphics[width=0.135\textwidth,height=0.07\textheight]{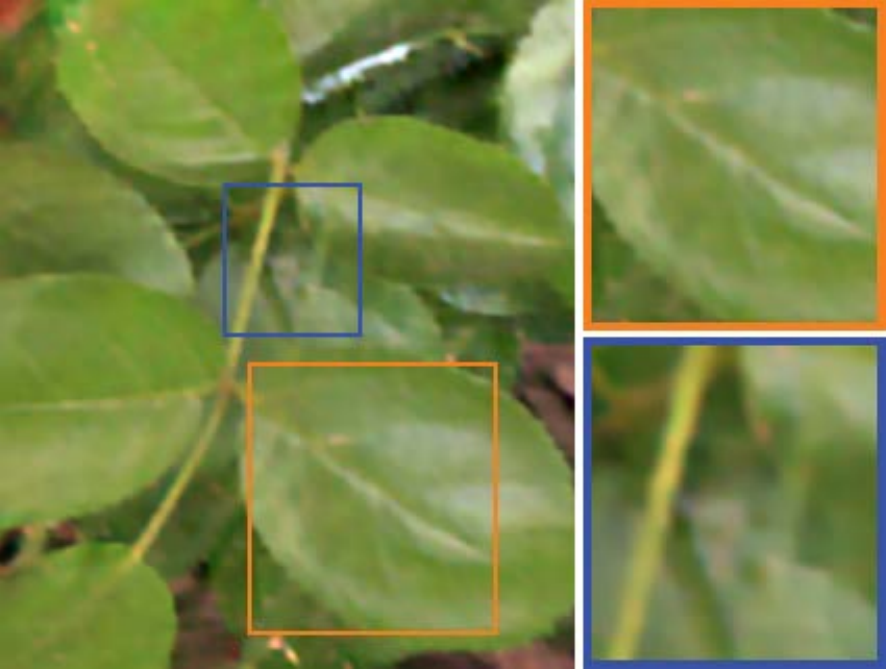}\\
			
			\includegraphics[width=0.135\textwidth,height=0.09\textheight]{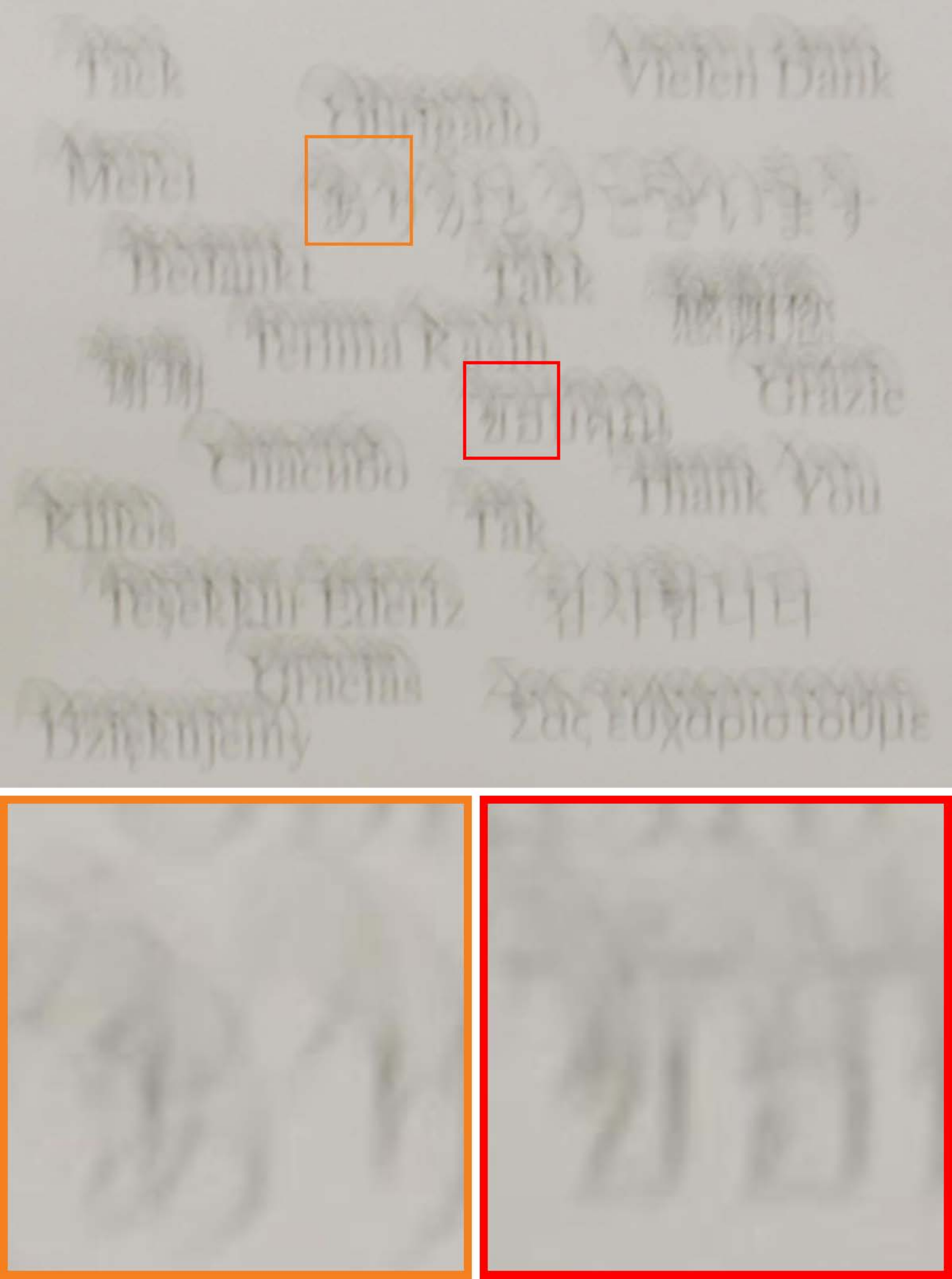}
			&\includegraphics[width=0.135\textwidth,height=0.09\textheight]{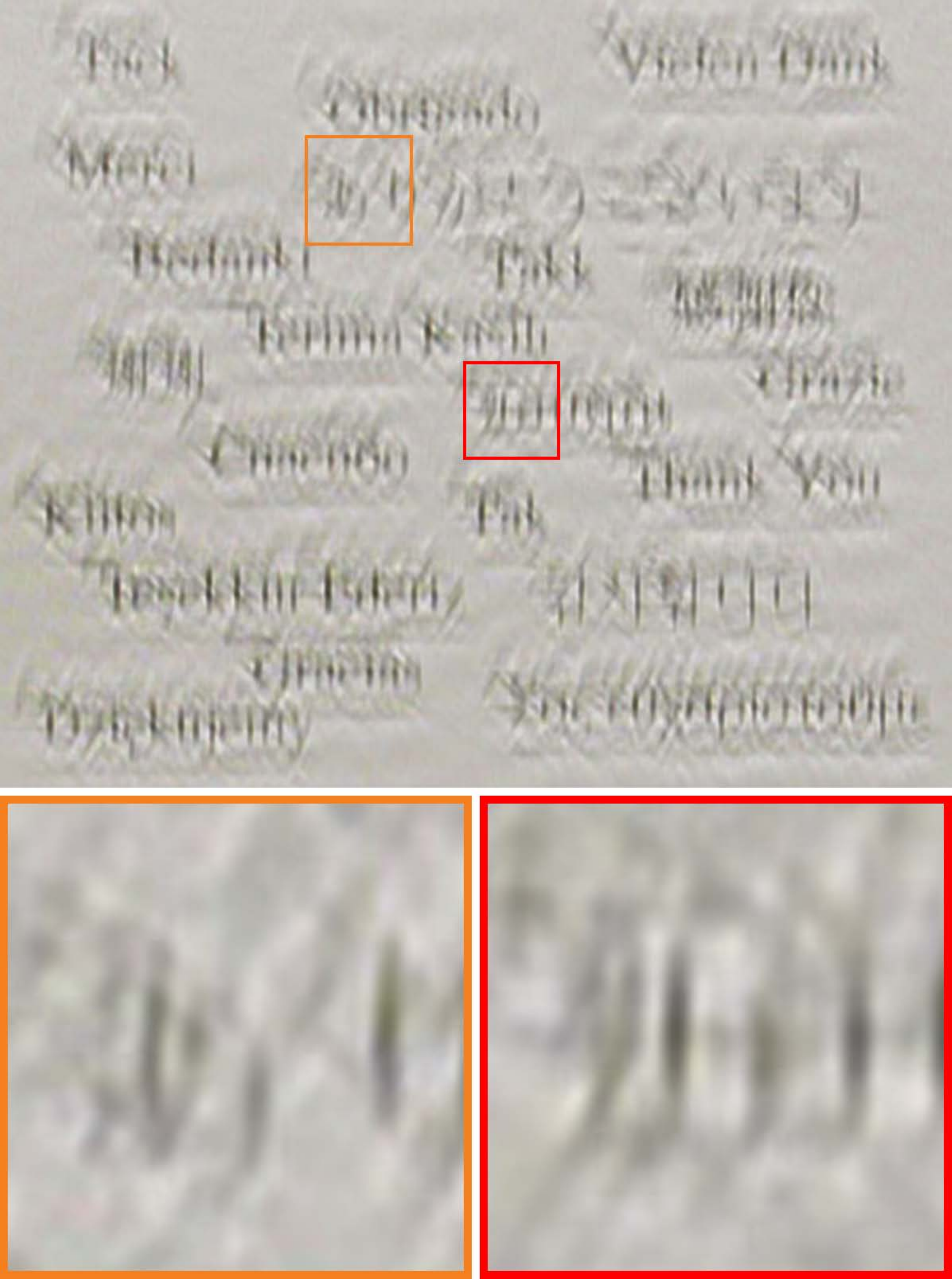}
			&\includegraphics[width=0.135\textwidth,height=0.09\textheight]{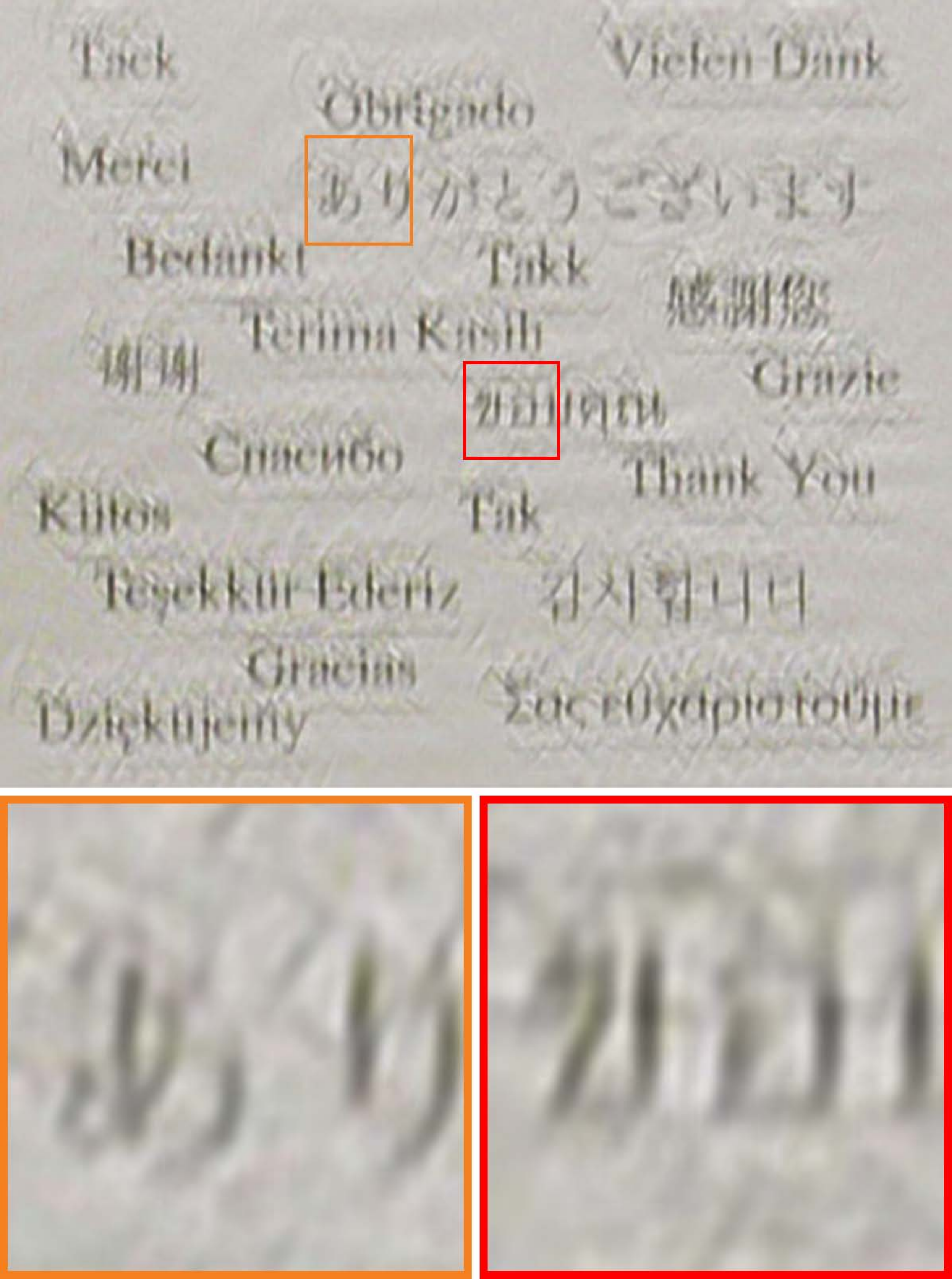}
			&\includegraphics[width=0.135\textwidth,height=0.09\textheight]{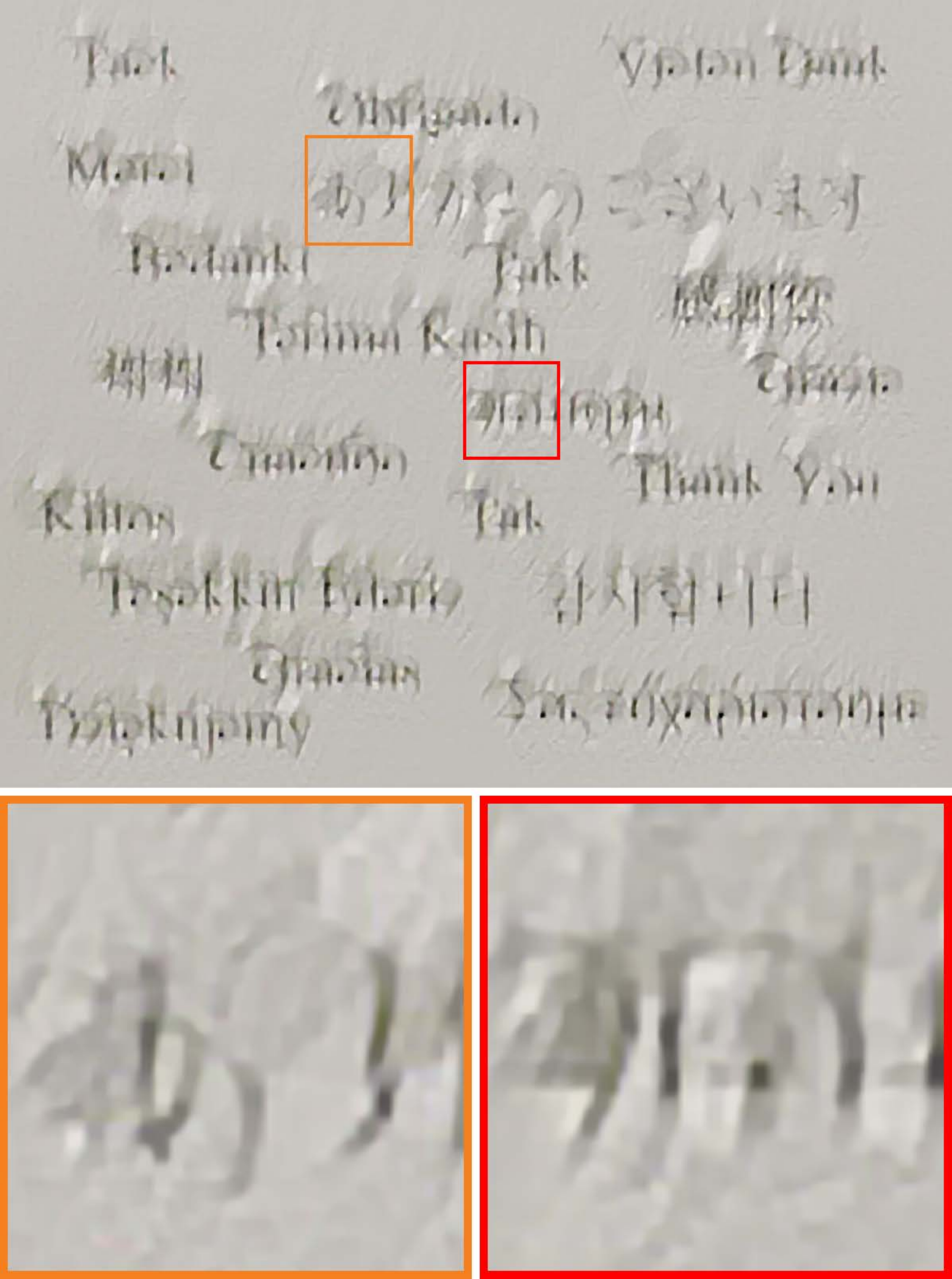}
			&\includegraphics[width=0.135\textwidth,height=0.09\textheight]{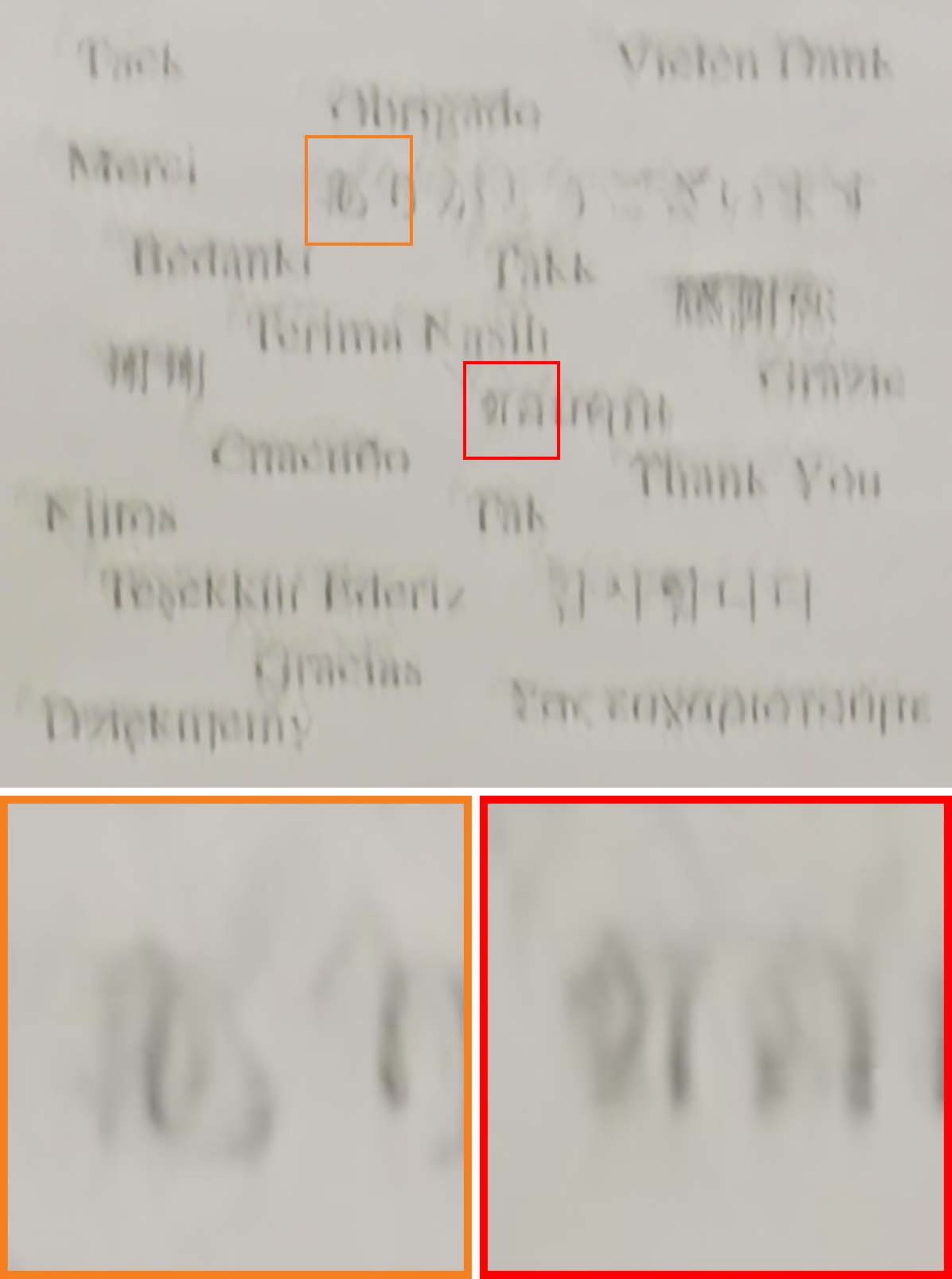}
			&\includegraphics[width=0.135\textwidth,height=0.09\textheight]{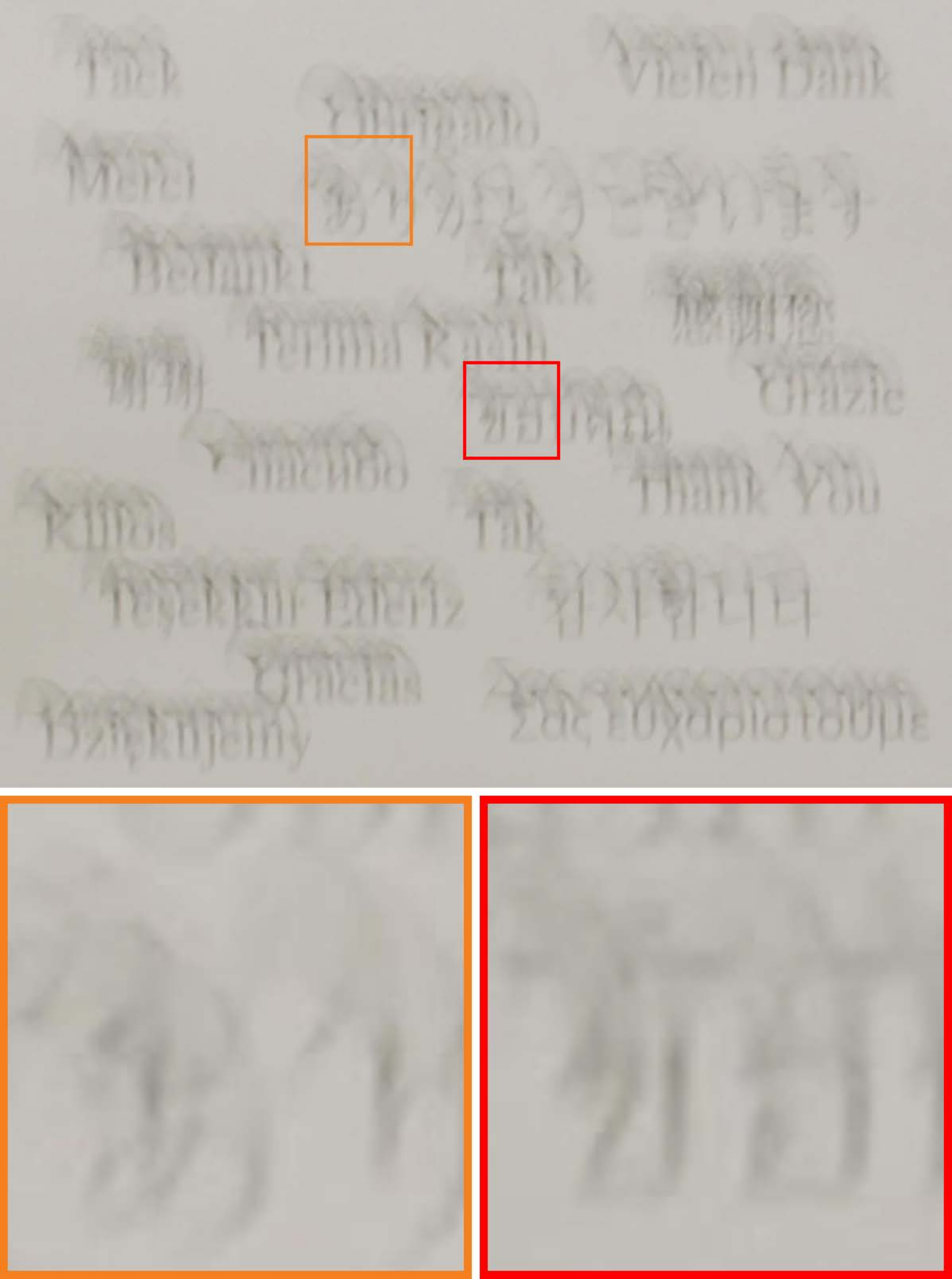}
			&\includegraphics[width=0.135\textwidth,height=0.09\textheight]{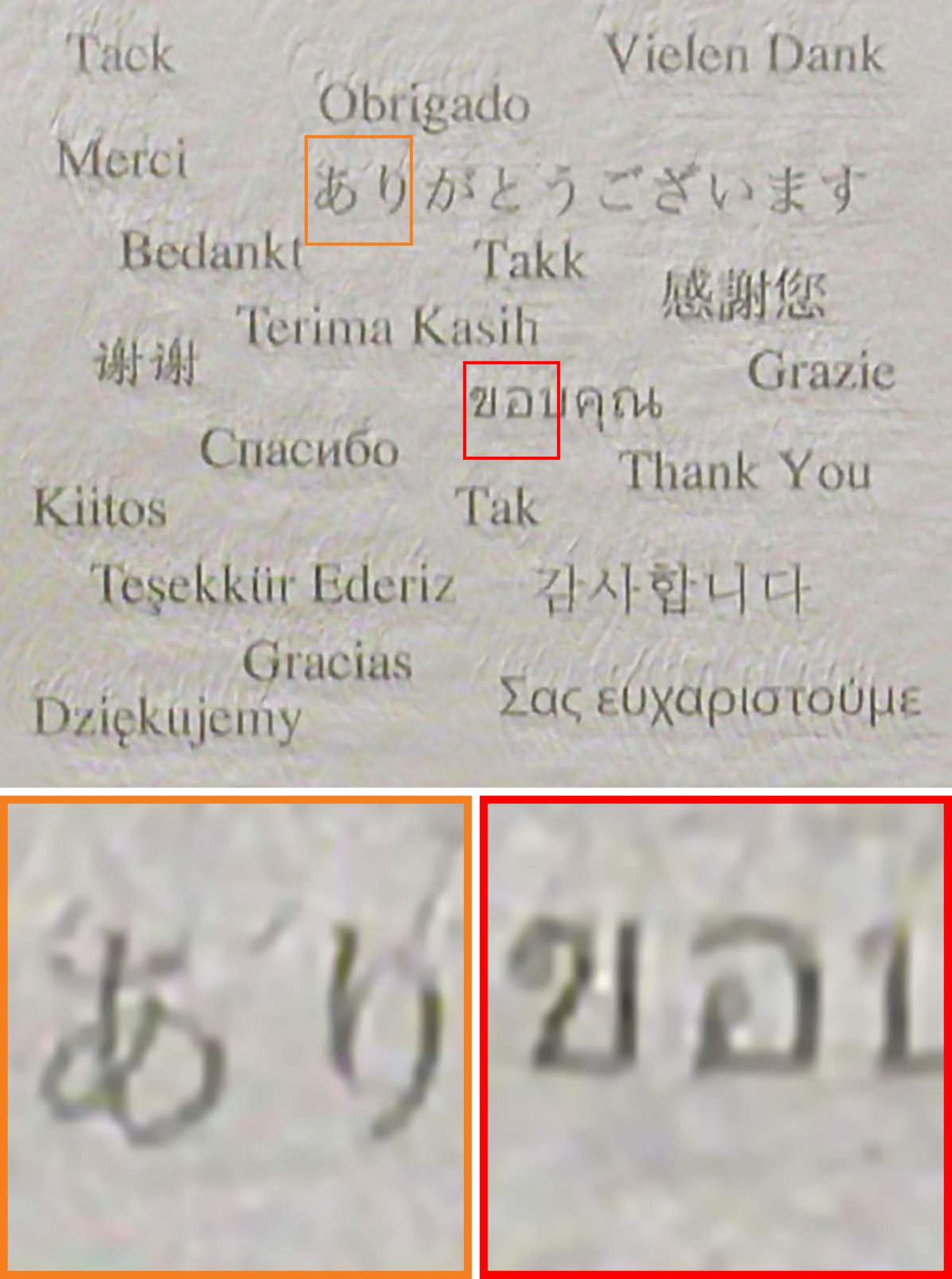}\\
			\footnotesize	(a) Input   &\footnotesize (b) DCP~\cite{pan2016blind}  & \footnotesize(c)  LMG~\cite{chen2019blind} & \footnotesize(d) PMP~\cite{wen2020simple} & \footnotesize (e)  DRUNet~\cite{Mou2022DGUNet} & \footnotesize (f)  Restomer~\cite{zamir2022restormer} & \footnotesize (g)  Ours\\
		\end{tabular}
		\caption{Image deblurring comparisons on the synthesized and real-world  blurred observations.}
		\label{fig:real-world_image}
	\end{figure*}
	
	\begin{table}
		\renewcommand{\arraystretch}{1.2}

		\caption{Average  interpolation results (PSNR/SSIM) on CBSD68~\cite{roth2009fields}.}
						\vspace{-0.2cm}
		\label{tab:data_set_result_inpainting]}
		\centering\footnotesize
		\setlength{\tabcolsep}{1.5mm}{
			\begin{tabular}{|c| c| c| c| c| c|}
				\hline
				Mask &  ~\cite{roth2009fields} &  ~\cite{he2014iterative}  & ~\cite{getreuer2012total} &~\cite{zhang2021plug} & Ours \\ \hline
				20\%  & 38.23/0.96  & 35.20/0.96 &37.55/0.97& 38.04/0.97& \textbf{38.78/0.98} \\ \hline
				
				80\%  & 27.64/0.65  & 24.92/0.70 &27.31/0.80& 26.44/0.79 & \textbf{27.77/0.82} \\ 
				\hline
			\end{tabular}	
		}
	\end{table}
	%
	%
	
	\textbf{Blind Image Deblurring:}
	First of all, we evaluated our methods with other deblurring competitors including    dark-channel prior~\cite{pan2016blind}, extreme-channel prior~\cite{yan2017image}, discriminative prior~\cite{li2018learning} and recently developed  LMG~\cite{chen2019blind} and  PMP~\cite{wen2020simple} on two synthetic datasets, \textit{i.e.,} Levin \emph{et al.}'s dataset~\cite{levin2009understanding}  and Sun \emph{et al.}'s dataset~\cite{sun2013edge}. 
To compare other blind deblurring approaches fairly, we introduced the non-blind deblurring method~\cite{zoran2011learning} to estimate  final latent images.
In  Table~\ref{tab:data_set_results}, GDC advanced state-of-the-art significantly on Levin \emph{et al.}'s dataset and cost the shortest running time notably. Meanwhile, the proposed scheme had the best PSNR result on the challenging  Sun \emph{et al.} ~\cite{sun2013edge}'s dataset, but it also consumed a much longer
time because of the inference of DM. We also report the numerical comparison with the latest learning-based schemes~\cite{Mou2022DGUNet,ji2022xydeblur,zamir2022restormer,Wang_2022_CVPR} in Table.~\ref{tab:dataset_sun} and obtain the consistent performance.
In Fig.~\ref{fig:real-world_image}, we depicted the visual comparisons of our schemes together with the DCP~\cite{pan2016blind}, LMG~\cite{chen2019blind} and  PMP~\cite{wen2020simple}, and   two learning-based methods  DRUNet~\cite{Mou2022DGUNet} and  Restomer~\cite{zamir2022restormer}. Levin \emph{et al.}'s dataset is a small dataset, only with 32 images. Noting that all schemes are not trained on this dataset. Moreover, the compared methods are all learning-based schemes. Without the estimation procedure of blur kernel estimation and prior correction, these methods are fragile facing the large blur kernels. 
Obviously, DCP  could not estimate an accurate kernel on this synthetic blurry image. Learning-based methods cannot address these difficult scenarios degraded by large kernel size.
GDC scheme could generate more visual-pleasant content and clear structure. Moreover, we validated the advantage of our schemes on real-world blurry images in the last row Fig.~\ref{fig:real-world_image}.    Obviously, GDC generated more contextual details and recovered better structural patterns.
	\begin{figure}
		\centering \begin{tabular}{c@{\extracolsep{0.1em}}c@{\extracolsep{0.1em}}c@{\extracolsep{0.1em}}c@{\extracolsep{0.1em}}c}
			\includegraphics[width=0.09\textwidth,]{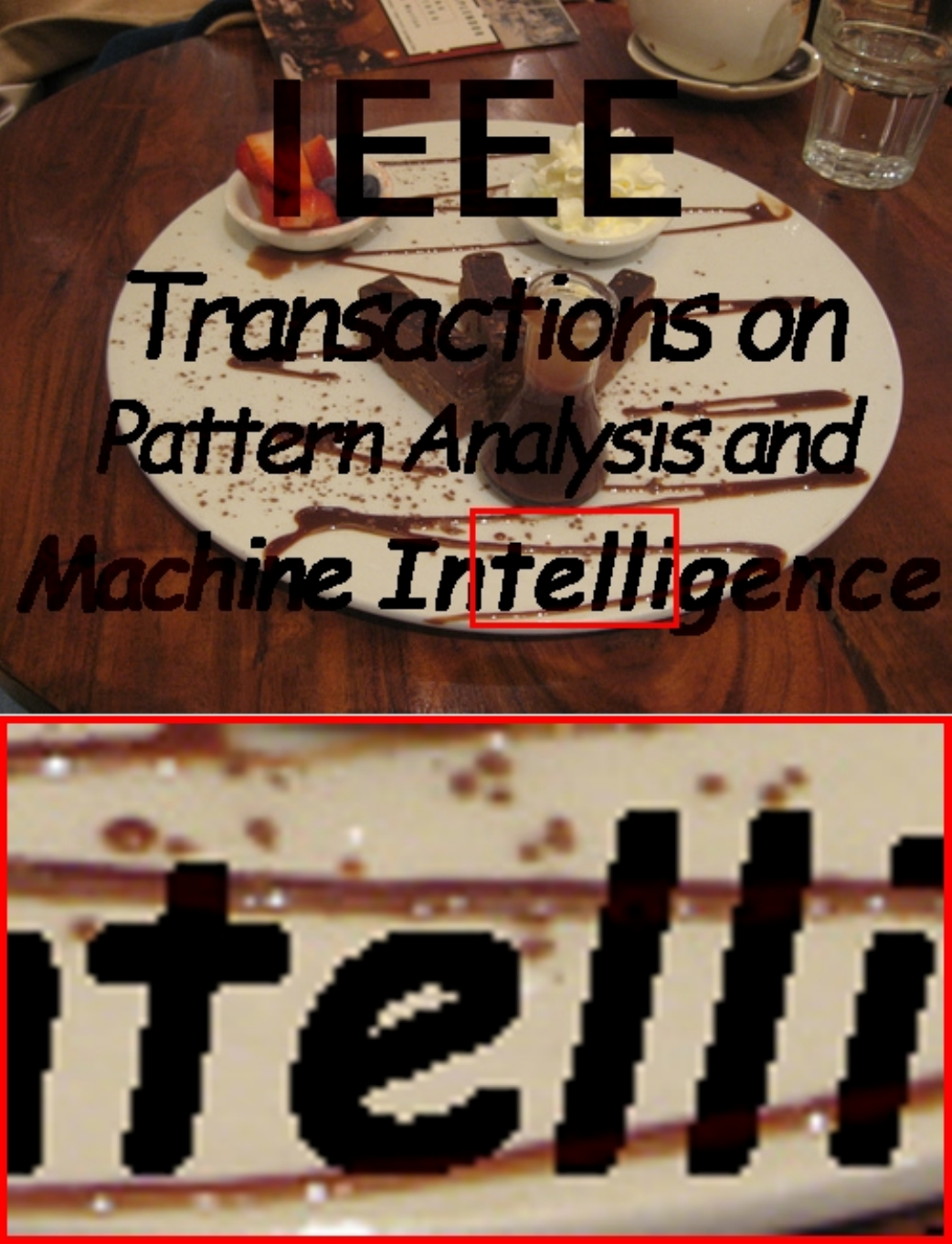}
			&\includegraphics[width=0.09\textwidth,]{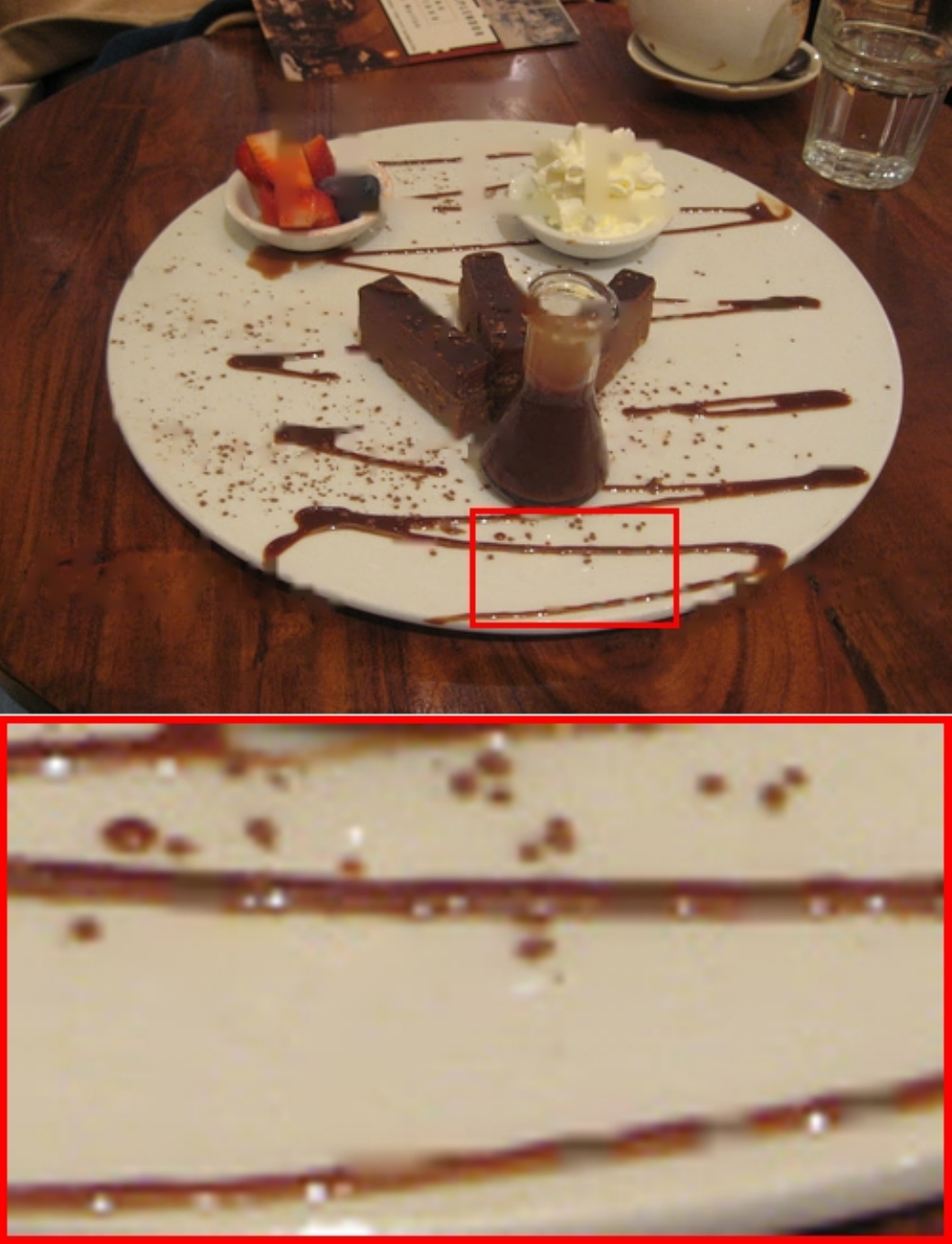}
			&\includegraphics[width=0.09\textwidth,]{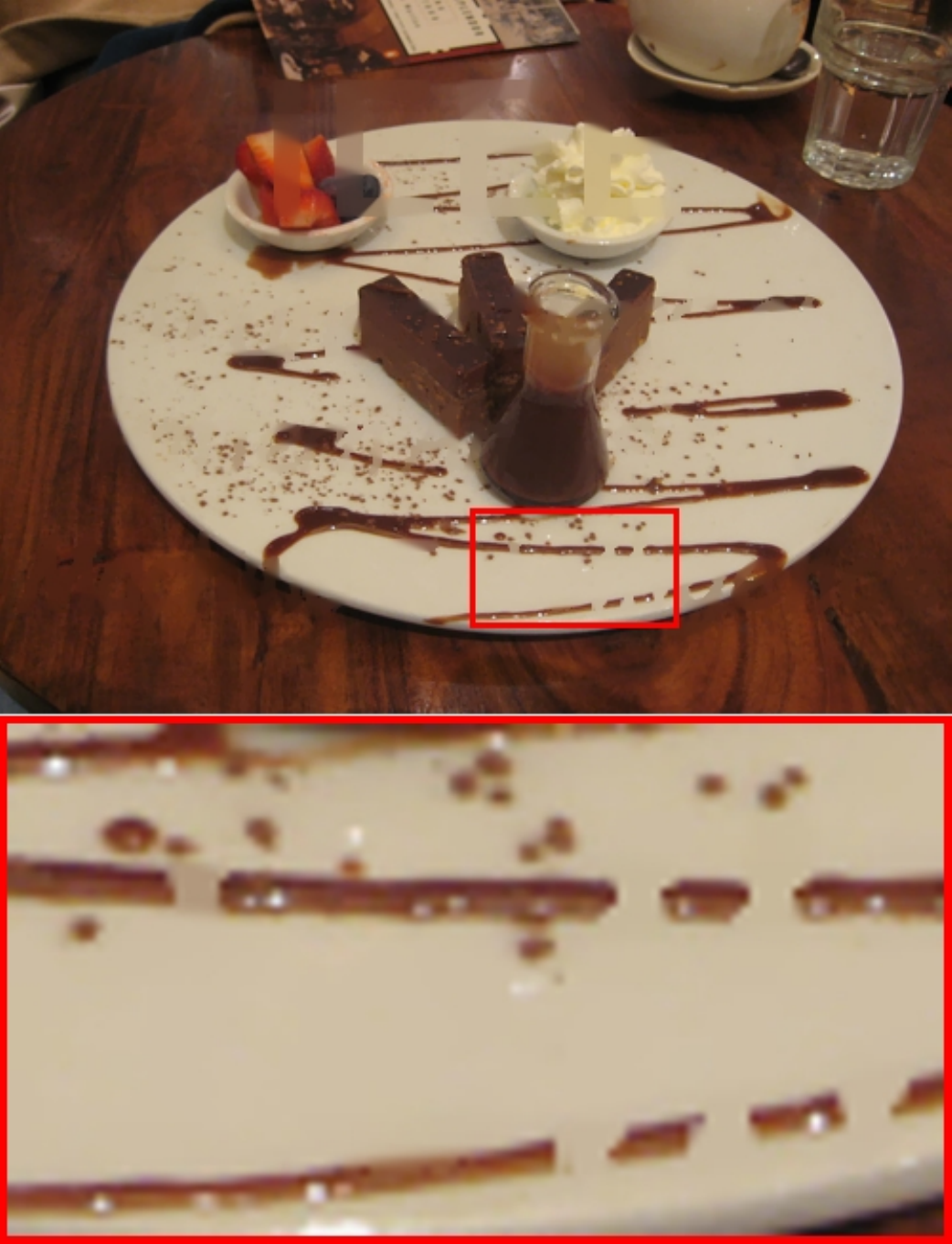}
			&\includegraphics[width=0.09\textwidth,]{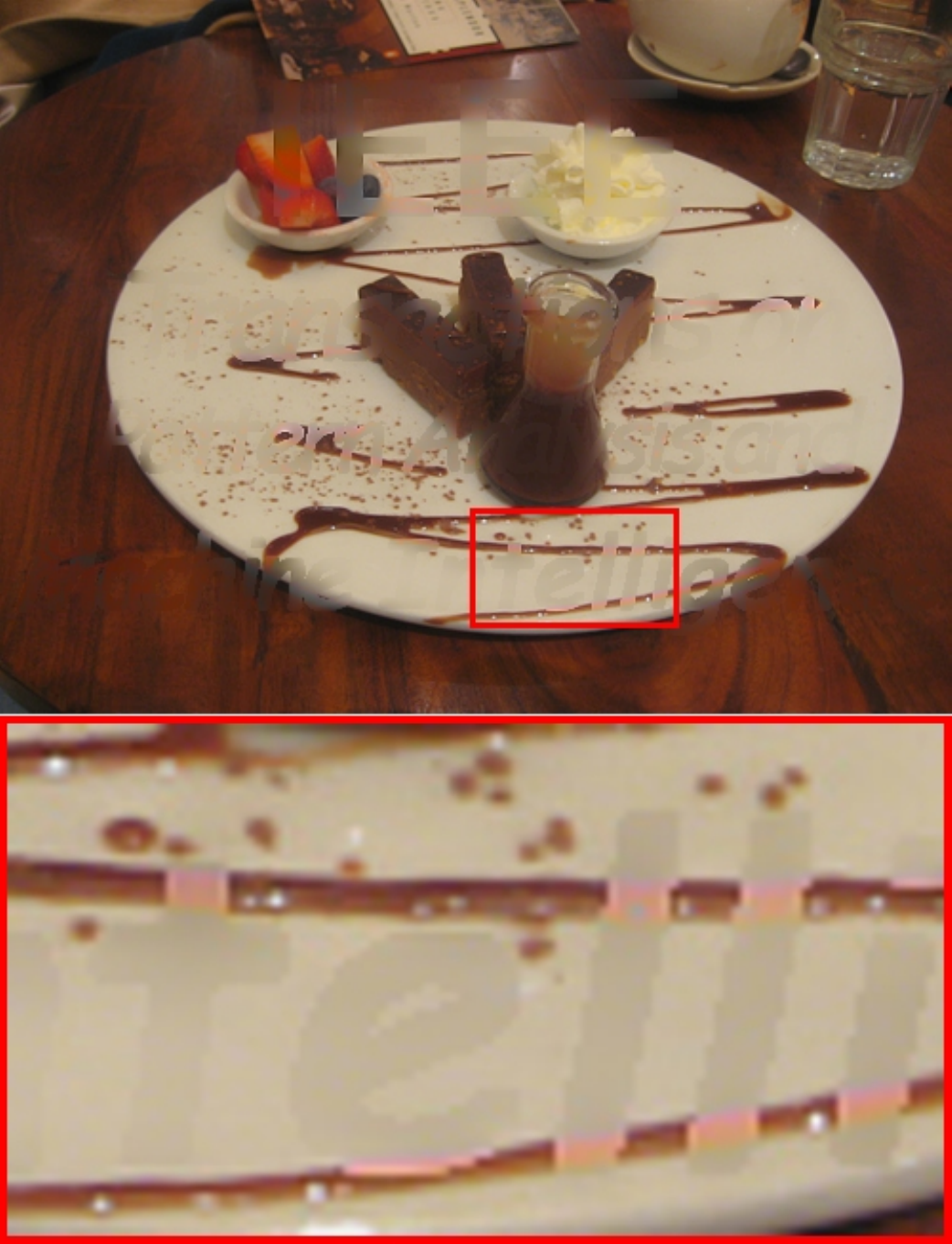}
			
			&\includegraphics[width=0.09\textwidth,]{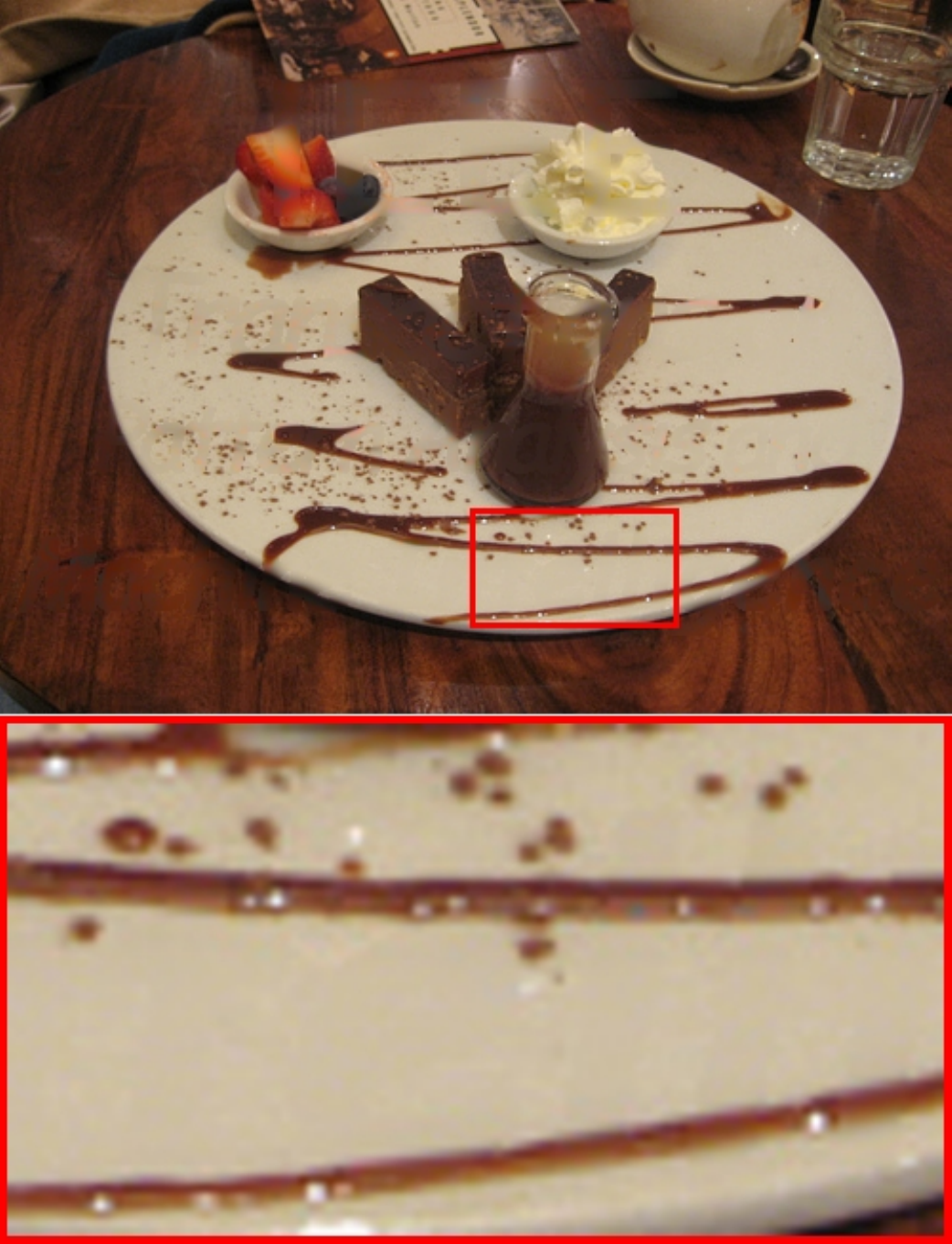}\\
			\footnotesize	PSNR/SSIM  & \footnotesize 30.82/0.98 & \footnotesize 29.53/0.97 &
			\footnotesize	26.19/0.96
			& \footnotesize \textbf{31.45/0.99} \\
			\includegraphics[width=0.09\textwidth,]{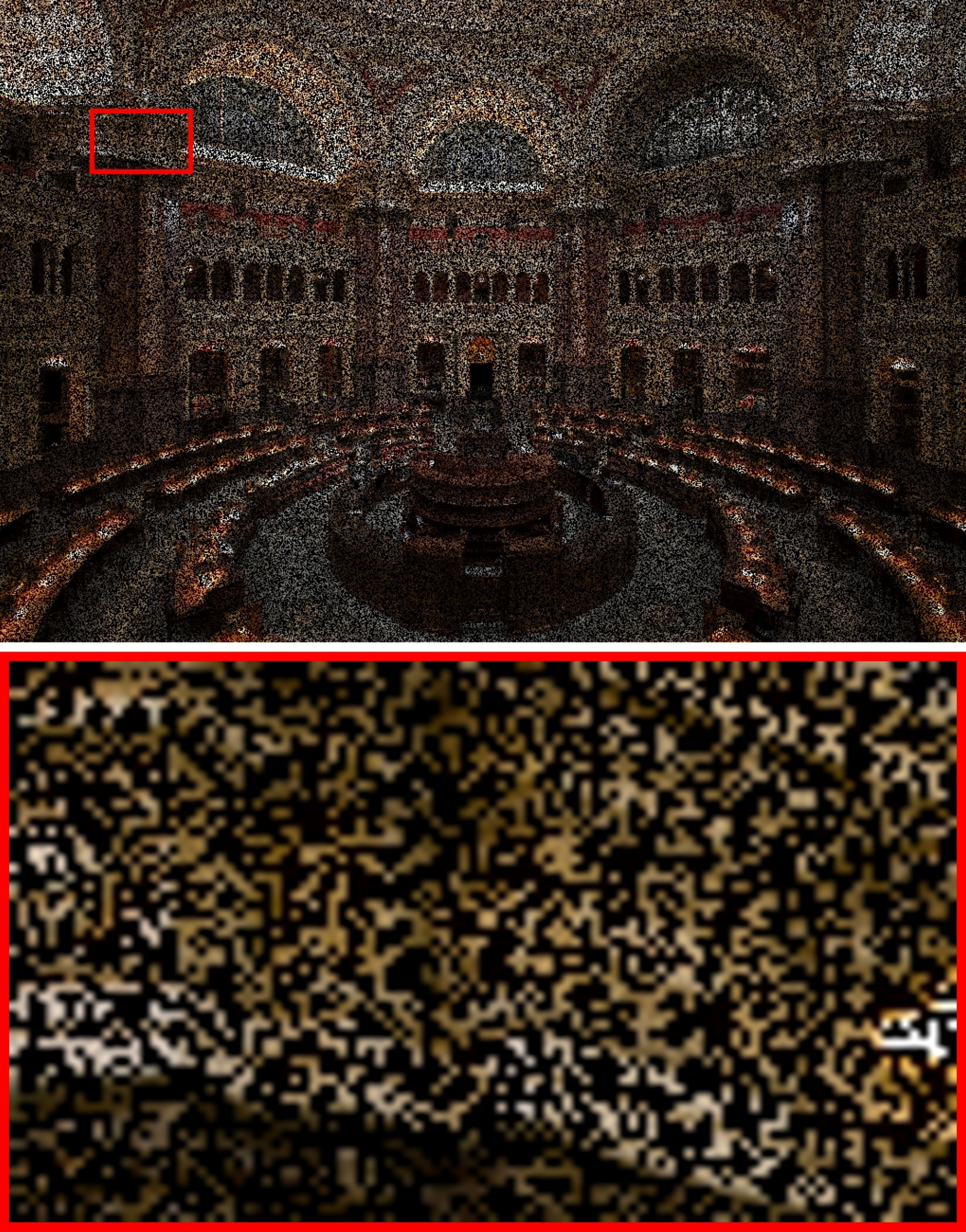}

			&\includegraphics[width=0.09\textwidth,]{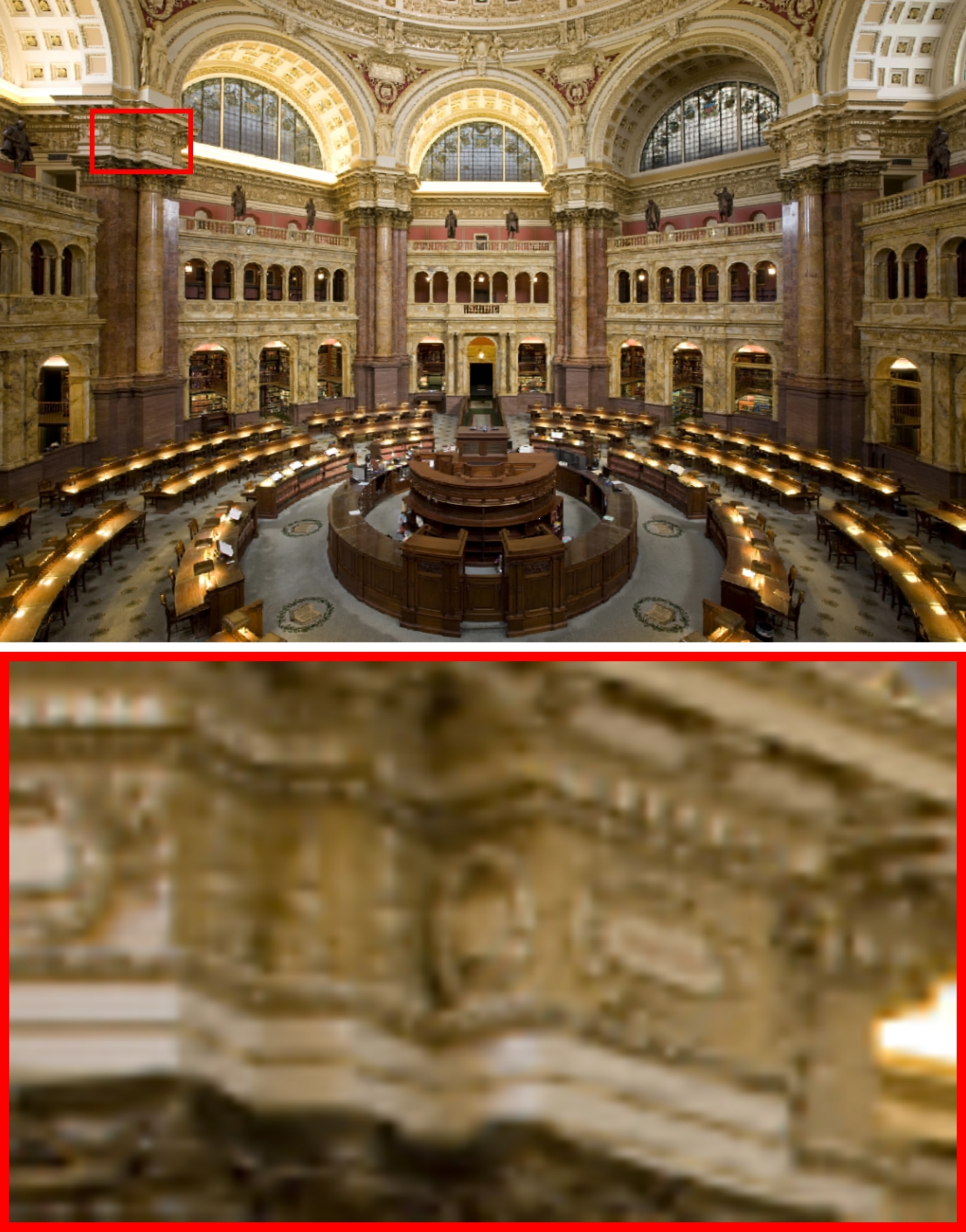}
			&\includegraphics[width=0.09\textwidth,]{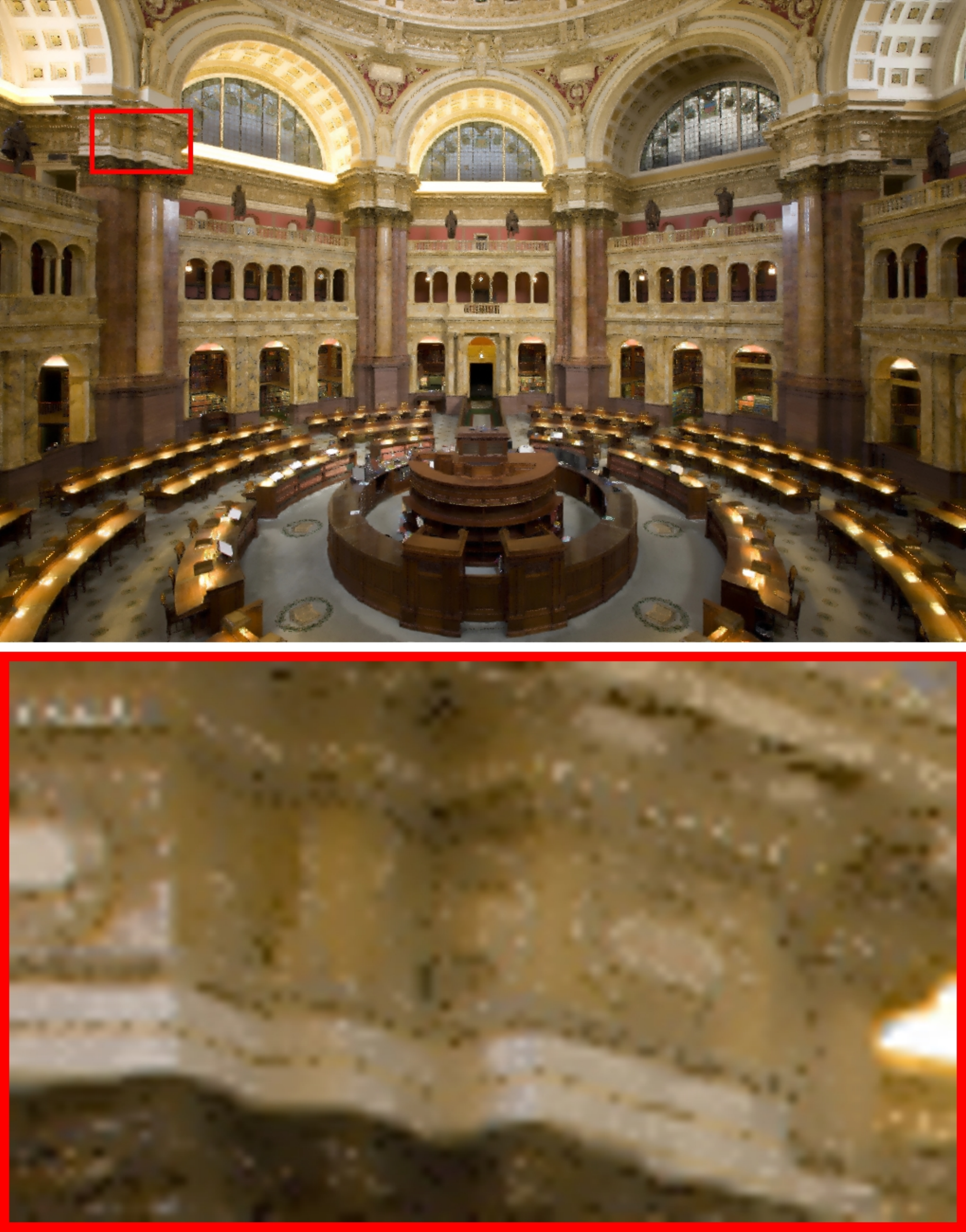}
			&\includegraphics[width=0.09\textwidth,]{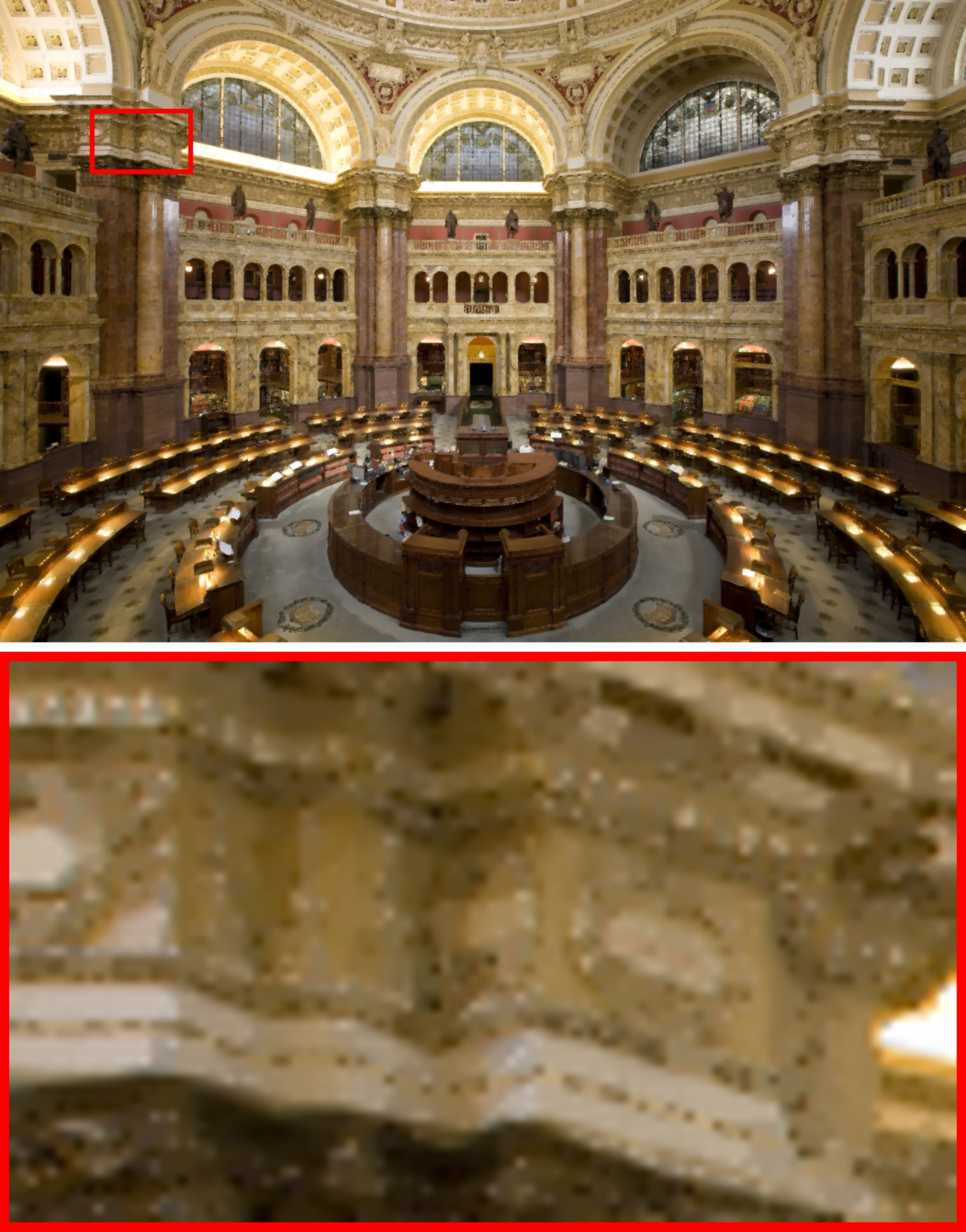}
			%
			%
			&\includegraphics[width=0.09\textwidth,]{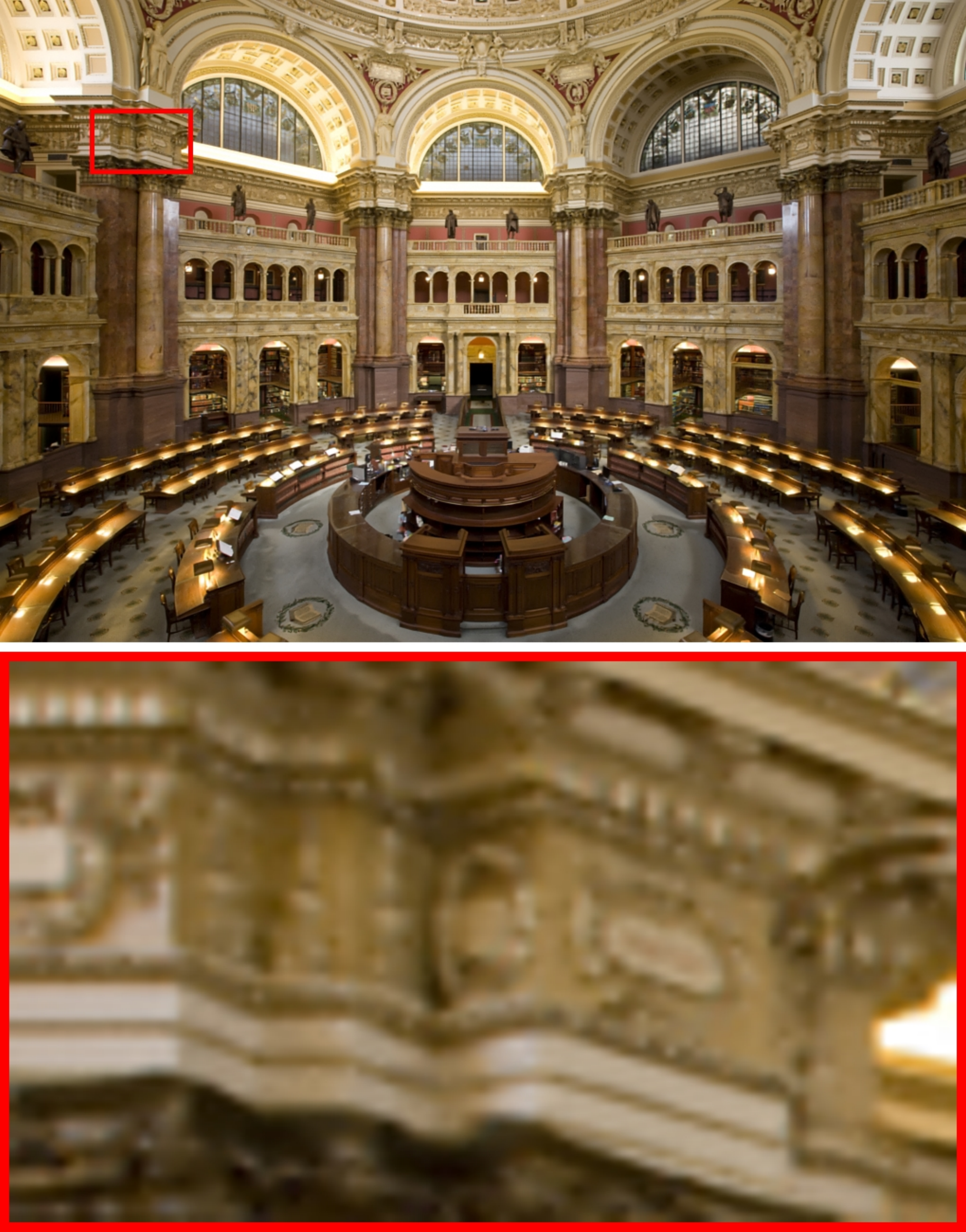}\\
			\footnotesize	PSNR/SSIM  & \footnotesize29.69/0.97& \footnotesize 26.21/0.93 & \footnotesize 27.13/0.94   & \footnotesize \textbf{30.68/0.97} \\
			\footnotesize	(a) Input & \footnotesize (b) FoE~\cite{roth2009fields} & \footnotesize (c) ISD~\cite{he2014iterative} & \footnotesize (d) TV~\cite{getreuer2012total}  & \footnotesize (e)  Ours\\	
		\end{tabular}
		\caption{Interpretation comparisons with one mask (\textit{i.e.,}  $\mathbf{60\%}$  pixels missing). Quantitative metrics are listed below each image.}
		\label{fig:inpaintting result}
	\end{figure}
	
	\begin{figure}
		\centering \begin{tabular}{c@{\extracolsep{0.2em}}c@{\extracolsep{0.2em}}c@{\extracolsep{0.2em}}c@{\extracolsep{0.2em}}c@{\extracolsep{0.2em}}c}
			\includegraphics[width=0.15\textwidth,height=0.07\textheight]{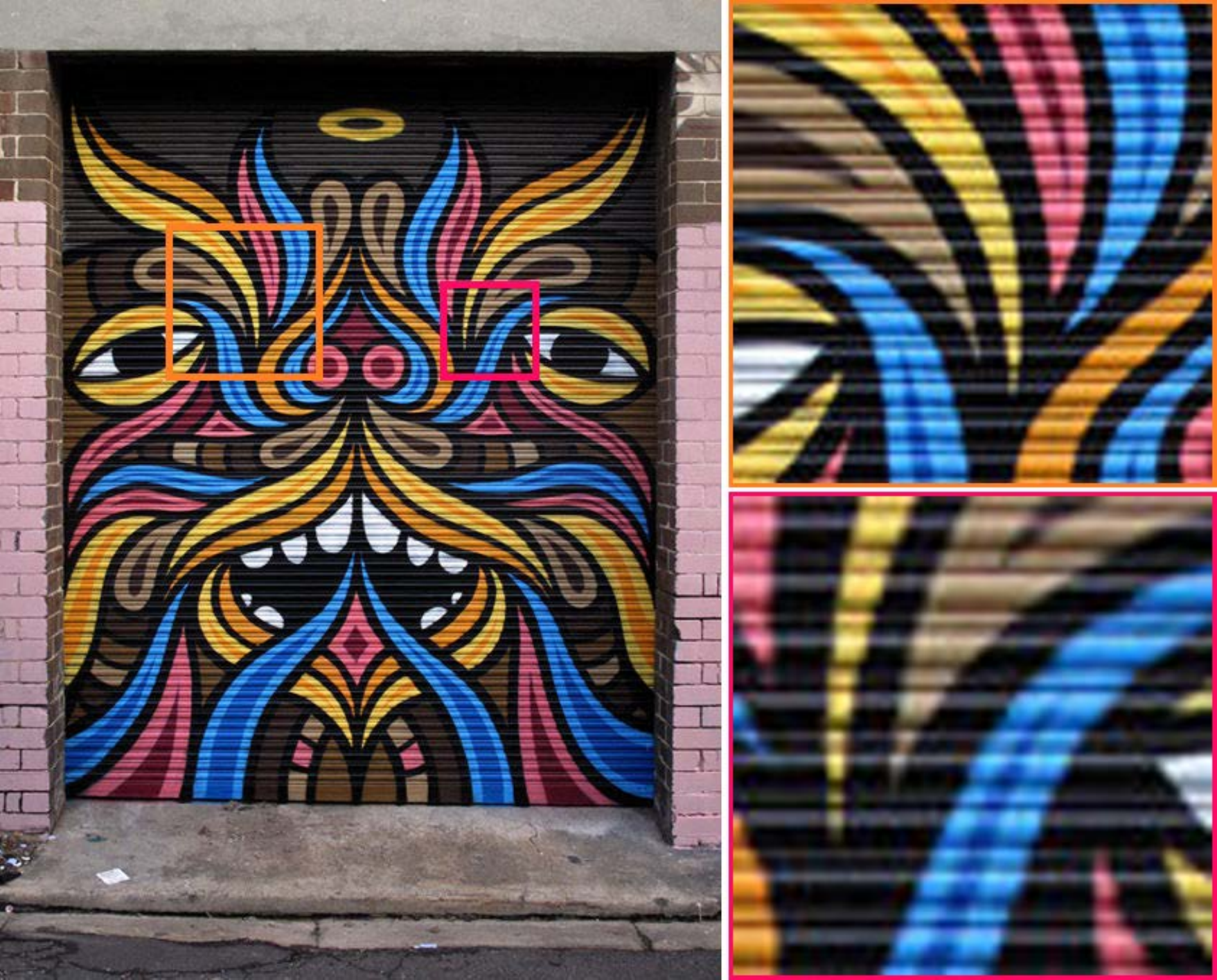}
			&\includegraphics[width=0.15\textwidth,height=0.07\textheight]{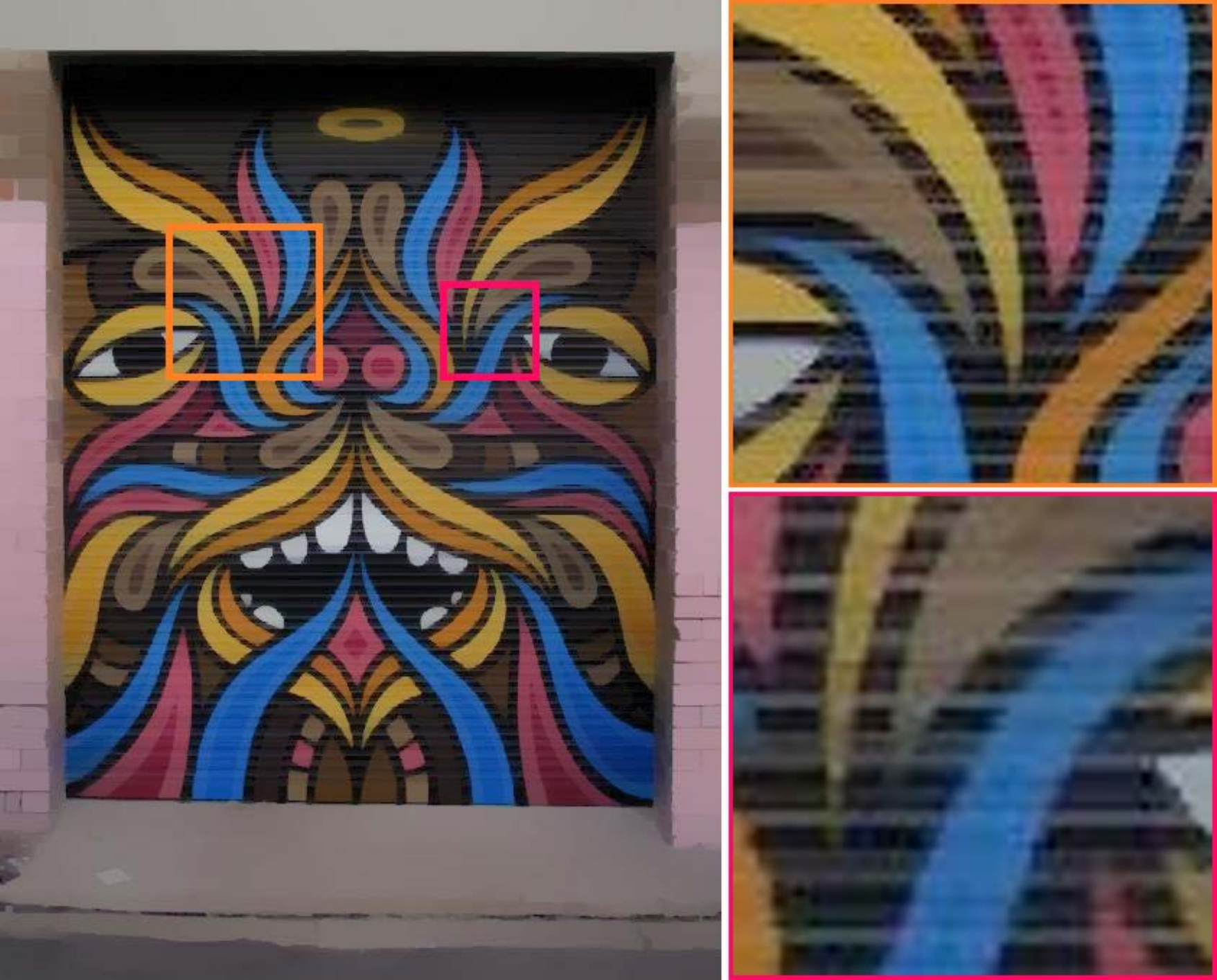}
			&\includegraphics[width=0.15\textwidth,height=0.07\textheight]{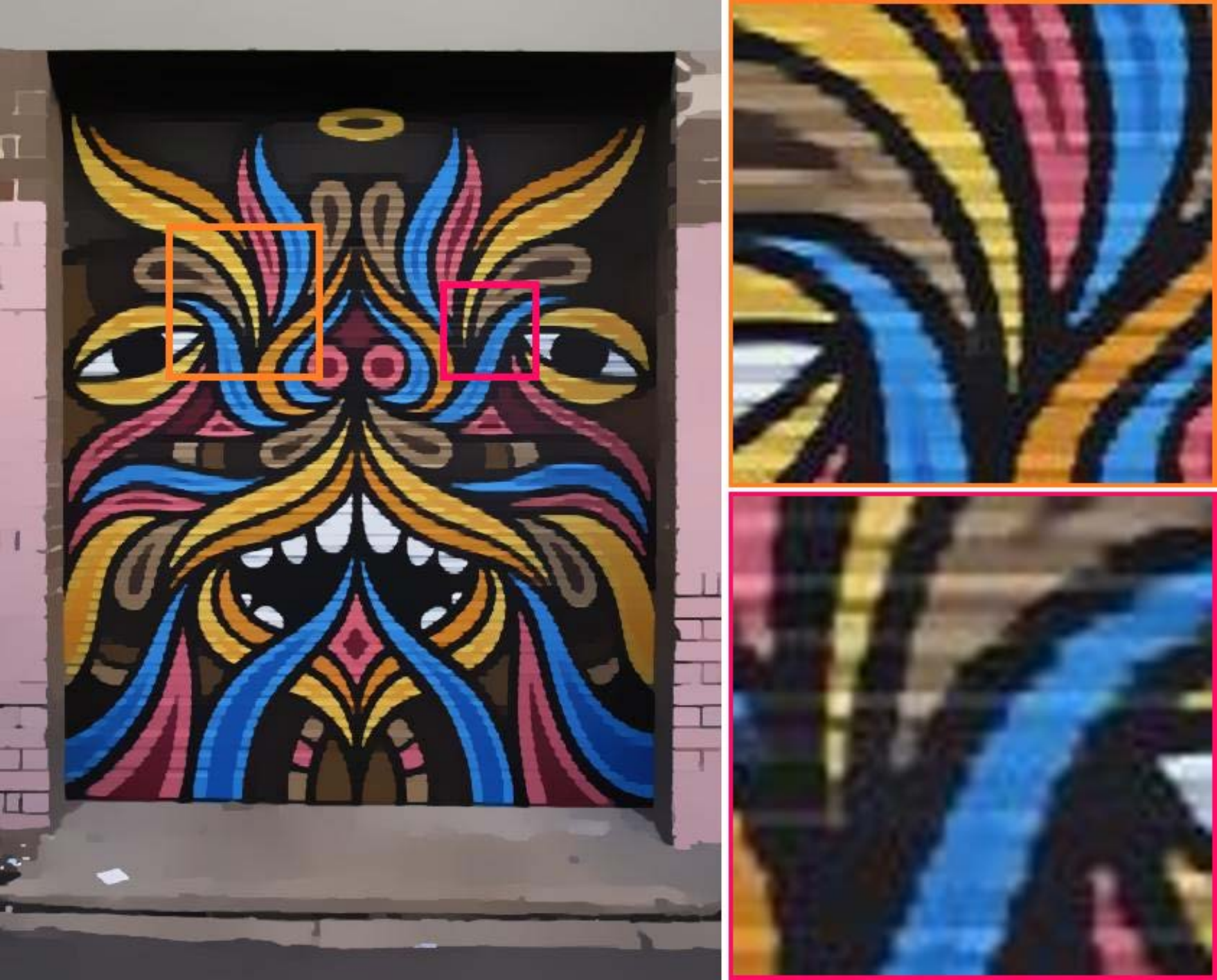}\\
			\footnotesize	(a) Input & \footnotesize(b) WLS~\cite{farbman2008edge} & \footnotesize(c) RTV~\cite{tsmoothing2012} \\
			\includegraphics[width=0.15\textwidth,height=0.07\textheight]{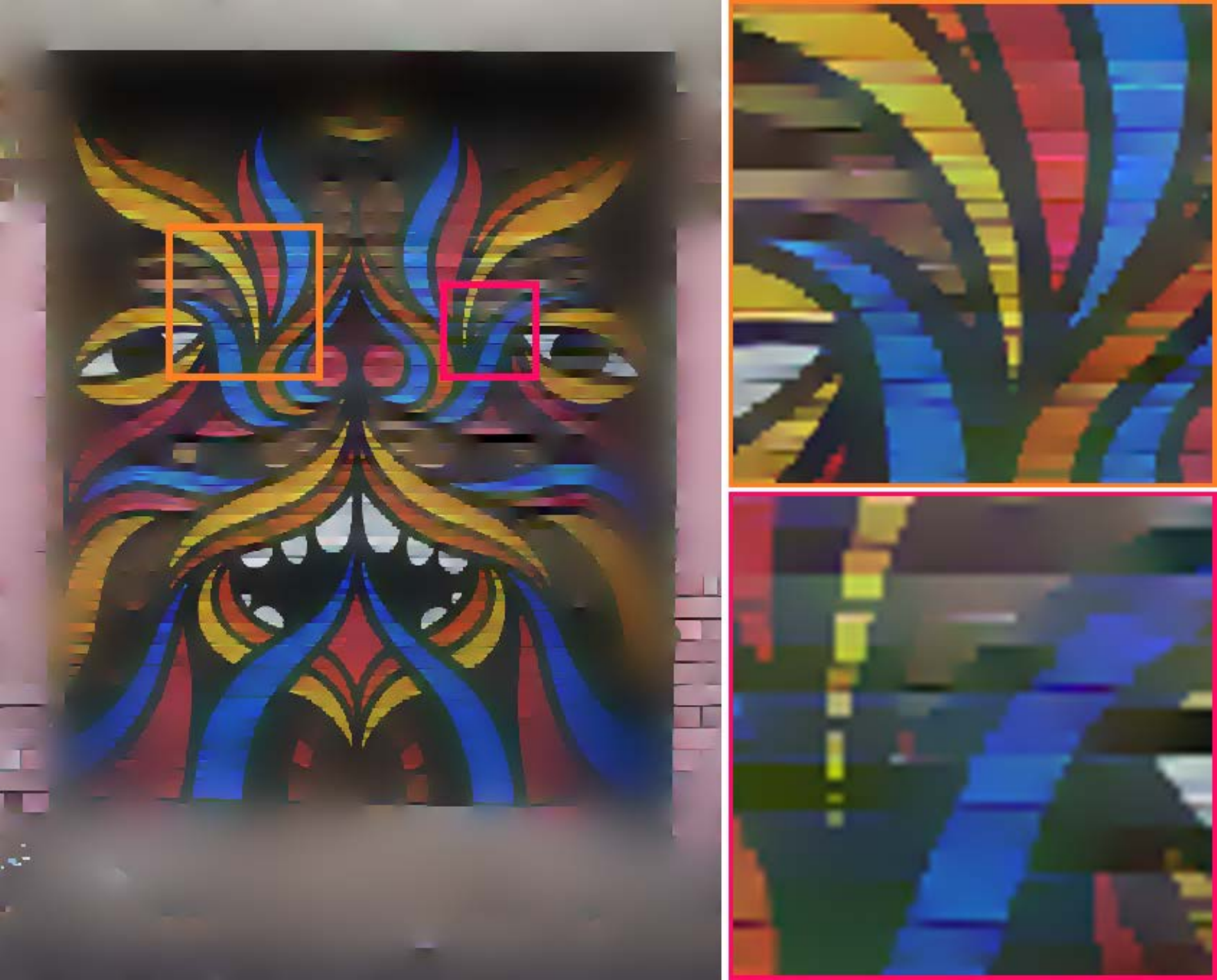}
			&\includegraphics[width=0.15\textwidth,height=0.07\textheight]{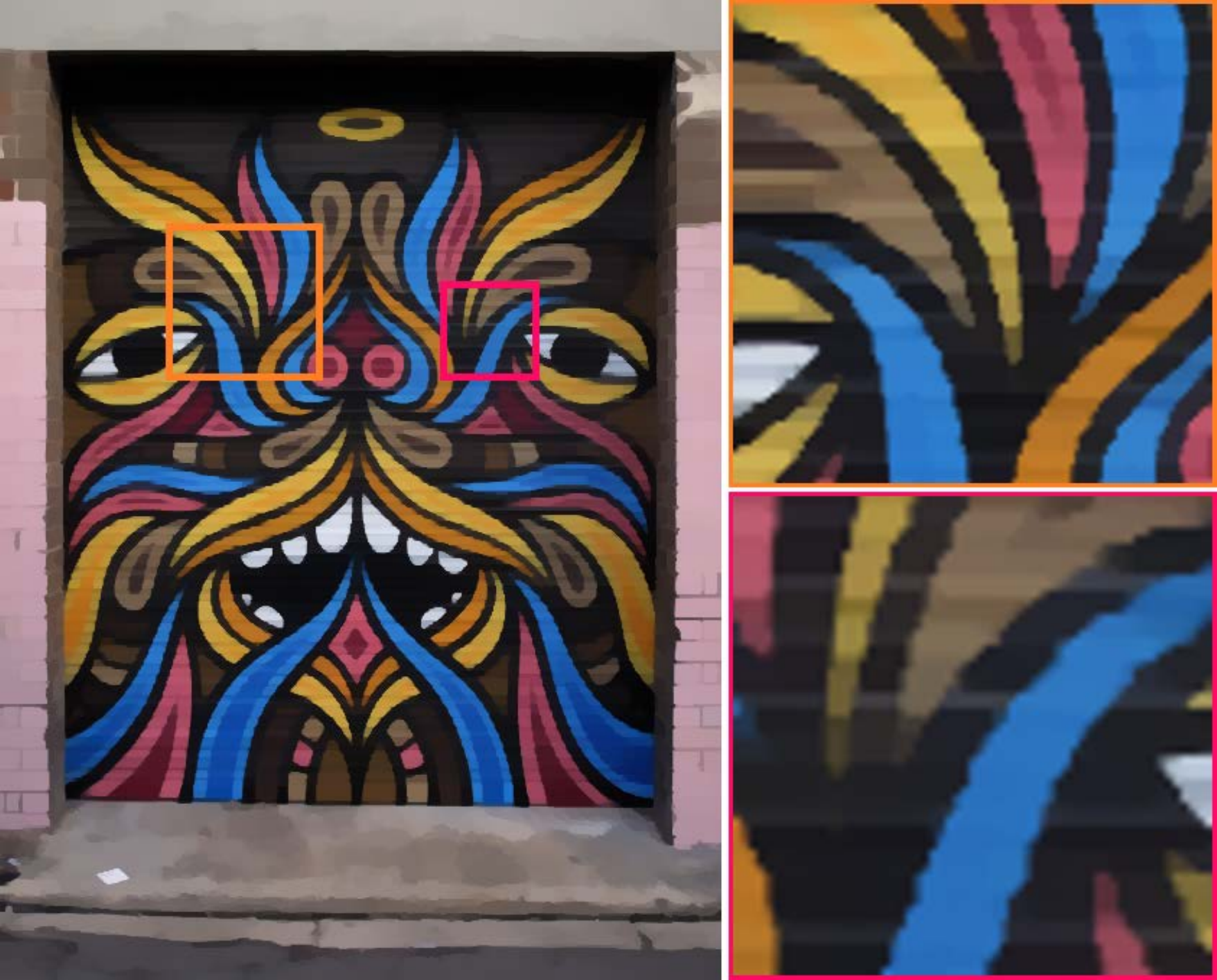}
			&\includegraphics[width=0.15\textwidth,height=0.07\textheight]{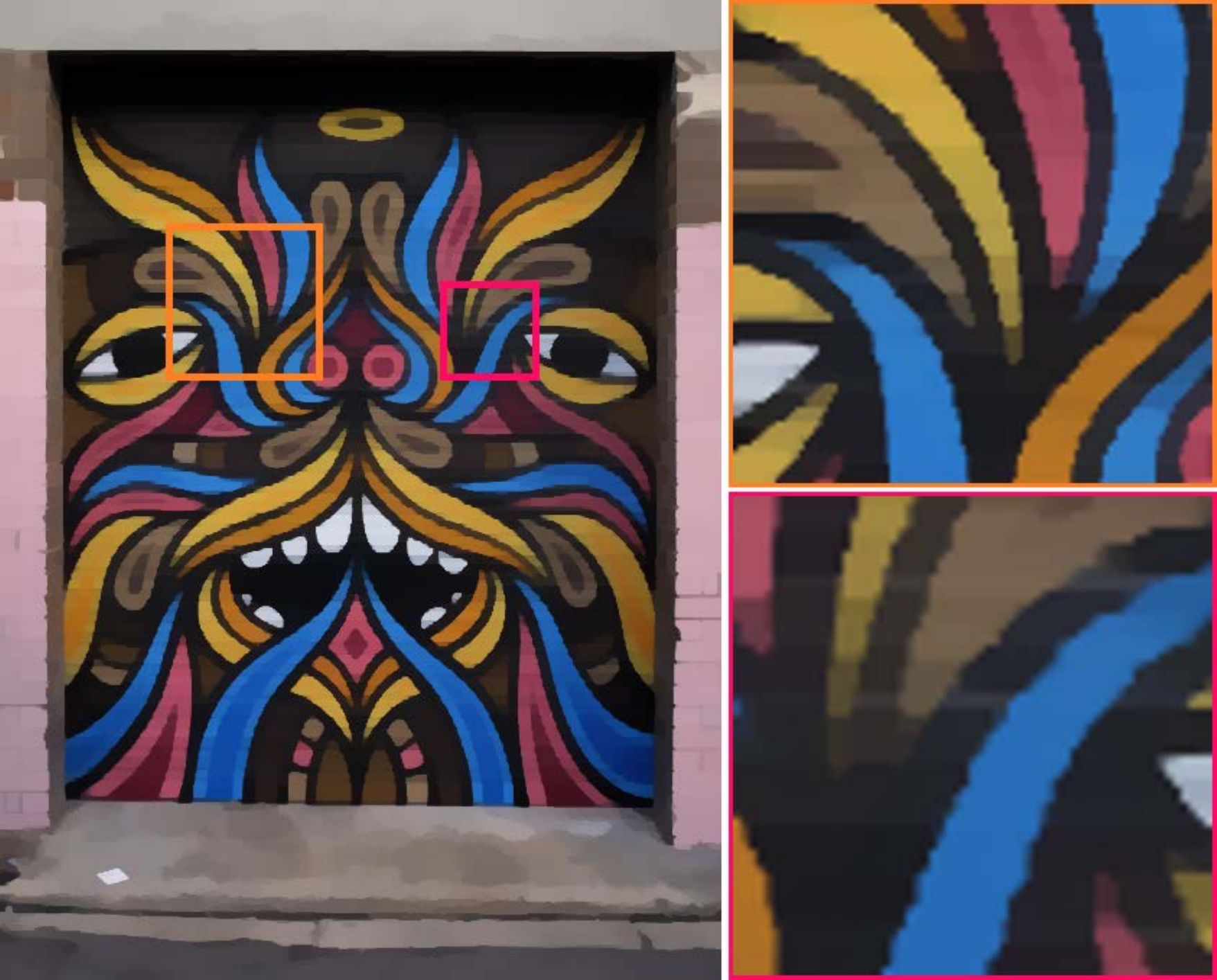}\\
			\footnotesize	(d) ILS~\cite{liu2020real}  & \footnotesize(e) GSF~\cite{liu2021generalized} & \footnotesize(f) Ours \\	
		\end{tabular}
		\caption{Smoothing comparisons on an example with horizontal textures.}
		\label{fig:smoothing result}
	\end{figure}
	
	\textbf{Image Interpolation:}
	To analyze the superiority of our scheme, we compared GDC against other state-of-the-art competing methods including  FoE~\cite{roth2009fields}, ISD~\cite{he2014iterative}, TV~\cite{getreuer2012total} and DPIR~\cite{zhang2021plug} on two kinds of masks (random masks of pixel missing). 
We reported the average quantitative results on the CBSD68 dataset~\cite{roth2009fields} in Table~\ref{tab:data_set_result_inpainting]}. We conducted the experiment under 20\% and 80\% of pixels missing. GDC scheme achieved the best scores under diverse missing rates.   In Fig.~\ref{fig:inpaintting result}, our GDC achieved better visual and quantitative results compared with other interpolation methods. 
It is obvious that TV~\cite{getreuer2012total} generated more pixel-missing regions. ISD~\cite{he2014iterative} and FoE~\cite{roth2009fields}  restored the most regions of images. However, these methods yielded much blurry results.

		\textbf{Edge-preserved Smoothing:}
		Edge-preserved smoothing involves two parts, \textit{i.e.,} the undesired textual removal and the structure preservation. We conducted this comparison with state-of-the-art competitors, such as   WLS~\cite{farbman2008edge}, RTV~\cite{tsmoothing2012} and recent proposed ILS~\cite{liu2020real} and
GSF~\cite{liu2021generalized}.
In Fig.~\ref{fig:smoothing result}, we depicted their visual results with a representative example, which is a rolling door with horizontal  textures.  Furthermore, 
RTV~\cite{tsmoothing2012} kept the boundaries considerably better, though the result also has some trivial  lines. 
Our  scheme removed the most undesired artifacts effectively and obtained a smoother visual performance.

\begin{figure*}
		\centering \begin{tabular}{c@{\extracolsep{0.2em}}c@{\extracolsep{0.2em}}c@{\extracolsep{0.2em}}c@{\extracolsep{0.2em}}c@{\extracolsep{0.2em}}c@{\extracolsep{0.2em}}c}
			\includegraphics[width=0.135\textwidth,]{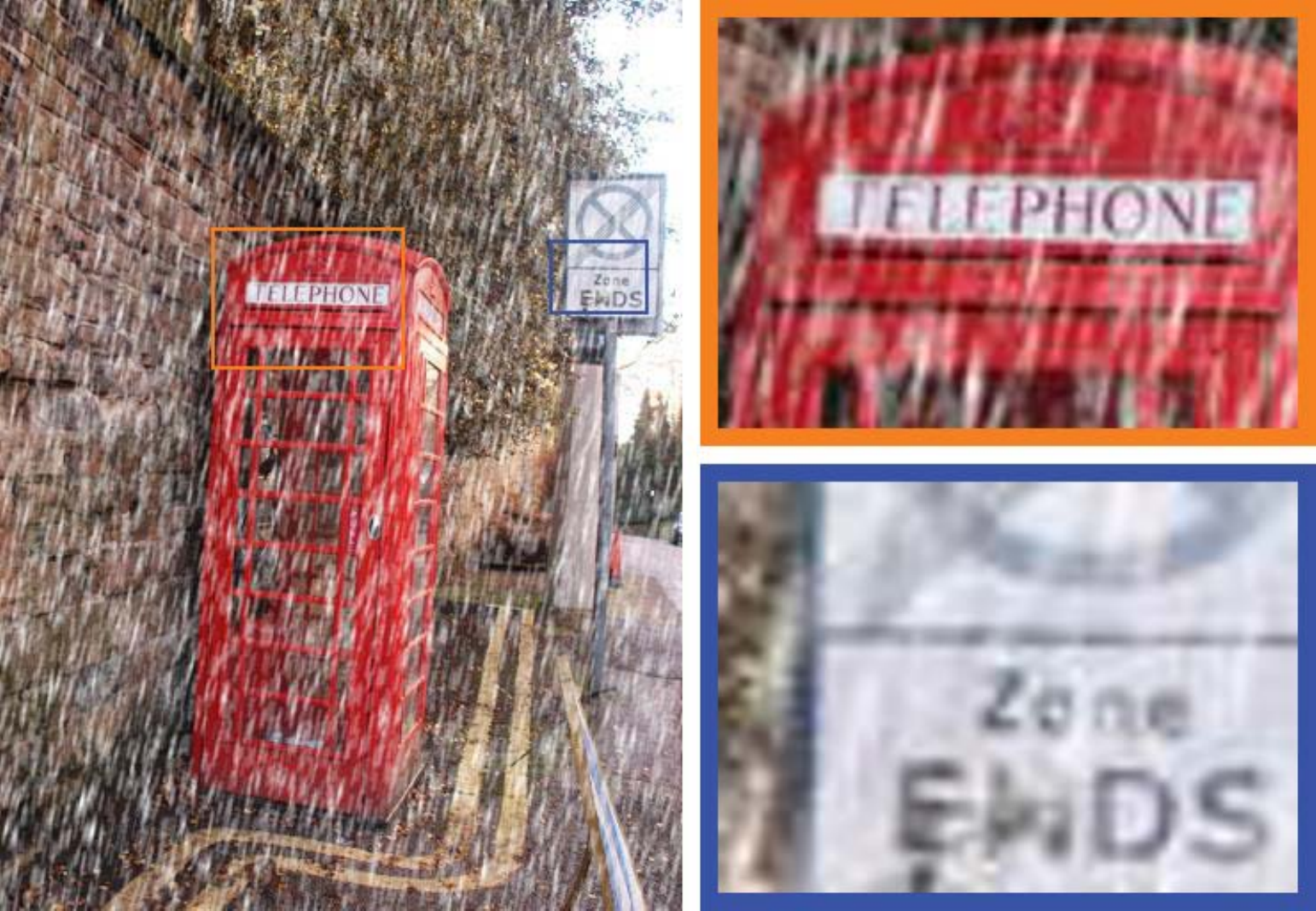}
			&\includegraphics[width=0.135\textwidth,]{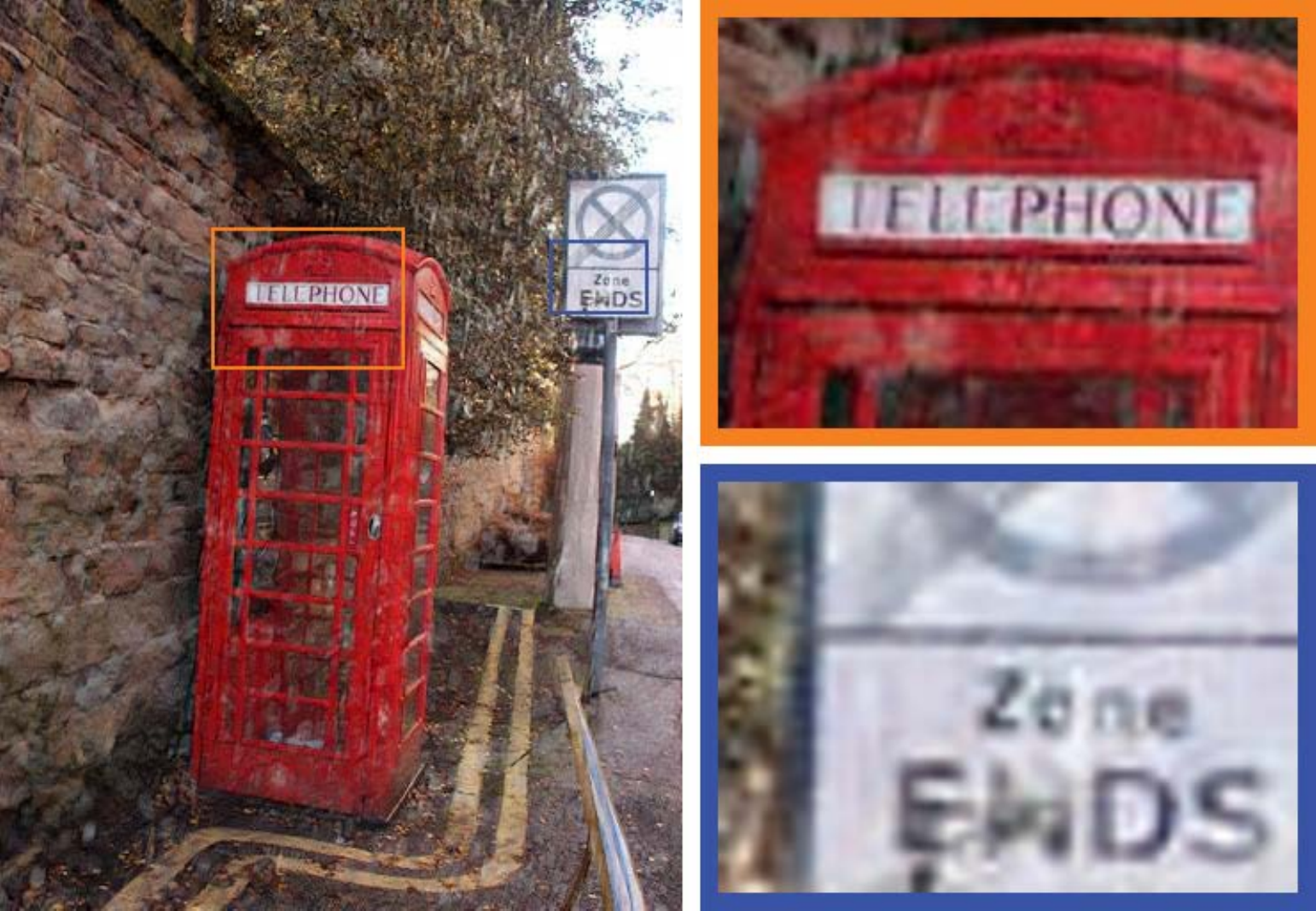}
			&\includegraphics[width=0.135\textwidth,]{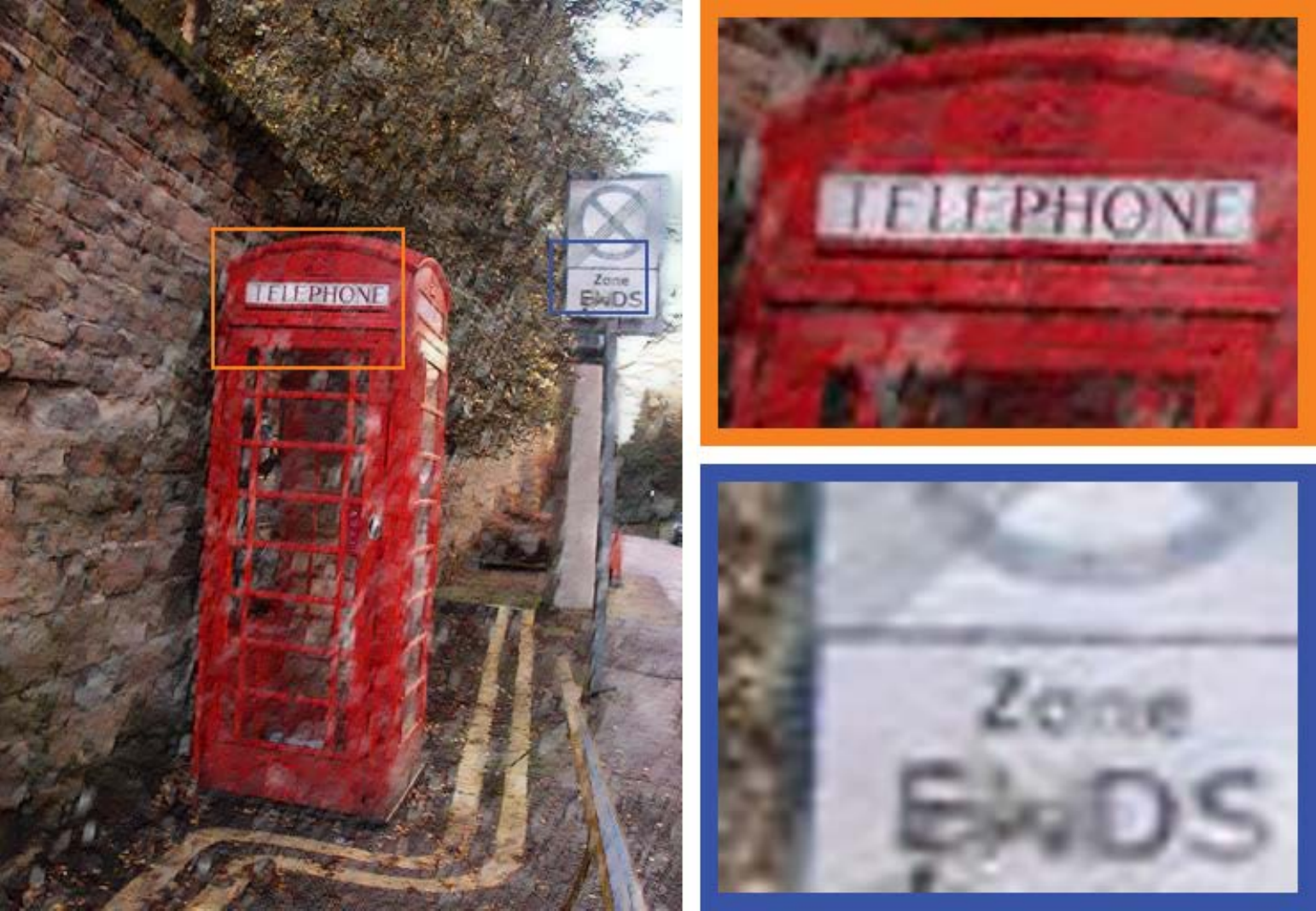}
			&\includegraphics[width=0.135\textwidth,]{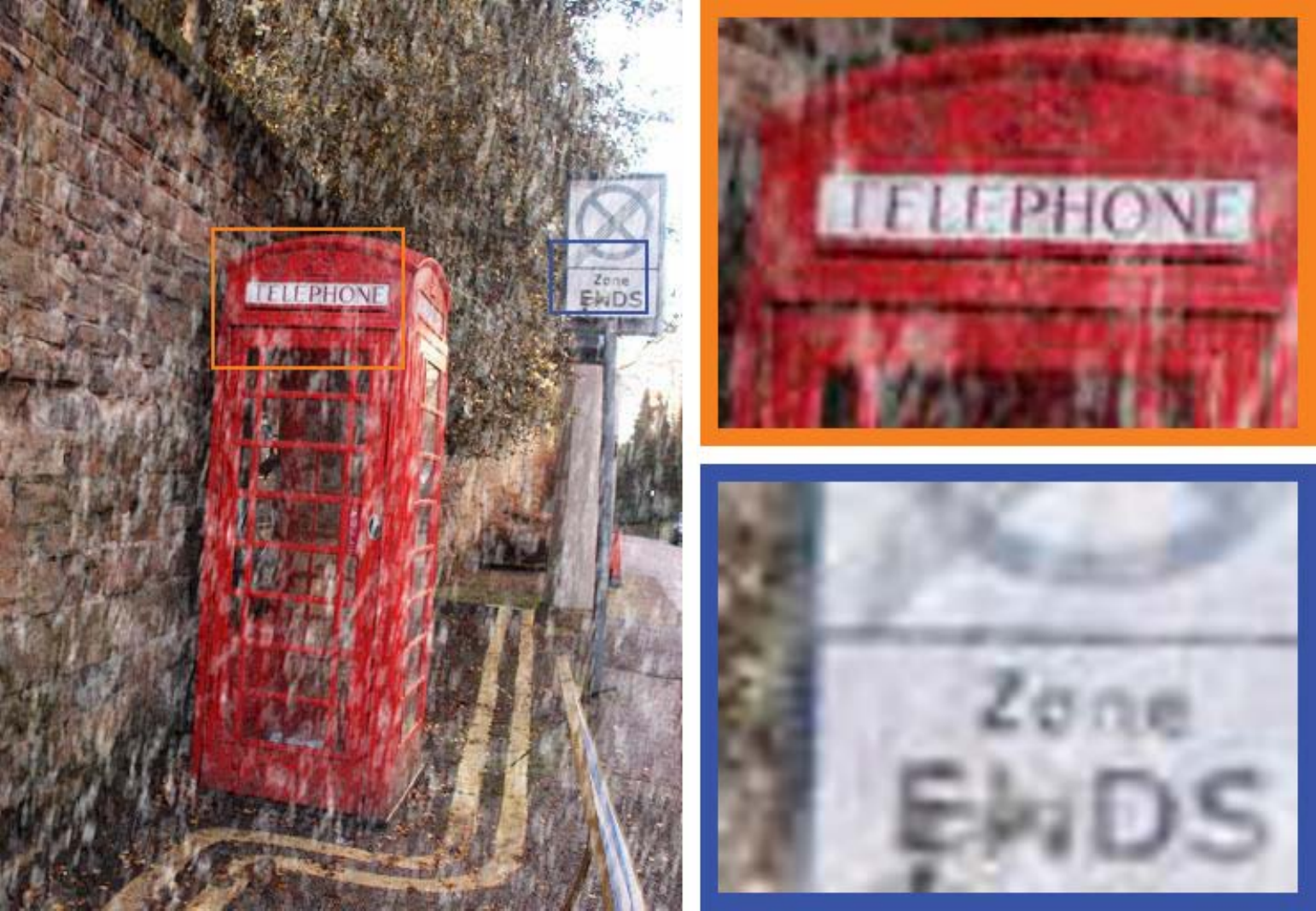}
			&\includegraphics[width=0.135\textwidth,]{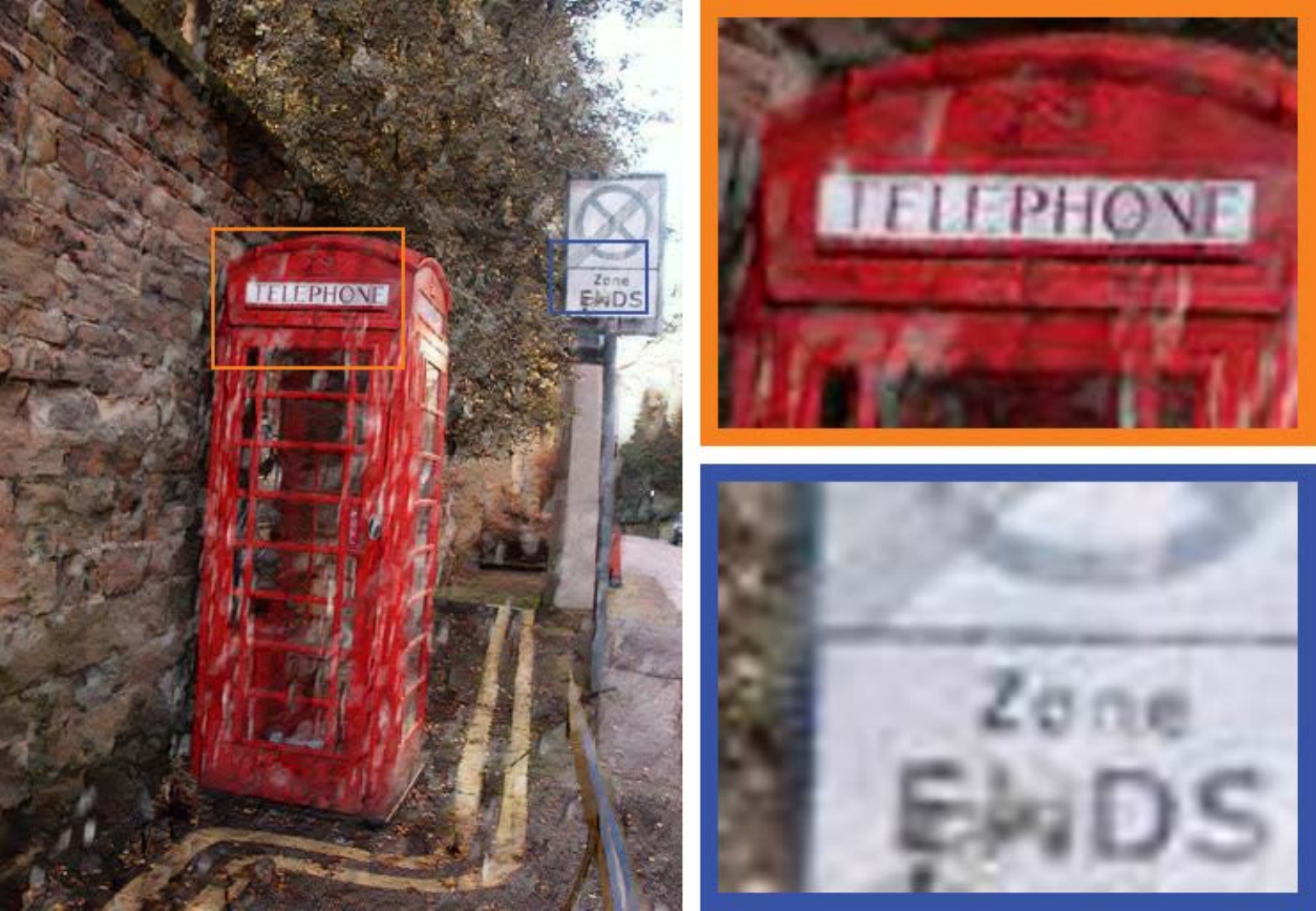}
			&\includegraphics[width=0.135\textwidth,]{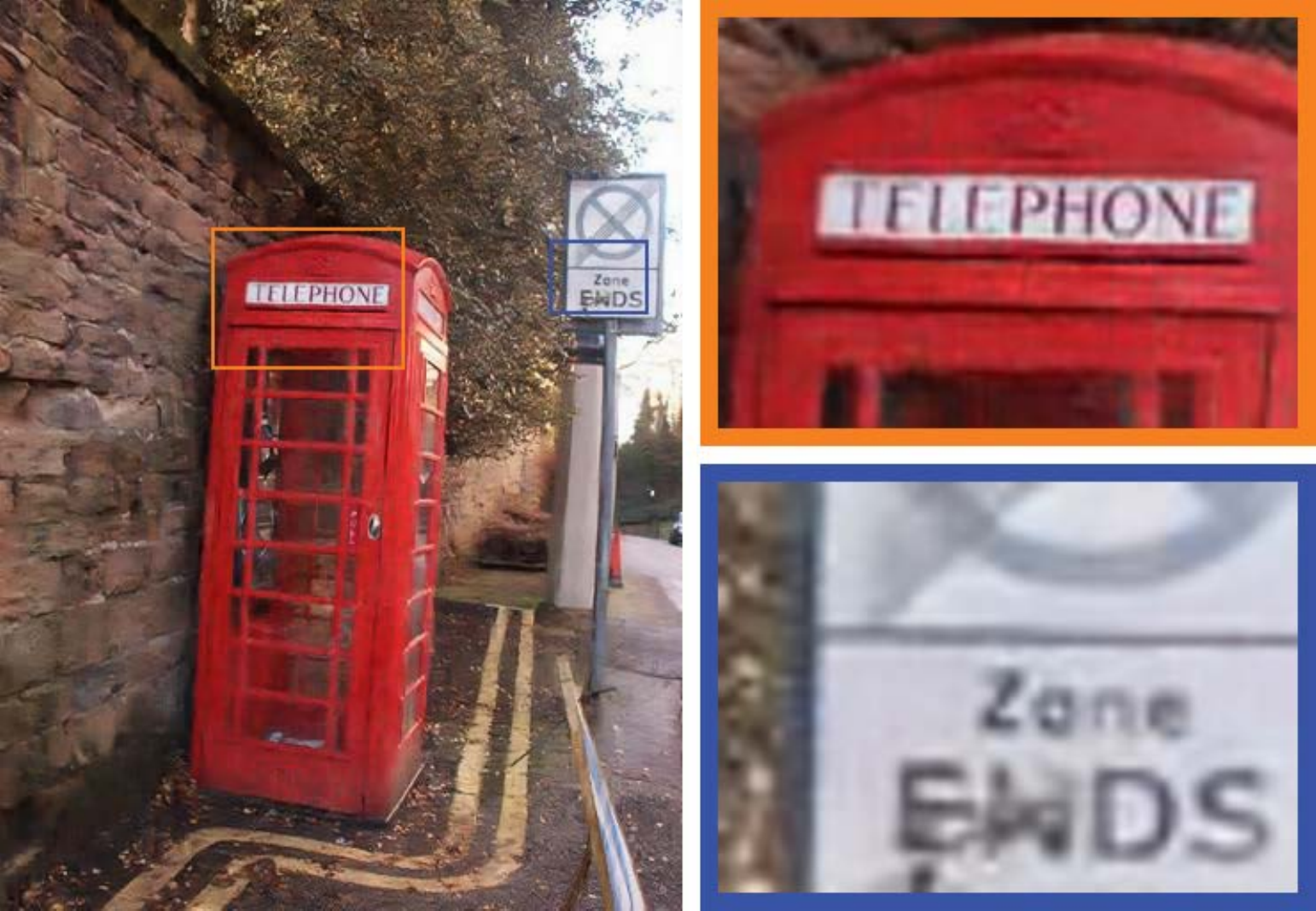}
			&\includegraphics[width=0.135\textwidth,]{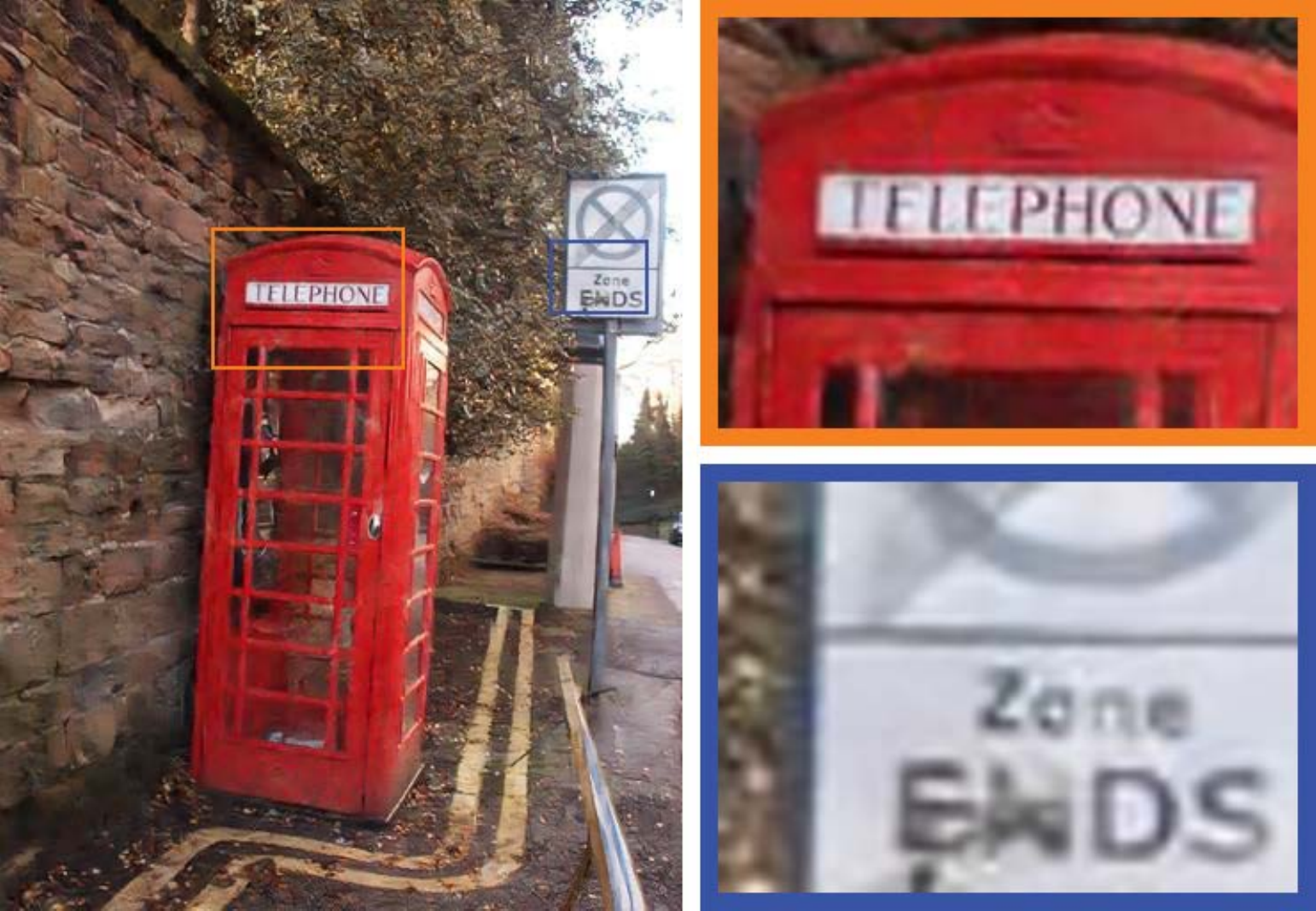}
			\\
			
			\includegraphics[width=0.135\textwidth,]{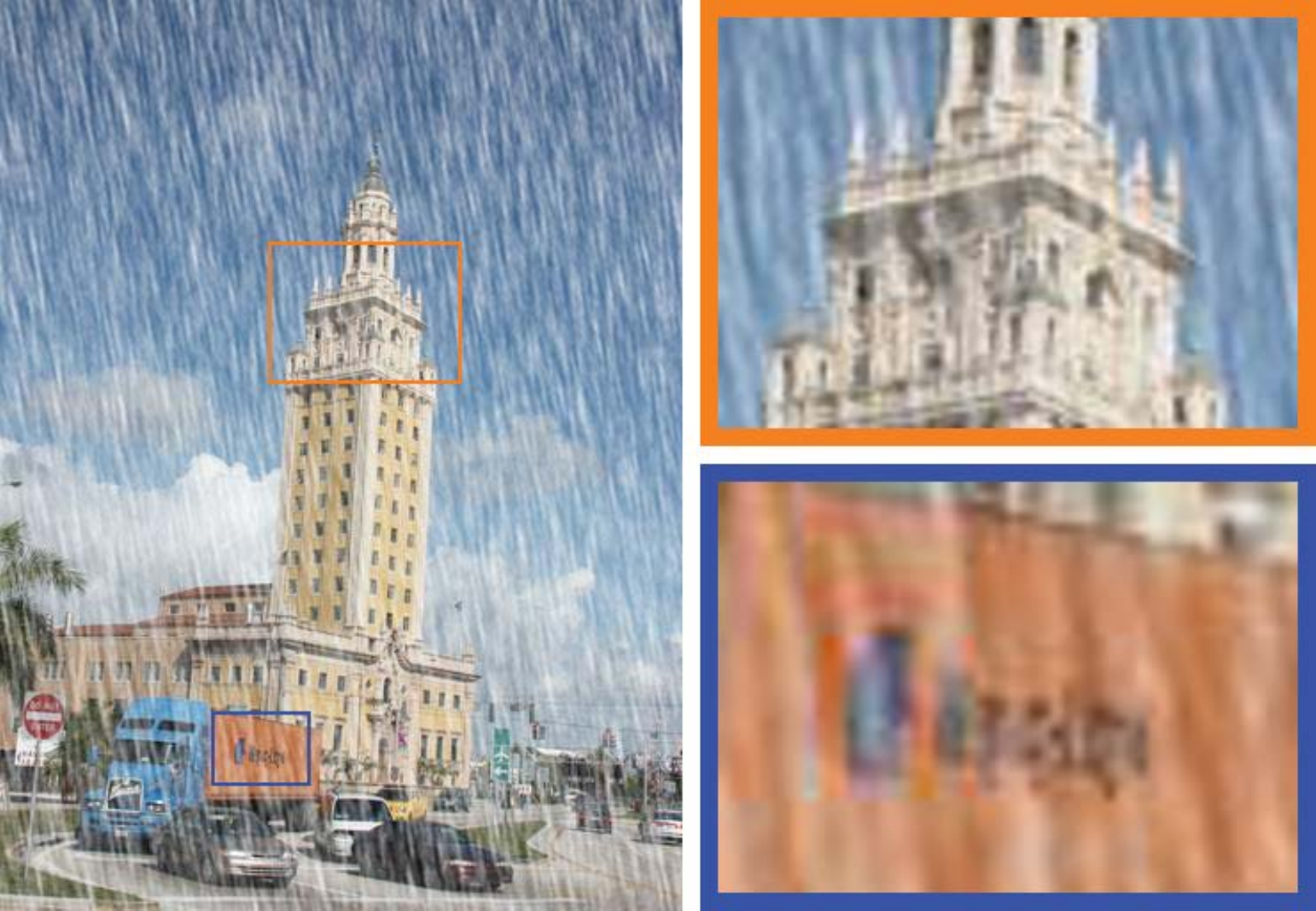}
			&\includegraphics[width=0.135\textwidth,]{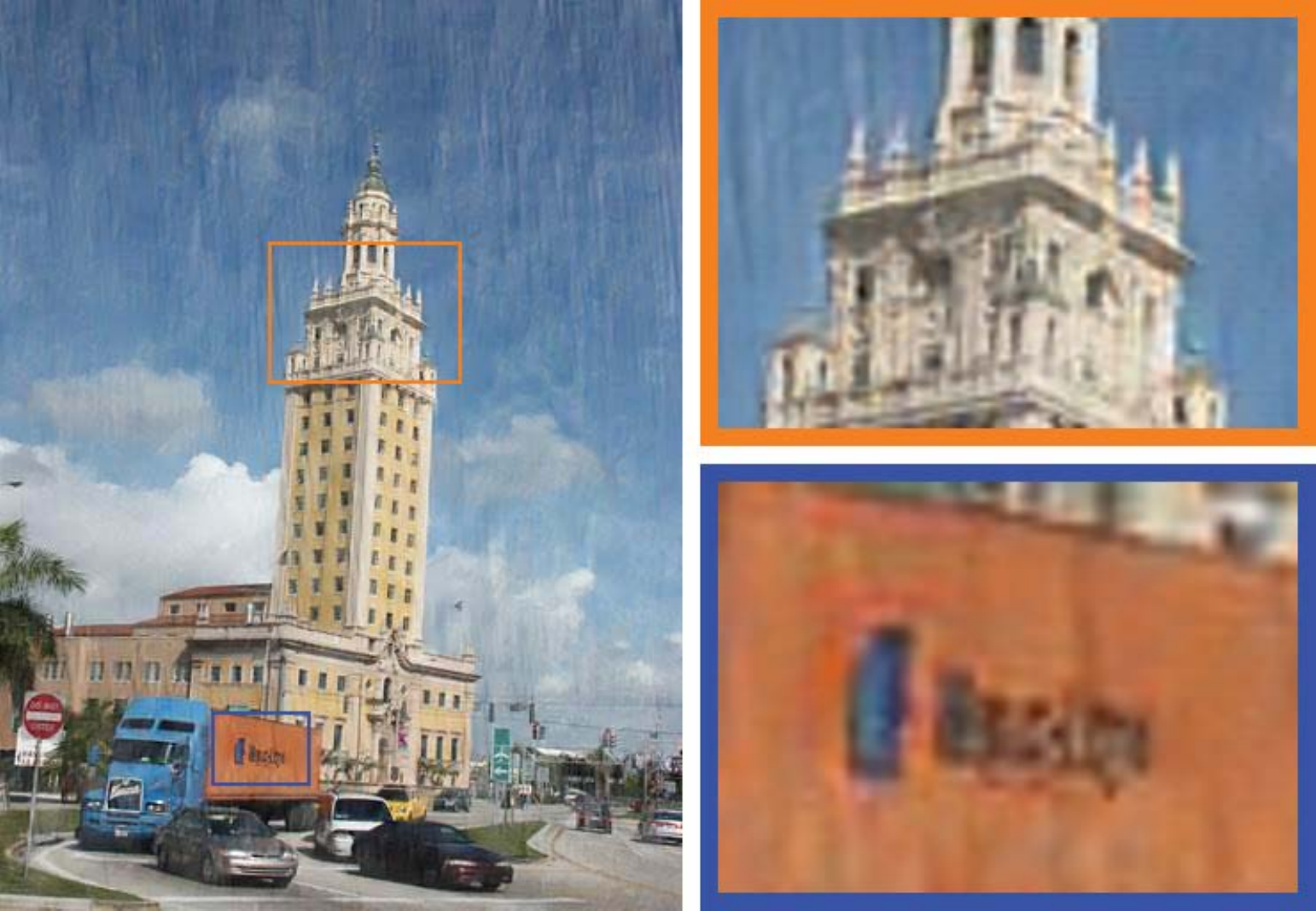}
			&\includegraphics[width=0.135\textwidth,]{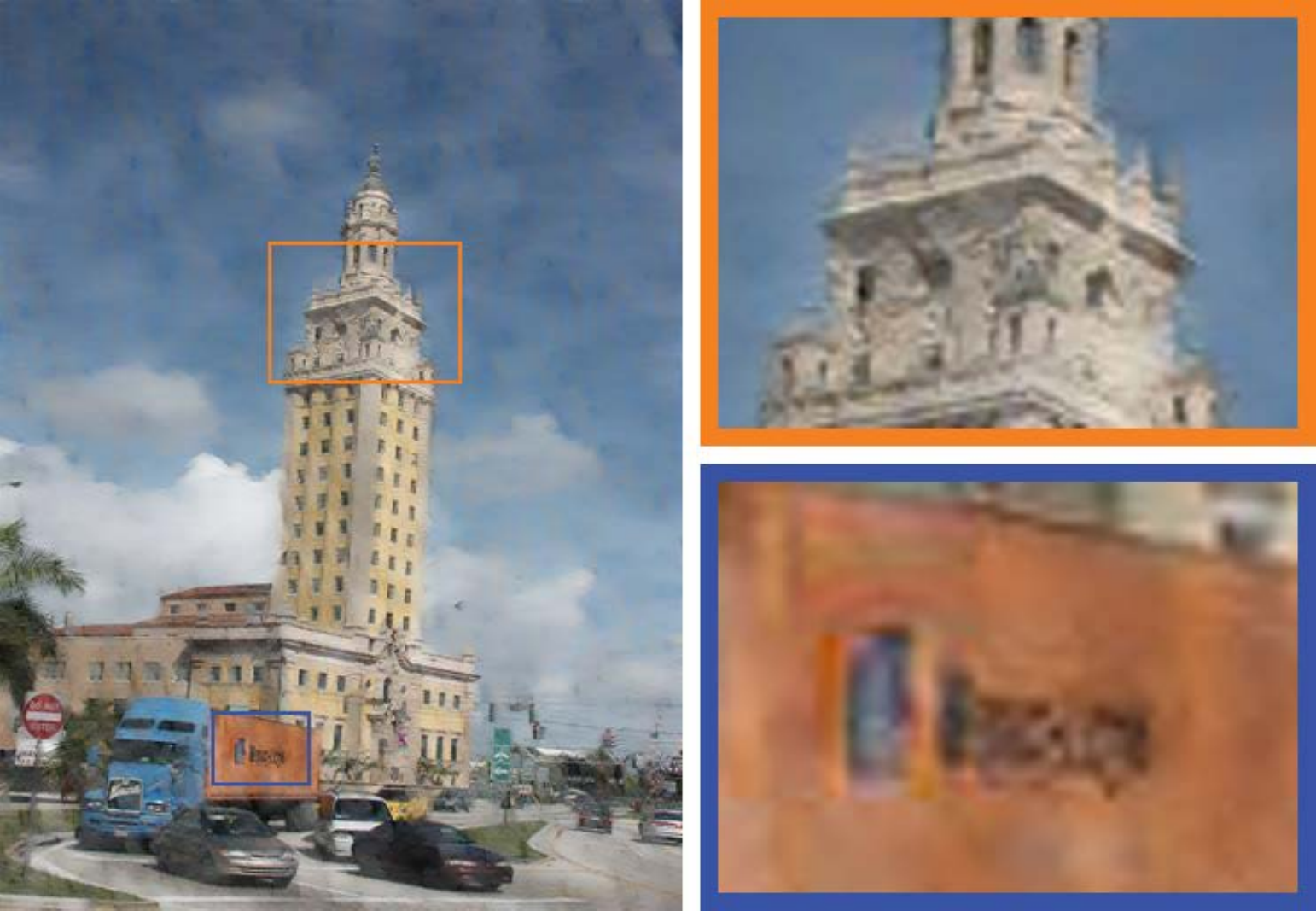}
			&\includegraphics[width=0.135\textwidth,]{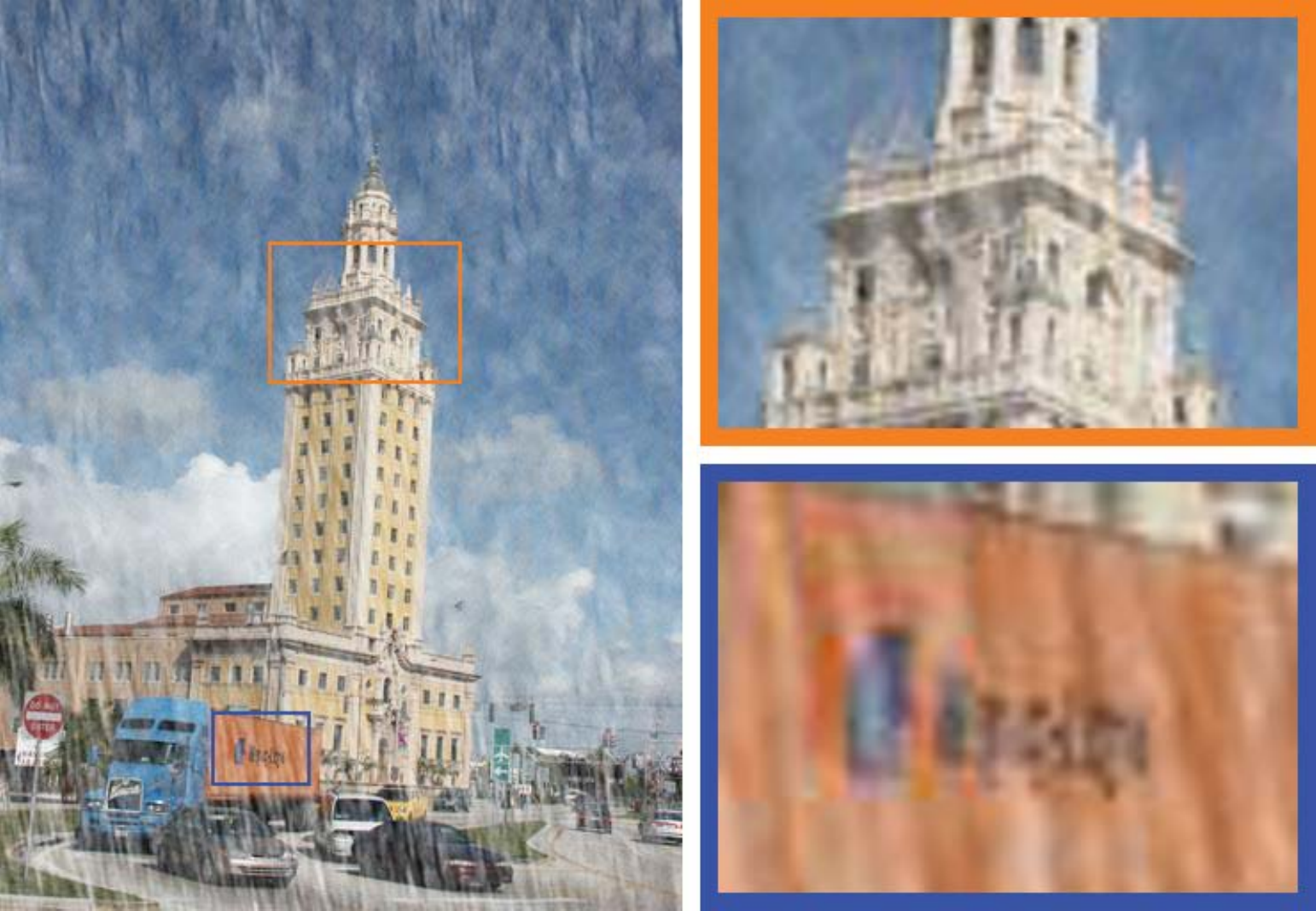}
			&\includegraphics[width=0.135\textwidth,]{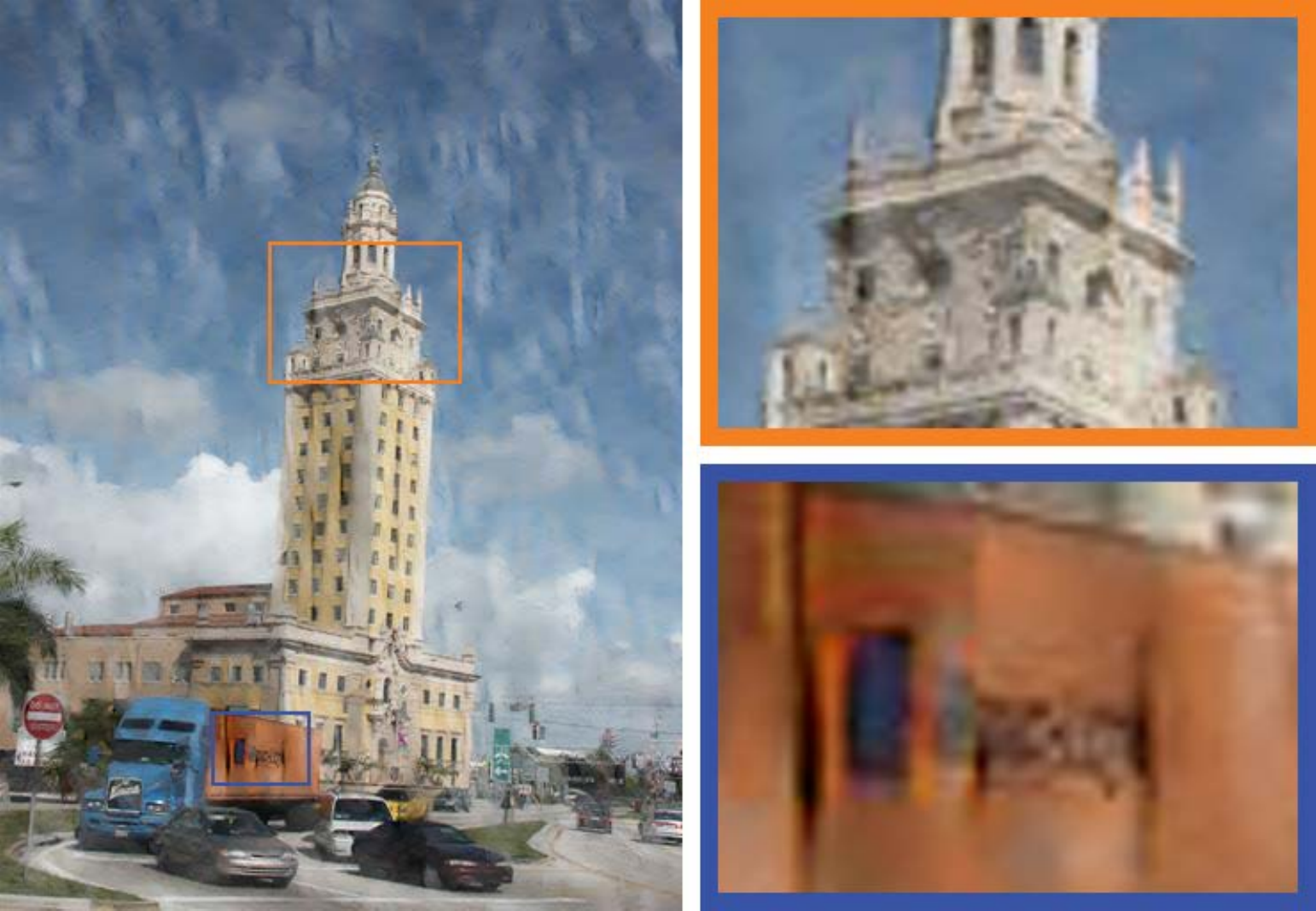}
			&\includegraphics[width=0.135\textwidth,]{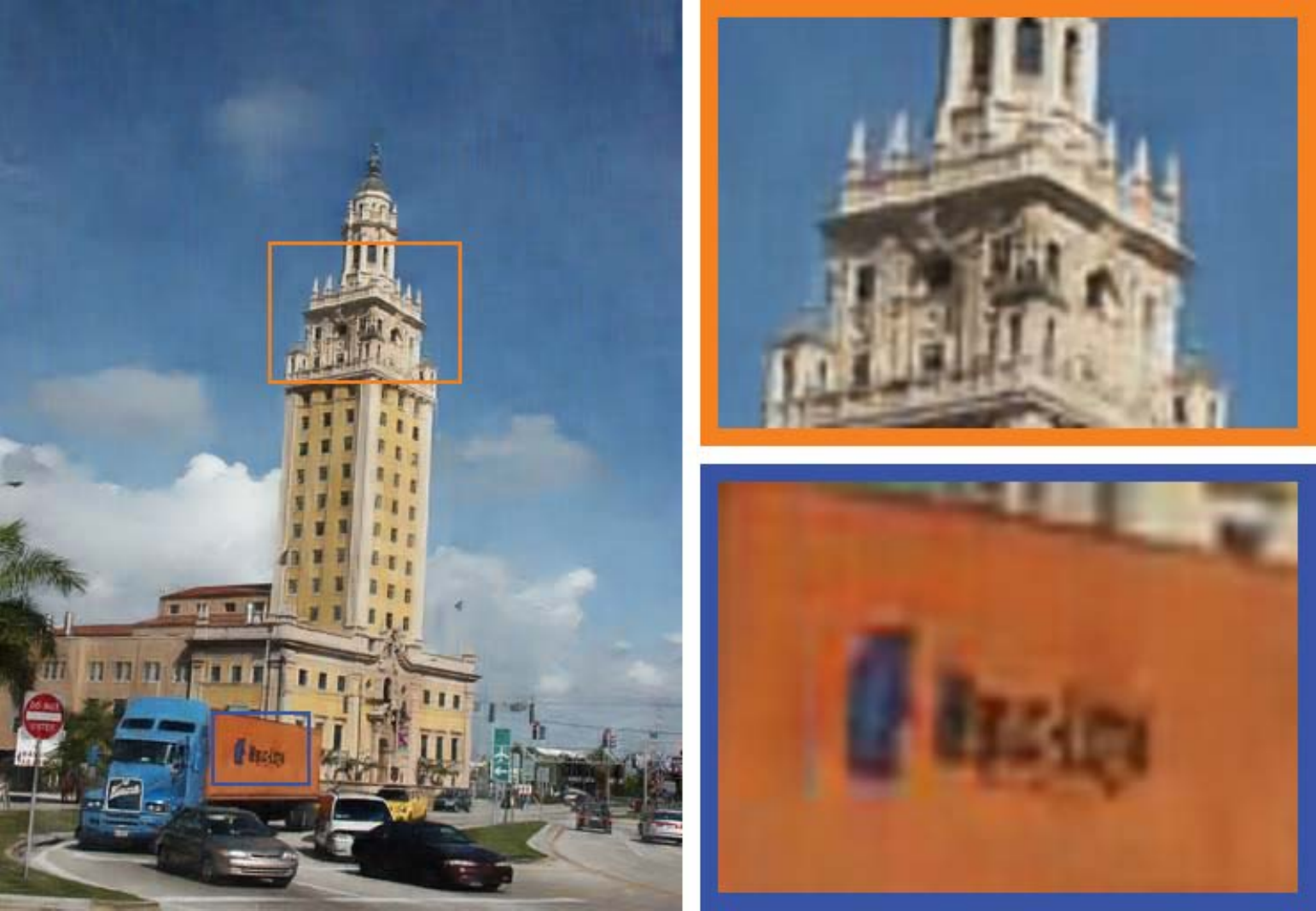}
			&\includegraphics[width=0.135\textwidth,]{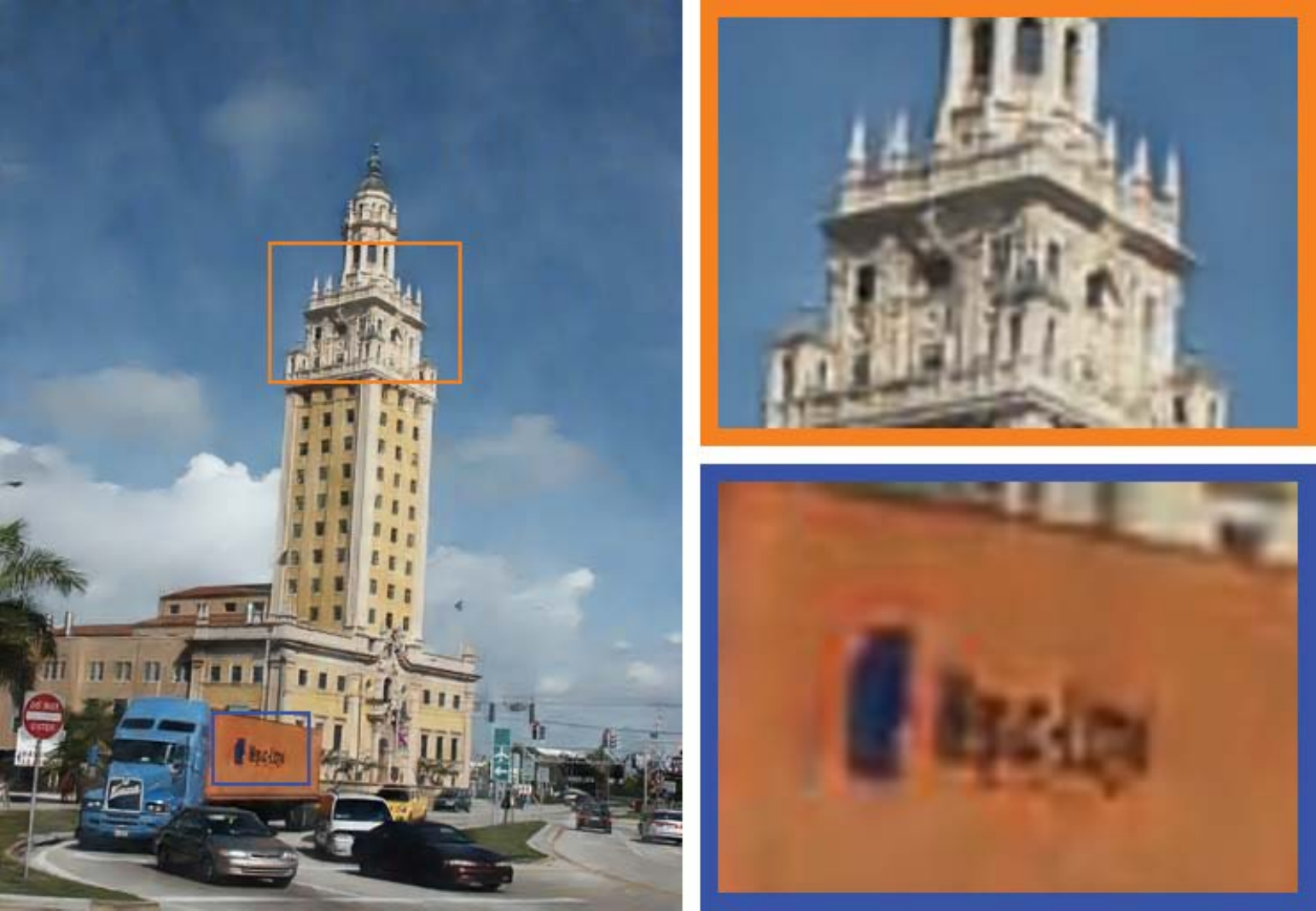}
			\\
			\footnotesize	(a) Input &\footnotesize (b) MWR~\cite{Chen2022MultiWeatherRemoval} &\footnotesize (c)  SAPNet~\cite{zheng2022sapnet} &\footnotesize	(d) AirNet~\cite{AirNet} &\footnotesize (e) DualCNN~\cite{pan2022dual}   &\footnotesize (f)  CLEARER~\cite{gou2020clearer}  & \footnotesize (g)  Ours\\
		\end{tabular}
		\caption{Rain removal comparisons with other competitive methods.}
		\label{fig:derain_result}
	\end{figure*}

	\textbf{Rain Removal}:
		Emphatically, we  introduced AS to address this complicated rain removal task. We competed our schemes against other state-of-the-art algorithms, including 
	MWR~\cite{Chen2022MultiWeatherRemoval}, SAPNet~\cite{zheng2022sapnet}, AirNet~\cite{AirNet}, ConNet~\cite{chen2021contourletnet}, DualCNN~\cite{pan2022dual} and NAS-based HiNAS~\cite{zhang2020memory} and CLEARER~\cite{gou2020clearer}. We retrained DualCNN using similar configurations to ensure a fair comparison.
	As reported in Table~\ref{tab:data_set_result_rain}, our method achieved the best quantitative results in terms of PSNR and SSIM measurements, which also demonstrated the flexibility and efficiency of our AS for GDC. We also provide two representative qualitative instances in Fig.~\ref{fig:derain_result}. GDC scheme can effectively remove most of the rain streaks and maintain the texture details, compared with the latest image deraining methods.
	
	\begin{table}	\renewcommand{\arraystretch}{1.2}
		\caption{Quantitative results about rain removal on Rain800.}
		\label{tab:data_set_result_rain}
				\vspace{-0.2cm}
		\centering\footnotesize
		\setlength{\tabcolsep}{1.2mm}{
			\begin{tabular}{|c| c| c| c| c| c|c| c| c| c|}
				\hline
				Metrics&	~\cite{Chen2022MultiWeatherRemoval} & ~\cite{zheng2022sapnet} &  ~\cite{AirNet} & ~\cite{chen2021contourletnet} &  ~\cite{pan2022dual}  & ~\cite{zhang2020memory}  & ~\cite{gou2020clearer}  &  Ours  \\ \hline		
				PSNR & 22.45 & 21.94 & 21.99 & 27.89 & 24.63 &26.31 &27.23 & \textbf{ 28.14} \\ \hline
				SSIM & 0.76 & 0.76 & 0.71 &0.88  &0.83 & 0.86 &0.86 & \textbf{0.88} \\ 
				\hline
			\end{tabular}	
		}
	\end{table}

	\subsection{Extending to Feature-Level: Optical Flow}
	Furthermore, we also expand our GDC scheme to feature-level optimization scenarios.
Optical flow which relates to the pixel correspondence across two consecutive frames.  Specifically, we denote the optical flow as $\mathbf{u}$ and formulate the fidelity as $f(\mathbf{u})=\|\mathbf{I}_1(\mathbf{u})-\mathbf{I}_0\|^2$ 
and $\phi(\mathbf{u})=\|\nabla\mathbf{u}\|_1$, where $\mathbf{I}_0$, $\mathbf{I}_1$ are two consecutive frames and $\mathbf{I}_1(\mathbf{u})$ denotes the estimated locations of previous $\mathbf{I}_0$ warped by $\mathbf{u}$. The fidelity imposes the  pixel correspondence of images and penalizes the brightness constancy between two frames. Moreover, the widely used TV regularization for optical flow is introduced to penalize the high variants of flow for local smoothness. Instead of utilizing the homologous  network for image restoration, we develop a task-specific GM to establish feature propagation by introducing the pixel correspondence in the pyramid feature domain.  

\begin{figure}[htb]
		\centering \begin{tabular}{c@{\extracolsep{0.2em}}c@{\extracolsep{0.2em}}c@{\extracolsep{0.2em}}c}

			\includegraphics[width=0.11\textwidth]{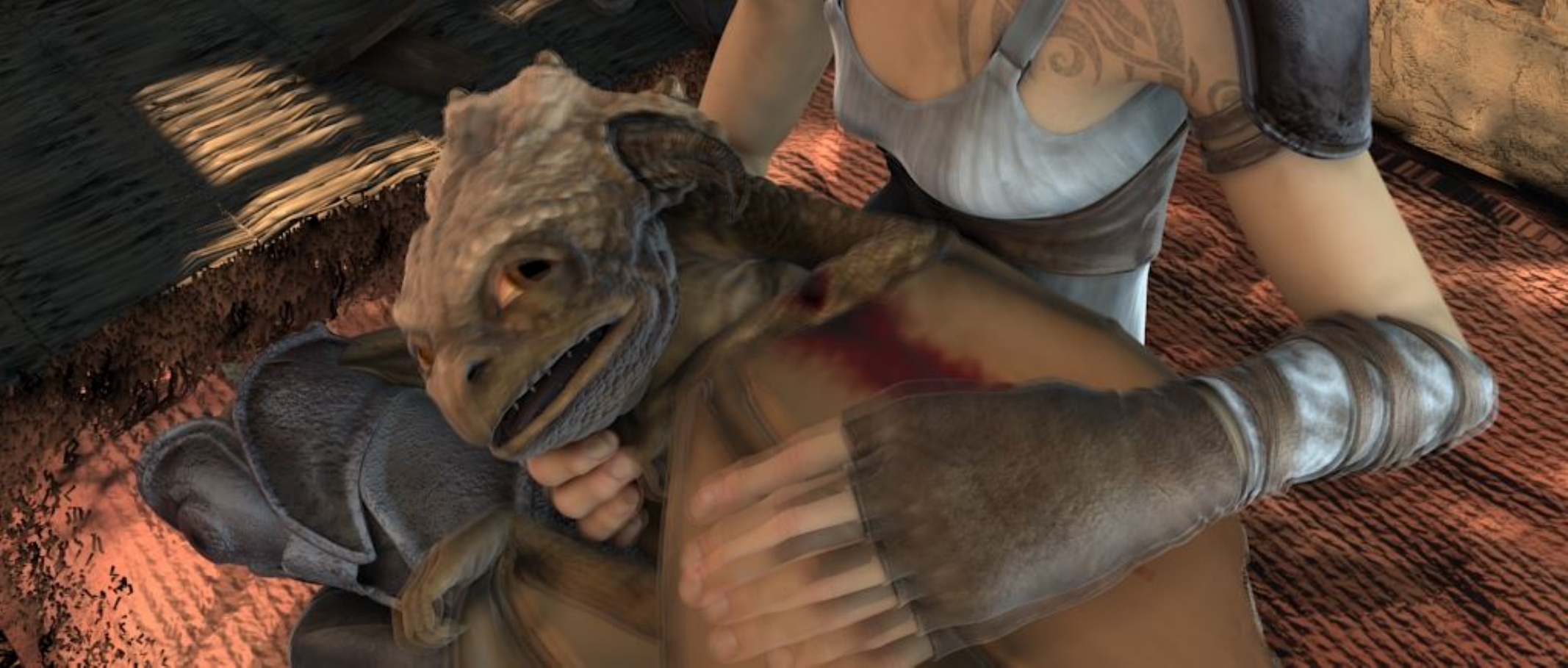}
			&\includegraphics[width=0.11\textwidth]{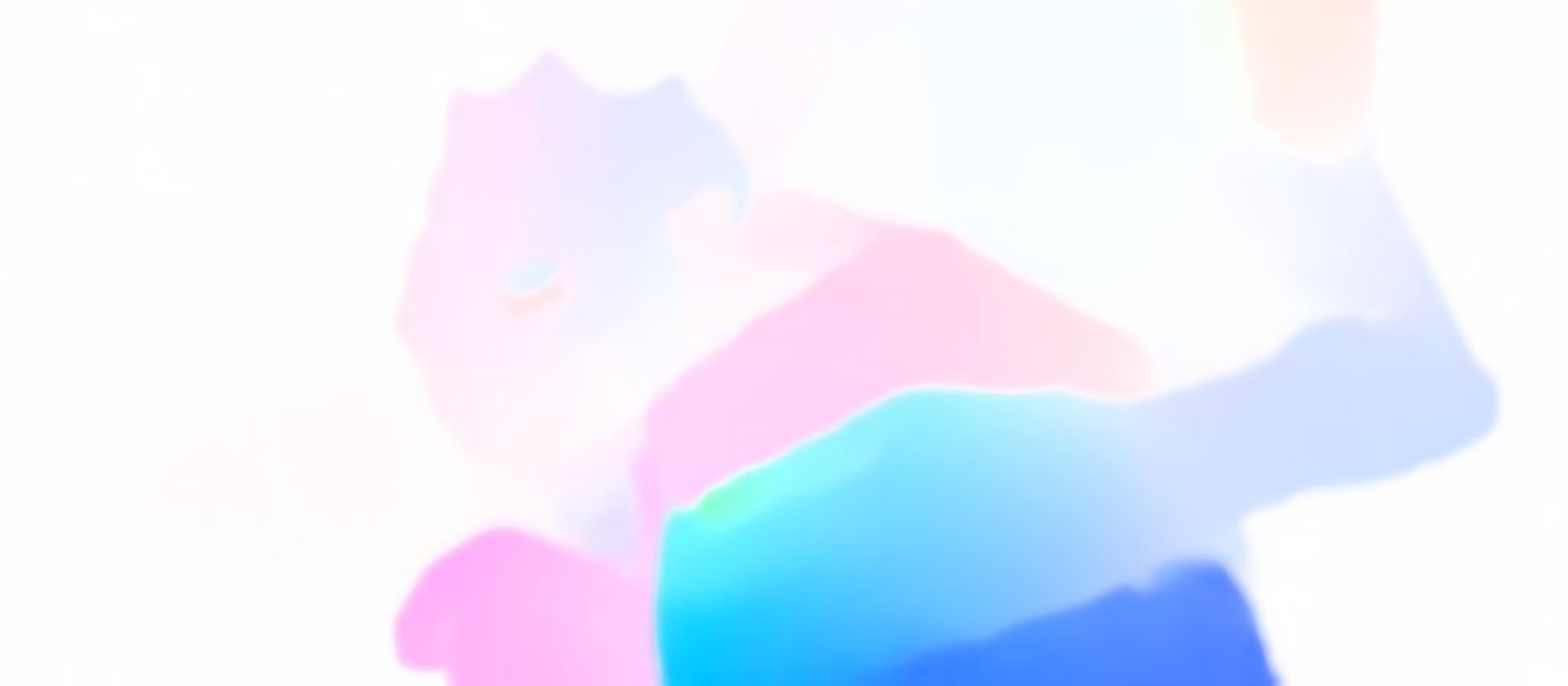}
			&\includegraphics[width=0.11\textwidth]{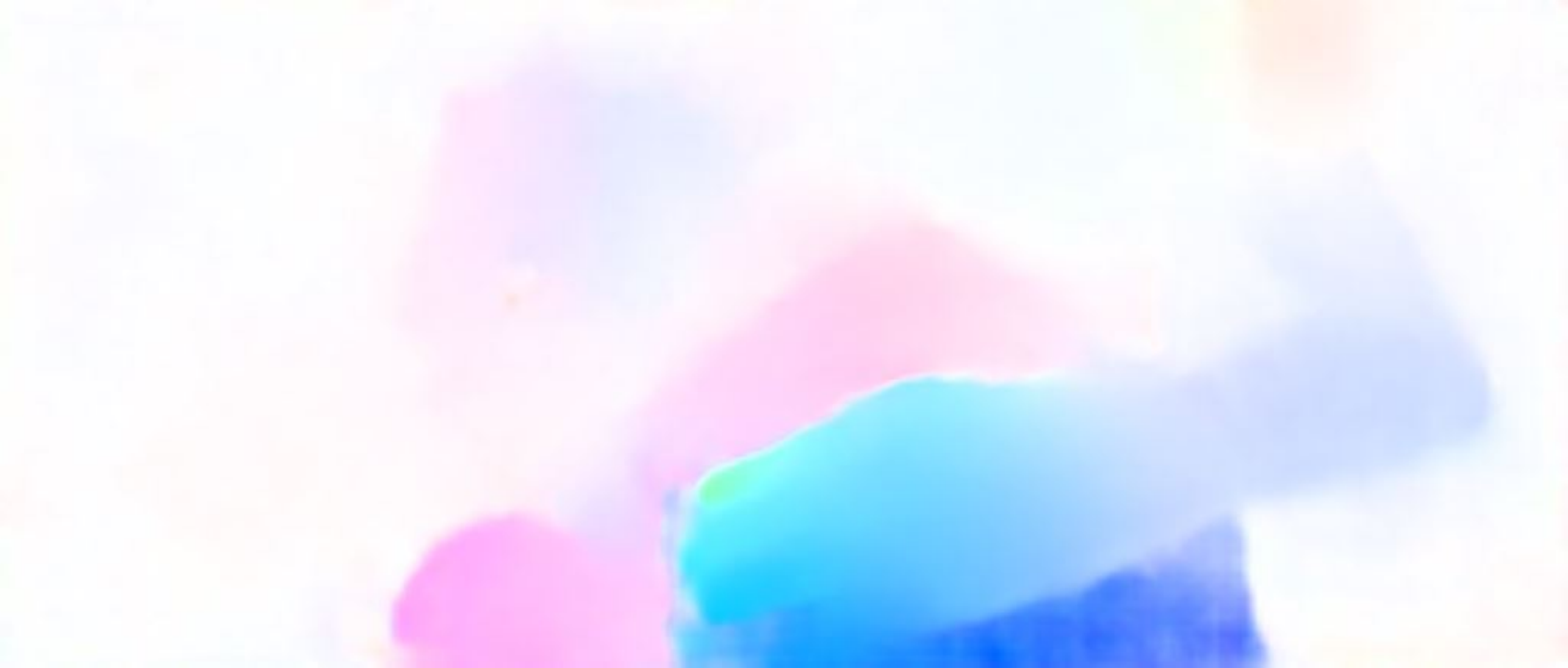}
			&\includegraphics[width=0.11\textwidth]{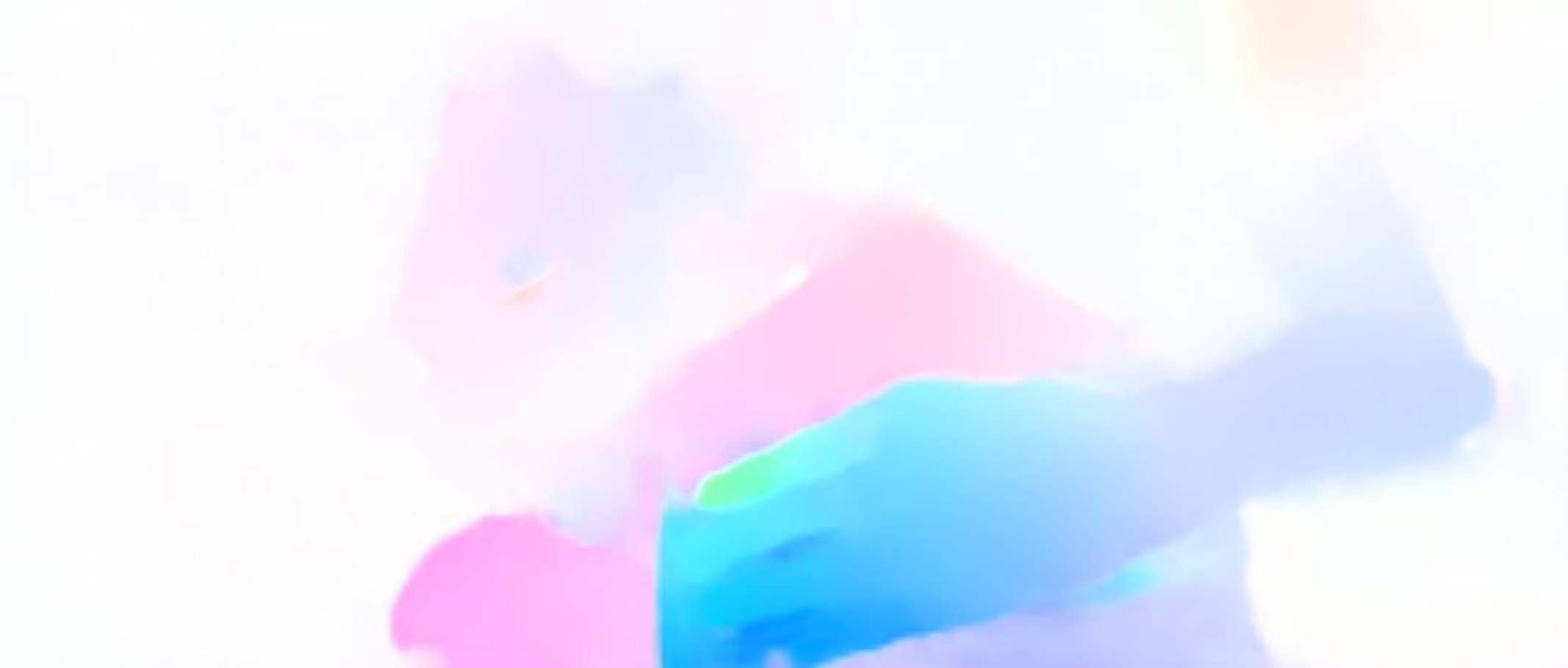}\\
			
			\textemdash & \footnotesize 1.95 & \footnotesize 1.31& \footnotesize\textbf{1.03} \\
			\includegraphics[width=0.11\textwidth]{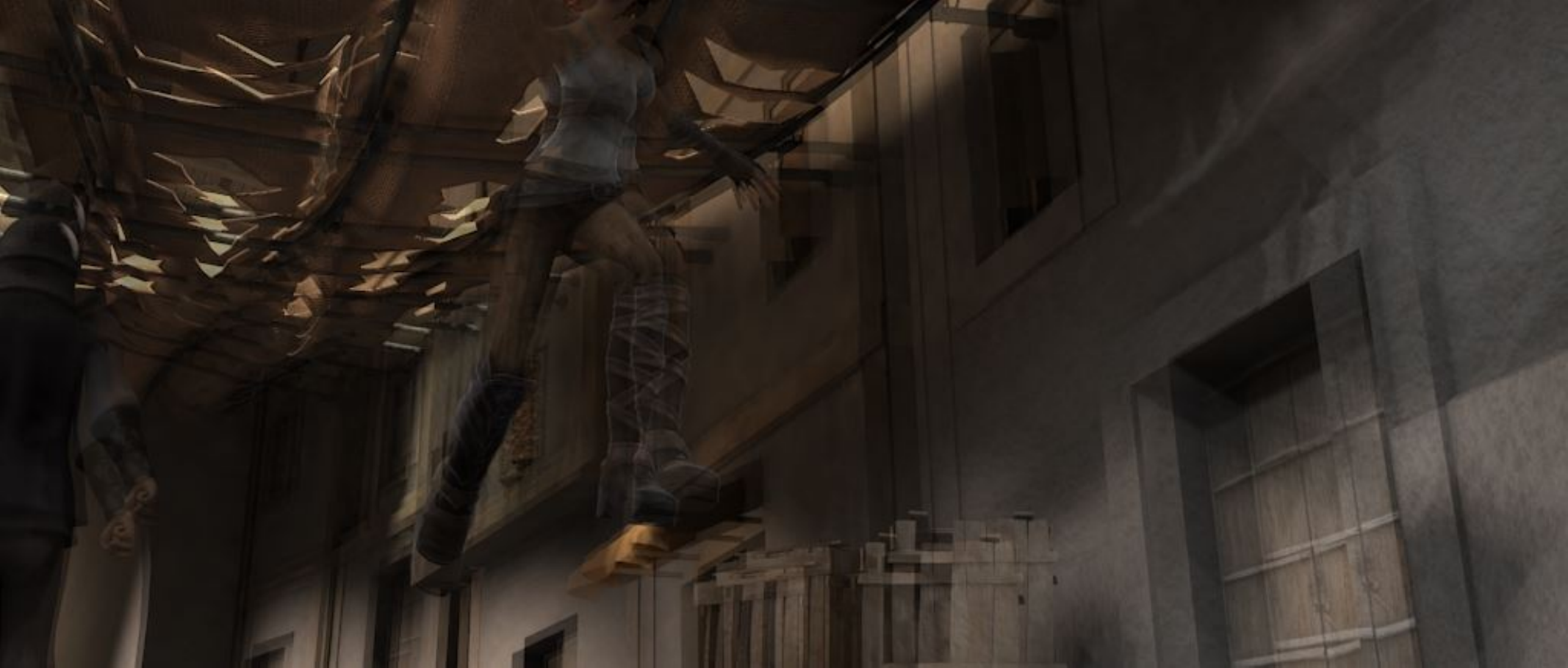}
			
			&\includegraphics[width=0.11\textwidth]{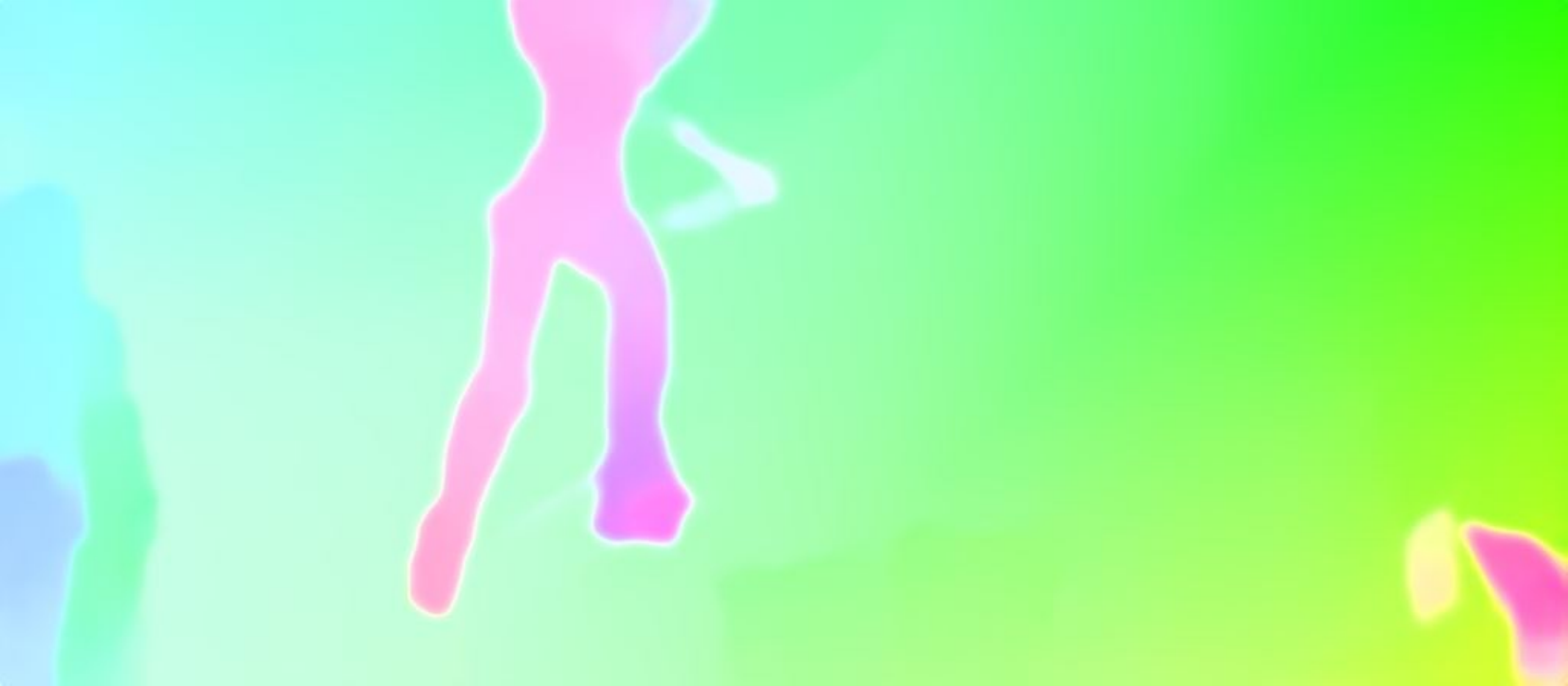}
			&\includegraphics[width=0.11\textwidth]{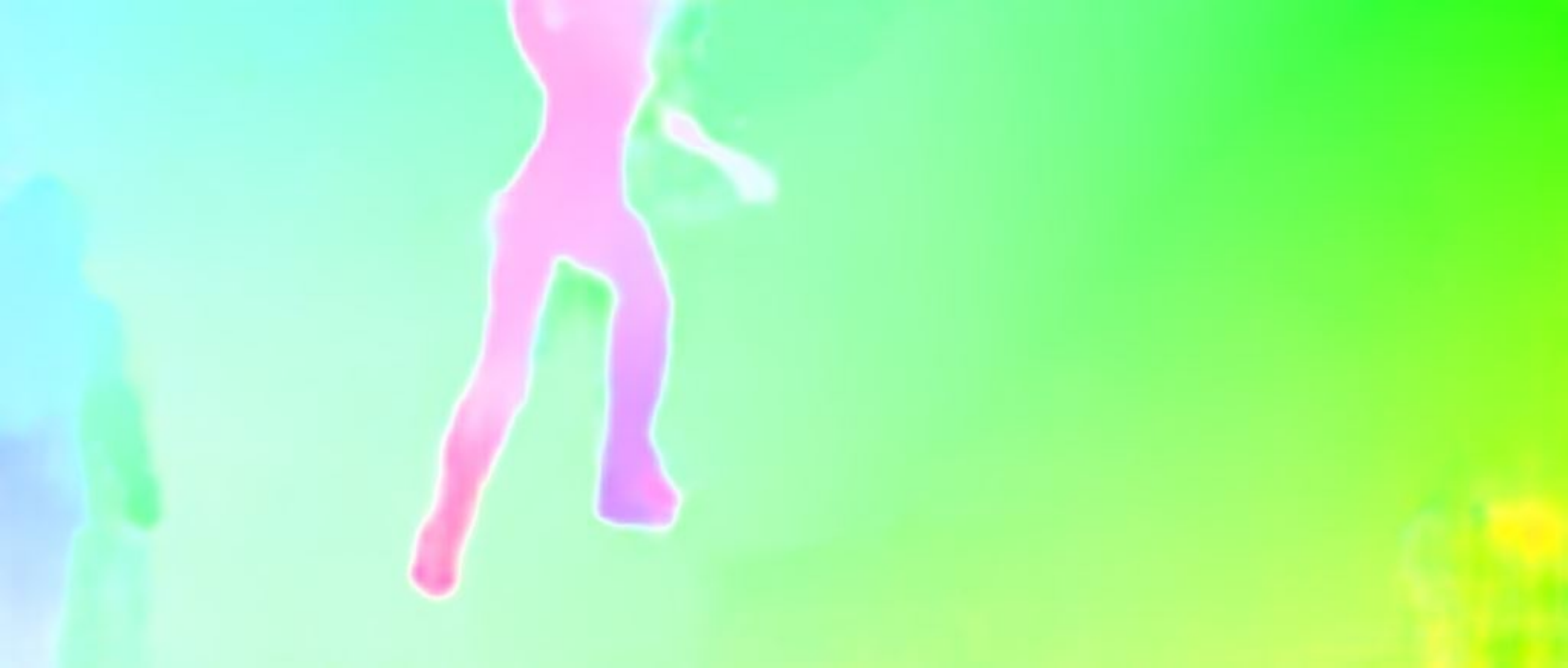}
			&\includegraphics[width=0.11\textwidth]{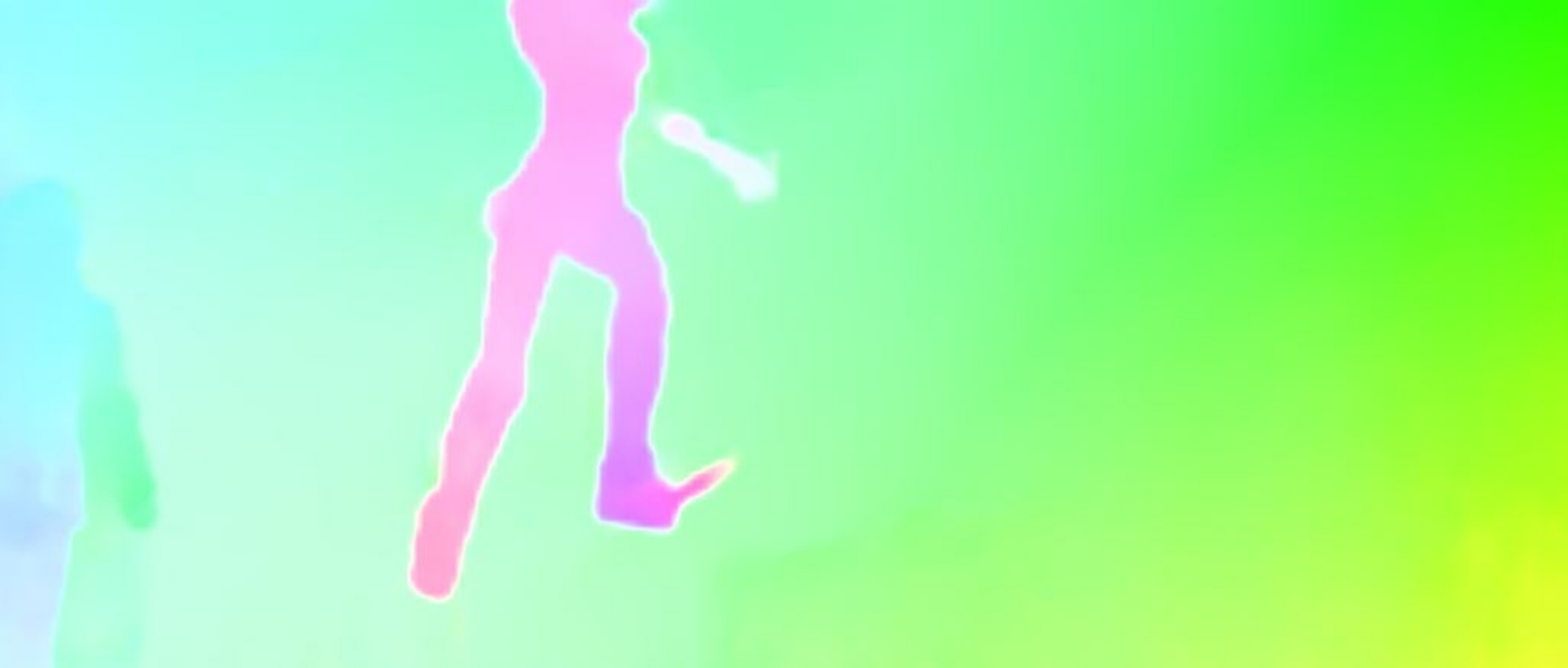}\\
			
			\textemdash & \footnotesize 2.08 & \footnotesize 1.44 &\footnotesize \textbf{1.26} \\
			\footnotesize (a) Frames & \footnotesize(b) Flownet~\cite{Ranjan_CVPR_2017}  & \footnotesize (c) PWCNet~\cite{Sun2018PWC-Net}  & \footnotesize (d) Ours\\
		\end{tabular}
		\caption{Optical flow comparisons on two frames from Sintel~\cite{Butler:ECCV:2012}. Quantitative metrics (EPE) are reported below each result.
		}
		\label{fig:optical_flow}
	\end{figure}

We compared our method with  specific 
optical flow estimation schemes, including FlowNet~\cite{DFIB15}, SpyNet~\cite{Ranjan_CVPR_2017}, Liteflownet~\cite{hui18liteflownet} and PWCNet~\cite{Sun2018PWC-Net}. 
Technically, as for the architecture of GM, we first constructed  pyramid encoders to extract the multi-scale feature of both frames. By utilizing the warping technique and correlation layers,
which have been leveraged in~\cite{DFIB15,Sun2018PWC-Net} to calculate the correlation of corresponding pixels, we imposed this consistency in the feature domain and established a residual CNN to estimate  optical flow in different scales. Behind each level of GM, we adopted the CM to regularize the estimated flow. In Table~\ref{tab:data_set_result}, we provided  average End Point Error (EPE) on different  benchmarks. We
illustrated that our framework is superior to all other
networks~\cite{Ranjan_CVPR_2017,hui18liteflownet,Sun2018PWC-Net} on the Sintel final~\cite{Butler:ECCV:2012} It is clear to see that the generated flows of our method have lesser artifacts and sharper boundaries in Fig.~\ref{fig:optical_flow}. Our method is better to capture  motion details (\textit{e.g.,}, hand movement).
	
	\begin{table}	\renewcommand{\arraystretch}{1.2}

		\caption{{Average EPE of GDC compared with other optical flow estimation methods on two benchmarks.}}
		\label{tab:data_set_result}
		\centering\footnotesize
				\vspace{-0.2cm}
		\setlength{\tabcolsep}{3.5mm}{
			\begin{tabular}{|c| c| c| c|  c| c|  c|}
				\hline
				Metrics&	~\cite{DFIB15} &  ~\cite{Ranjan_CVPR_2017} &  ~\cite{hui18liteflownet}   & ~\cite{Sun2018PWC-Net}    &  Ours  \\ \hline		
				Sintel Clean & 3.04 & 4.12  & 2.48 &2.55 & \textbf{ 2.28} \\ \hline
				Sintel	Final & 	4.60  & 5.57 & 4.04 & 3.93  & \textbf{3.74} \\ 
				\hline
			\end{tabular}	
		}
	\end{table}

	\section{Conclusions}
	This paper provided a unified optimization-inspired learning framework, aggregating  generative learning, discriminative metric, and
model  correction principles from the ``optimization-inspired learning'' perspective. We also proposed two architecture augmentation strategies (\textit{i.e.,}  normalization and architecture search)  to enhance the stability of  propagations and discover task-specific architecture in the propagation.  We introduced two propagative control mechanisms to guarantee  convergent iterations for both fully- and partially-defined optimization models.
Extensive experimental results demonstrated the high flexibility and effectiveness of our framework for diverse low-level vision applications.
	
	\appendix
	
	\subsection{Proof for Proposition~1}
	We first consider following Condition~\ref{assum:function} to introduce the convergent application scenarios of AN, where the proof of Proposition~\ref{prop:p2} is based on this condition.
	\begin{con}\label{assum:function}
		The function $f$ is $\rho$-strong convex   and its gradient is Lipschitz with constant $L$, where  $\rho>0$ and  $L>0$.  $\phi$ is proper convex function.
	\end{con}
	\begin{proof}
		Based on the above Condition~\ref{assum:function}, we can prove the fixed-point convergence of AN.
		In accordance with  Condition~\ref{assum:function} of the convexity of $\phi$, the nonexpansive property of  $\mathtt{prox}_{\gamma^t, \phi} ({\mathbf{u}}^{t+1}_{d} - \frac{1}{\gamma^{t}}\nabla  f({\mathbf{u}}^{t+1}_{d}))$ holds~\cite{bauschke2011convex} which means
		\begin{equation*}
		\begin{array}{l}
		\|\mathcal{T}_{\mathcal{C}}({\mathbf{u}}^{t+1}_d) -\mathcal{T}_{\mathcal{C}}( {\mathbf{u}}^{t}_d)\|^2\\  \leq  \|{\mathbf{u}}^{t+1}_d - \frac{1}{\gamma^{t}} \nabla f({\mathbf{u}}^{t+1}_d) - {\mathbf{u}}^{t}_d + \frac{1}{\gamma^{t}} \nabla f({\mathbf{u}}^{t}_d)\|^2 \\ = \|{\mathbf{u}}^{t+1}_d - {\mathbf{u}}^{t}_d\|^2 - \frac{2}{\gamma^{t}}
		\langle \nabla f({\mathbf{u}}^{t+1}_d) - \nabla f({\mathbf{u}}^{t}_d), {\mathbf{u}}^{t+1}_d - {\mathbf{u}}^{t}_d\rangle\\
		+ (\frac{1}{\gamma^{t}})^2 \|\nabla f({\mathbf{u}}^{t+1}_d) -\nabla f({\mathbf{u}}^{t}_d)\|^2,
		\end{array}
		\end{equation*}
		where we denote $\mathcal{T}_{\mathcal{G}}$, $\mathcal{T}_{\mathcal{D}}$ and $\mathcal{T}_{\mathcal{C}}$ as the operators of GDC propagation respectively. 
		Since $f$ is $\rho$-strong convex based on Condition~\ref{assum:function}, then we have that~\cite{yu2004introductory},
		\begin{equation*}\begin{array}{l}
		\langle \nabla f({\mathbf{u}}^{t+1}_d) - \nabla f({\mathbf{u}}^{t}_d), {\mathbf{u}}^{t+1}_d - {\mathbf{u}}^{t}_d\rangle \geq \frac{\rho L}{\rho+L}\| {\mathbf{u}}^{t+1}_d - {\mathbf{u}}^{t}_d\|^2 \\
		+\frac{1}{\rho + L} \|\nabla f({\mathbf{u}}^{t+1}_d) - \nabla f({\mathbf{u}}^{t}_d)\|^2.
		\end{array}
		\end{equation*}
		Combining with the above two inequalities, we have
		\begin{equation*}\label{eq:T_c}
		\begin{array}{l}
		\|\mathcal{T}_{\mathcal{C}}({\mathbf{u}}^{t+1}_d)- \mathcal{T}_{\mathcal{C}}({\mathbf{u}}^{t}_d)\|^2 \leq (1- \frac{2\rho L }{\gamma^{t}(\rho+L)})\|{\mathbf{u}}^{t+1}_d -{\mathbf{u}}^{t}_d\|^2\\
		+((\frac{1}{\gamma^t})^2 -\frac{{2}}{\gamma^t(\rho+L)}) \|\nabla f({\mathbf{u}}^{t+1}_d)-\nabla f({\mathbf{u}}^{t}_d)\|^2.
		\end{array}
		\end{equation*}
		Therefore, for any $\gamma^t \geq \frac{\rho+L}{2}$, we have $\|\mathcal{T}_{\mathcal{C}}({\mathbf{u}}^{t+1}_d)-\mathcal{T}_{\mathcal{C}}({\mathbf{u}}^{t}_d)\|\\
		\leq \sqrt{1- \frac{2\rho L}{\gamma^t(\rho+L)}} \|{\mathbf{u}}^{t+1}_d-{\mathbf{u}}^{t}_d\|.$
		With the augmentation of AN, assuming that the inner deep classifier of DM has $K$ layers, we have that
		\begin{equation*}\label{eq:lip}
		\begin{array}{l}
		{\| \mathcal{D}({\mathbf{u}}^{t+1}_{g}) \!-\!  \mathcal{D}({\mathbf{u}}^{t}_{g})\|}~\!/\!~ {\|{\mathbf{u}}^{t+1}_{g}  \!-\!{\mathbf{u}}^{t}_{g} \| } \!\leq\! \prod\limits_{k = 1}^{K} \max\limits_{\mathbf{u}^t \neq \mathbf{0}} \frac{\|\tilde{\bm{\omega}}_{d}{\mathbf{u}}^{t+1}_{g}\|}{\|{\mathbf{u}}^{t+1}_{g}\|} \leq \delta_d,
		\end{array}
		\end{equation*}
		where $\tilde{\bm{\omega}}_{d}$ denotes the normalized weights. The above equation implies that 
		\begin{equation*}
		\begin{array}{l}
		\|{\mathbf{u}}_g^{t+1}+\mathcal{D}({\mathbf{u}}^{t+1}_{g})-{\mathbf{u}}_g^{t+1} -({\mathbf{u}}_g^{t}+\mathcal{D}({\mathbf{u}}^{t}_{g})) \|^2 \\ \leq \delta^2_d \|{\mathbf{u}}_g^{t+1} - {\mathbf{u}}_g^{t}\|^2.
		\end{array}
		\end{equation*}
		The above inequality yields
		\begin{equation*}\label{eq:T_d}
		\begin{array}{l}
		\frac{1}{(1+\delta_d)^2}\|{\mathcal{T}}_{\mathcal{D}}( {\mathbf{u}}_g^{t+1}) - {\mathcal{T}}_{\mathcal{D}}( {\mathbf{u}}_g^{t})\|^2  + \frac{1-\delta_d^2}{(1+\delta_d)^2}\|{\mathbf{u}}_g^{t+1}-{\mathbf{u}}_g^{t}\|^2\\
		- \frac{2}{1+\delta_d}\langle {\mathbf{u}}_g^{t+1}-{\mathbf{u}}_g^{t}, \frac{1}{1+\delta_d}({\mathcal{T}}_{\mathcal{D}}( {\mathbf{u}}_g^{t+1})- {\mathcal{T}}_{\mathcal{D}}( {\mathbf{u}}_g^{t}))\rangle \leq 0.
		\end{array}
		\end{equation*}
		According to the properties of averaged nonexpansive operators in~\cite{bauschke2011convex}, 
		we know that 
		$\frac{{\mathcal{T}}_{\mathcal{D}}({\mathbf{u}}_g^{t+1})}{1+\delta_d}$ is  $\frac{\delta_d}{1+\delta_d}$- averaged and also is nonexpansive, thus 
		${\mathcal{T}}_{\mathcal{D}}( {\mathbf{u}}_g^{t+1})$ is ($1+\delta_d$)-Lipschitz.
		Similarly, we have $\mathcal{T}_{\mathcal{G}}(\mathbf{u}^t)$ is $(1+\delta_g)$-Lipschitz. We denote that $\mathcal{T}(\mathbf{u}^t)= \mathcal{T}_{\mathcal{C}} \circ \mathcal{T}_{\mathcal{D}} \circ \mathcal{T}_{\mathcal{G}} (\mathbf{u}^{t})$.
		Combining the Lipschitz property of ${\mathcal{T}}_{\mathcal{G}}( {\mathbf{u}}^{t})$, ${\mathcal{T}}_{\mathcal{D}}( {\mathbf{u}}_g^{t+1})$ and ${\mathcal{T}}_{\mathcal{C}}( {\mathbf{u}}_d^{t+1})$, we conclude that
		\begin{equation*}\begin{array}{l}
		\!\|\mathcal{T}(\mathbf{u}^{t+1})\!-\!\mathcal{T}(\mathbf{u}^{t})\|\!=\! \|\mathcal{T}_{\mathcal{C}}\!\circ\!\mathcal{T}_{\mathcal{D}}\!\circ\!\mathcal{T}_{\mathcal{G}}(\mathbf{u}^{t+1})\!-\!\mathcal{T}_{\mathcal{C}}\!\circ\!\mathcal{T}_{\mathcal{D}}\!\circ\!\mathcal{T}_{\mathcal{G}}(\mathbf{u}^{t})\|\\
		\!\leq
		\sqrt{1- \frac{2\rho L}{\gamma^t(\rho+L)}}\|\mathcal{T}_{\mathcal{D}}\circ\mathcal{T}_{\mathcal{G}}(\mathbf{u}^{t+1})-\mathcal{T}_{\mathcal{D}}\circ\mathcal{T}_{\mathcal{G}}(\mathbf{u}^{t})\|\\
		\!\leq
		\sqrt{1- \frac{2\rho L}{\gamma^t(\rho+L)}}(1+\delta_d)(1+\delta_g)\|\mathbf{u}^{t+1}-\mathbf{u}^{t}\|.
		\end{array}
		\end{equation*}
		We set $\delta = \sqrt{1- \frac{2\rho L}{\gamma^t(\rho+L)}}(1+\delta_d)(1+\delta_g).$ Then we have that $\mathcal{T}$ is $\delta$-Lipschitz. 
		Obviously, if $\frac{\rho+L}{2} \leq \gamma^{t} \leq \frac{2\rho L}{ \rho+L} (1+ \frac{1}{(1+\delta_d)^2(1+\delta_g)^2 -1})$
		constant of $\mathcal{T}(\mathbf{u}^{t})$ is less than 1. Indeed, such a $\gamma^t$  exists if $(1+\delta_d)(1+\delta_g) < \frac{\rho+L}{|\rho -L|}$. Then, there exists $\mathbf{u}^{\ast}$ satisfying $\mathbf{u}^{\ast} =\mathcal{T}(\mathbf{u}^{\ast}) $. 	Obviously, $\|{\mathbf{u}}^{t+1}-{\mathbf{u}}^{t}\|\to 0$ as $t\to\infty$ and the sum of $\|{\mathbf{u}}^{t+1}-{\mathbf{u}}^{t}\|$ is bounded, \textit{i.e.,}  $\sum_{t=1}^{\infty}\|{\mathbf{u}}^{t+1}-{\mathbf{u}}^{t}\|<\infty$. This means that the propagative sequence  converges to a fixed-point. 
	\end{proof}
	
	\subsection{Proof for  Proposition~2}
	Based on the above assumption, we provide the detailed proof for  Proposition~2.
	\begin{proof}
		After the checking of PCM, the step~\ref{step:pg_u_new} of Alg.~\ref{alg:correction1} implies that
		$
		{\mathbf{u}}^{t+1} \leftarrow  \mathtt{prox}_{\gamma^t, \phi} ({\mathbf{v}}^{t}- \frac{1}{\gamma^{t}} \nabla f({\mathbf{v}}^{t})).
		$ 
		Thus we have 
		$${\mathbf{u}}^{t+1} \in \arg \min\limits_{\mathbf{u}} \{ \phi(\mathbf{u}) + \frac{\gamma^t}{2}\| \mathbf{u} -
		\mathbf{v}^{t} +  \frac{1}{\gamma^{t}} \nabla f({\mathbf{v}}^{t})\|^2\}.$$
		Subsequently, the above equation yields
		\begin{equation}
		\label{eq:eq12}
		\phi(\mathbf{u}^{t+1}) \leq \phi(\mathbf{v}^{t})- \frac{\gamma^{t}}{2} d_t -
		(\nabla f({\mathbf{v}}^{t}))^{\top}(\mathbf{u}^{t+1} -\mathbf{v}^{t}),
		\end{equation}
		where $d_t=\|\mathbf{u}^{t+1}\!-\!\mathbf{v}^{t} \|^2$. Considering the convexity and the Lipschitz smooth properties of $f$, 
		the following 
		\begin{equation}\label{eq:eq13}
		f(\mathbf{u}^{t+1})\leq f(\mathbf{v}^{t})+(\nabla f({\mathbf{v}}^{t}))^{\top}(\mathbf{u}^{t+1} -\mathbf{v}^{t})- \frac{L}{2} d_t,
		\end{equation}
		holds. Summing up Eq.~\eqref{eq:eq12} and Eq.~\eqref{eq:eq13}, we can have
		$\Psi(\mathbf{u}^t)\geq\Psi(\mathbf{u}^{t+1})+\beta^t d_t$ with $\beta^t=\frac{L +\gamma^t}{2}$.
		If $\mathbf{v}^{t} \leftarrow \mathbf{u}^{t}$, the first conclusion in  Proposition~\ref{prop:p4}  is completed, else we have $\Psi(\mathbf{v}^{t}) = \Psi({\mathbf{u}}^{t+1}_{c}) \leq \Psi({\mathbf{u}}^{t})$. Obviously, we obtain the sufficient descent property. This implies that $\|\mathbf{u}^{t+1}-\mathbf{v}^{t}\|\to 0$ as $t\to\infty$. 
		As the updated step after optimality checking of PCM is computed by  proximal gradient  scheme in~Alg.~\ref{alg:correction1}), the optimal condition 
		$0 \in \partial \phi(\mathbf{u}^{t+1}) + \gamma^t(\mathbf{u}^{t+1}-\mathbf{v}^{t} +\frac{1}{\gamma^{t}} \nabla f(\mathbf{v}^{t}) )$ 
		implies 
		\begin{equation*}
		-{\gamma^{t}}(\mathbf{u}^{t+1} \!-\!\mathbf{v}^{t}) - \nabla f(\mathbf{v}^{t}) +\nabla f(\mathbf{u}^{t+1}) \in  \partial  \phi(\mathbf{u}^{t+1}) + \nabla  f(\mathbf{u}^{t+1}).
		\end{equation*}
		Thus, we conclude that 
		\begin{equation*}
		\begin{array}{l}
		\!\!\|\!-\!{\gamma^{t}}(\mathbf{u}^{t+1}\!-\!\mathbf{v}^{t}) \!-\! \nabla f(\mathbf{v}^{t}) \!+\!\nabla f(\mathbf{u}^{t+1}) \| 
		\!\leq\!  C \|\mathbf{u}^{t+1}\!-\!\mathbf{v}^{t}\|\!\to\! 0,
		\end{array}
		\end{equation*}
		where $C = L+\gamma^t$. 
		Let $\mathbf{u}^{\ast}$ becomes any accumulation point of $\{\mathbf{u}^{t+1}\}$, called $\{\mathbf{u}^{t_p+1}\}\to\mathbf{u}^{\ast}$, as $p\to\infty$. Incorporating the lower semi-continuity of $\phi(\mathbf{u})$ and the supreme principle, we obtain that $	\lim_{p\rightarrow\infty}\Psi(\mathbf{u}^{t_p+1})=\Psi(\mathbf{u}^{\ast})$. 
		Thus, we conclude that $\mathbf{0}\in\partial\Psi(\mathbf{u}^{\ast})$. 
		This completes the proof. 
	\end{proof}
	
	\subsection{Proof for  Proposition~3}
	The following Condition~\ref{cond:bound}  provides the foundational bounded constraint for  deep modules (\textit{i.e.,} GM and DM) to guarantee the convergence of PDM. We prove that, by the control of PCM, the sequence can converge to a fixed-point.
	\begin{con}\label{cond:bound}
		For any input ${\mathbf{u}}$, the proposed deep modules GM and DM should satisfy
		$\|\mathcal{T}_{\mathcal{D}}\!\circ\! \mathcal{T}_{\mathcal{G}}({\mathbf{u}^{t}}) - {\mathbf{u}^{t}} \| \leq   \sqrt{c/\gamma^{t}}$, where $c>0$,
		denoted the step 3 and 4 of Alg.~\ref{alg:correction3} as $ \mathcal{T}_{\mathcal{G}}$ and $ \mathcal{T}_{\mathcal{D}}$ respectively.
	\end{con}
	\begin{proof}
		Based on  Condition~\ref{cond:bound}, we can prove the convergence of PCM for PDM optimization.
		For convenience, we  denote $\ \mathcal{T}_{\mathcal{G}}$, $ \mathcal{T}_{\mathcal{D}}$ are the operators of step 3 and 4 in Alg.~\ref{alg:correction3} and $\rho^{t} = \sqrt{c/\gamma^{t}}$ firstly. Then, based on  Condition \ref{cond:bound},  following inequality is satisfied
		\begin{equation*}\label{eq:Nondeterministic-eq2}
		\begin{array}{l}
		\! \|\mathbf{u}^{t} - {\mathbf{u}}^{t+1}_{d}\| =
		\| \mathcal{T}_{\mathcal{D}}\!\circ\! \mathcal{T}_{\mathcal{G}}(\mathbf{u}^{t}) - \mathbf{u}^{t}\|
		\!\leq \rho^t.
		\end{array}
		\end{equation*}
		For  PDM, the $\mathcal{C}_{\mathtt{PDM}}$ is defined as $\arg\min\limits_{\mathbf{u}} f(\mathbf{u}) + \frac{\gamma^{t}}{2}\|\mathbf{u}-{\mathbf{u}}^{t+1}_{d}\|^2$. With  the first order optimal condition, we have that $\mathbf{0}\in \nabla f({{\mathbf{u}}}^{t+1}_{c})+\gamma^{t}({{\mathbf{u}}^{t+1}_{c}} - {\mathbf{u}}^{t+1}_{d})$. 	
		If  optimality checking {$\|\nabla f(\mathbf{u}_c^{t})\| \leq L $} is satisfied, we can have the following inequality, \textit{i.e.,} $\gamma^{t}\|{{\mathbf{u}}^{t+1}_{c}} - {\mathbf{u}}^{t+1}_{d}\| =\| \nabla f({{\mathbf{u}}}^{t+1})\|\leq L$ which means $\|{{\mathbf{u}}^{t+1}_{c}} - {\mathbf{u}}^{t+1}_{d}\| \leq  \frac{L}{\gamma^{t}}$. With the above inequality, we then have that 
		\begin{equation}\label{eq:res_u}
		\begin{array}{l}
		\!\|{{\mathbf{u}}^{t+1}} -{{\mathbf{u}}^{t}} \| = \|{{\mathbf{u}}^{t+1}_{c}} - {\mathbf{u}}^{t+1}_{d} + ({\mathbf{u}}^{t+1}_{d} -{\mathbf{u}}^{t} )\|\\
		\!\leq \frac{L}{\gamma^{t}} +\rho^t.
		\end{array}
		\end{equation}
		When {$\|\nabla f(\mathbf{u}_c^{t})\| \leq L $} is not satisfied, we set $\mathbf{u}^{t+1}\gets\mathbf{u}^{t}$. Obviously, the above inequality of Eq. \eqref{eq:res_u} is also achieved. 
		As for  $\mathbf{u}^{t+1}_{d}$, with Condition \ref{cond:bound} we have that
		\begin{equation*}\begin{array}{l}
		\! \|{\mathbf{u}}^{t+1}_{d} - {\mathbf{u}}^{t}_{d}\|
		\! = \|\mathcal{T}_{\mathcal{D}}\!\circ\! \mathcal{T}_{\mathcal{G}}(\mathbf{u}^{t}) - \mathcal{T}_{\mathcal{D}}\!\circ\! \mathcal{T}_{\mathcal{G}}(\mathbf{u}^{t-1})\| \\ \leq \rho^t +\rho^{t-1} + \frac{L}{\gamma^{t-1}}.
		\end{array}
		\end{equation*} 
		Thus, sequence $\{\mathbf{u}^t,{\mathbf{u}}_d^{t}\}$ is a Cauchy sequence. This implies  there exists a fixed point $\mathbf{u}^*$ satisfying $\mathbf{u}^t\to\mathbf{u}^*$ as $t\to\infty$.  
	\end{proof}

	\ifCLASSOPTIONcaptionsoff
	\newpage
	\fi
	
	\bibliographystyle{IEEEtran}
	\bibliography{reference}

	\begin{IEEEbiography}[{\includegraphics[width=1in,height=1.25in,clip,keepaspectratio]{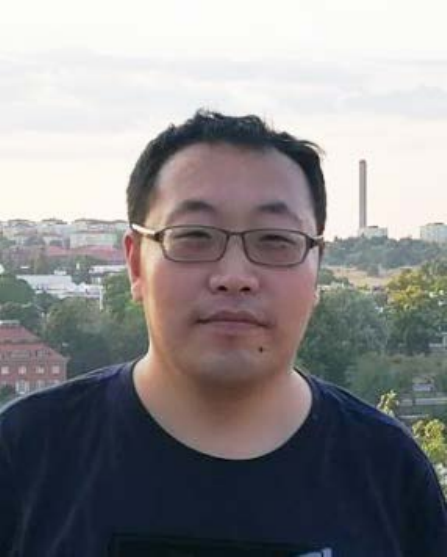}}] {Risheng Liu} received his B.Sc. (2007) and Ph.D. (2012) from Dalian University of Technology, China. From 2010 to 2012, he was doing research as joint Ph.D. in robotics institute at Carnegie Mellon University. From 2016 to 2018, He was doing research as Hong Kong Scholar at the Hong Kong Polytechnic University. He is currently a full professor with the Digital Media Department at International School of Information Science \& Engineering, Dalian University of Technology (DUT). He was awarded the ``Outstanding Youth Science Foundation" of the National Natural Science Foundation of China. He serves as associate editor for Pattern Recognition, the Journal of Electronic Imaging (Senior Editor), The Visual Computer, and IET Image Processing. His research interests include optimization, computer vision and multimedia.
	\end{IEEEbiography}
	\vspace{-1.2cm}
	\begin{IEEEbiography}[{\includegraphics[width=1in,height=1.25in,clip,keepaspectratio]{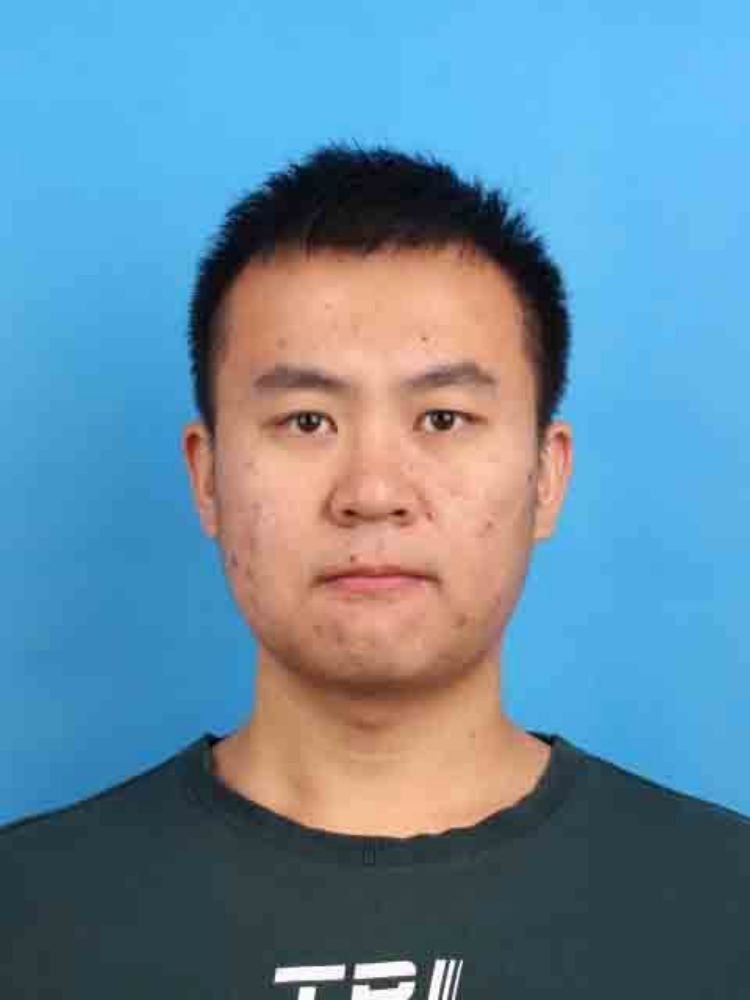}}] {Zhu Liu} received the B.S. degree in software
		engineering from the Dalian University of Technology, Dalian, China, in 2019. He received his M.S. degree in Software Engineering at Dalian University of Technology, Dalian, China, in 2022. He is currently pursuing the Ph.D. degree in Software Engineering at Dalian University of Technology, Dalian, China. His research interests include 
		optimization, image processing and fusion.
	\end{IEEEbiography}
		\vspace{-1.2cm}
	\begin{IEEEbiography}[{\includegraphics[width=1in,height=1.25in,clip,keepaspectratio]{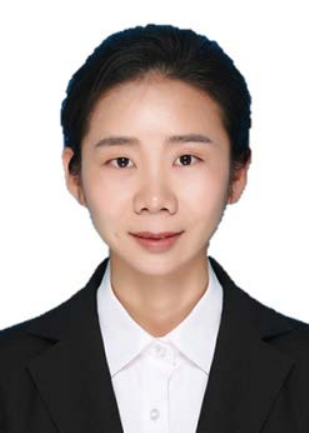}}] {Pan Mu} received the B.S. degree in Applied Mathematics from Henan University, China, in 2014, the M.S. degree in Operational Research and Cybernetics from Dalian University of Technology, China, in 2017 and the PhD degree in Computational Mathematics at Dalian University of Technology, Dalian, China, in 2021. She is currently a lecture with the Zhejiang University of Technology, Hangzhou, China. Her research interests include computer vision, machine learning, optimization and control.
	\end{IEEEbiography}
		\vspace{-1.2cm}
	\begin{IEEEbiography}[{\includegraphics[width=1in,height=1.25in,clip,keepaspectratio]{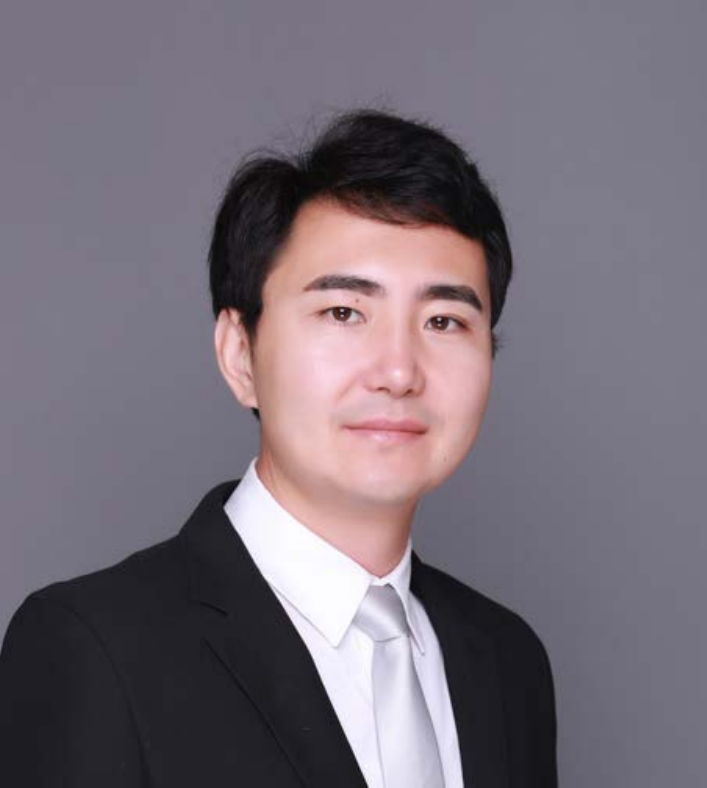}}] {Xin Fan} received the B.E. and Ph.D. degrees in information and communication engineering from Xian Jiaotong University, Xian, China, in 1998 and 2004, respectively. He was with Oklahoma State University, Stillwater, from 2006 to 2007, as a post-doctoral research fellow. He joined the School of Software, Dalian University of Technology, Dalian, China, in 2009. His current research interests include computational geometry and machine learning, and their applications to low-level image processing and DTI-MR image analysis.
	\end{IEEEbiography}
		\vspace{-1.2cm}
	\begin{IEEEbiography}[{\includegraphics[width=1in,height=1.25in,clip,keepaspectratio]{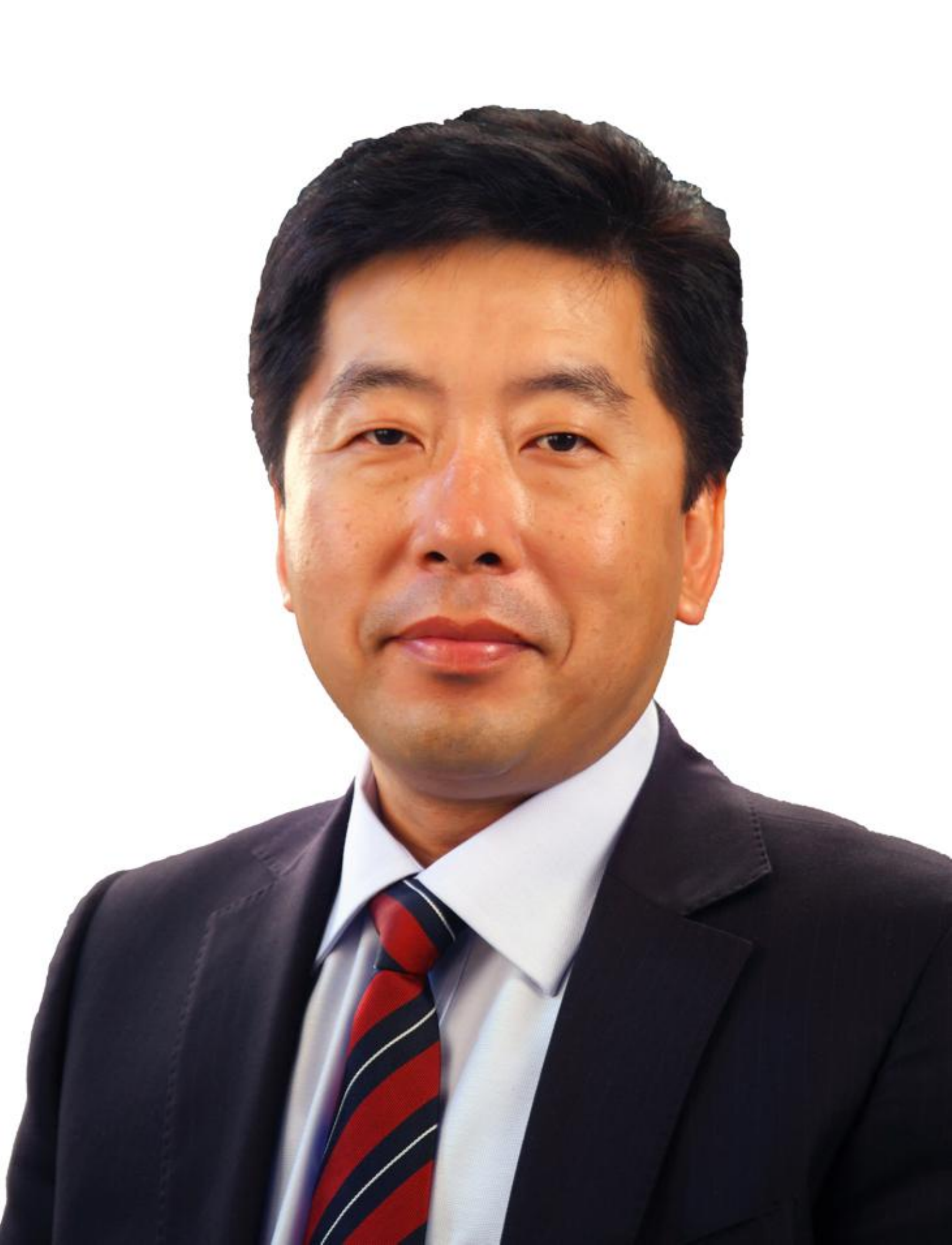}}] {Zhongxuan Luo} received the B.S. degree in Computational Mathematics from Jilin University, China, in 1985, the M.S. degree in Computational Mathematics from Jilin University in 1988, and the Ph.D. degree in Computational Mathematics from Dalian University of Technology, China, in 1991. He has been a full professor of the School of Mathematical Sciences at Dalian University of Technology since 1997. His research interests include computational geometry and computer vision.
	\end{IEEEbiography}
\end{document}